\documentclass[10pt,twocolumn,letterpaper]{article}

\usepackage{cvpr}
\usepackage{times}
\usepackage{epsfig}
\usepackage{graphicx}
\usepackage{amsmath}
\usepackage{amssymb}

\usepackage[pagebackref=true,breaklinks=true,letterpaper=true,colorlinks,bookmarks=false]{hyperref}

\cvprfinalcopy 

\def\cvprPaperID{2201} 

\ifcvprfinal\fi

\usepackage[]{graphicx}
\usepackage{amsfonts, amssymb}
\usepackage{mdwmath}
\usepackage{mdwtab}
\usepackage[tight,footnotesize]{subfigure}
\usepackage{algorithm}	
\usepackage{algpseudocode}
\usepackage{balance}
\usepackage{comment}
\usepackage{gensymb}
\usepackage{multirow}
\usepackage{comment}
\usepackage[]{appendix}

\begin{document}

\title{SO-Net: Self-Organizing Network for Point Cloud Analysis}

\author{Jiaxin Li \qquad Ben M. Chen \qquad Gim Hee Lee \\
National University of Singapore
}

\maketitle

\begin{abstract}
This paper presents SO-Net, a permutation invariant architecture for deep learning with orderless point clouds. The SO-Net models the spatial distribution of point cloud by building a Self-Organizing Map (SOM). Based on the SOM, SO-Net performs hierarchical feature extraction on individual points and SOM nodes, and ultimately represents the input point cloud by a single feature vector. The receptive field of the network can be systematically adjusted by conducting point-to-node $k$ nearest neighbor search. In recognition tasks such as point cloud reconstruction, classification, object part segmentation and shape retrieval, our proposed network demonstrates performance that is similar with or better than state-of-the-art approaches. In addition, the training speed is significantly faster than existing point cloud recognition networks because of the parallelizability and simplicity of the proposed architecture. Our code is available at the project website.\footnote{https://github.com/lijx10/SO-Net}

\end{abstract}
\section{Introduction} \label{sec_intro}
After many years of intensive research, convolutional neural networks (ConvNets) is now the foundation for many state-of-the-art computer vision algorithms, \eg image recognition, object classification and semantic segmentation etc. Despite the great success of ConvNets for 2D images, the use of deep learning on 3D data remains a challenging problem. Although 3D convolution networks (3D ConvNets) can be applied to 3D data that is rasterized into voxel representations, most computations are redundant because of the sparsity of most 3D data. Additionally, the performance of naive 3D ConvNets is largely limited by the resolution loss and exponentially growing computational cost. Meanwhile, the accelerating development of depth sensors, and the huge demand from applications such as autonomous vehicles make it imperative to process 3D data efficiently. Recent availability of 3D datasets including ModelNet \cite{wu20153d}, ShapeNet \cite{chang2015shapenet}, 2D-3D-S \cite{2017arXiv170201105A} adds on to the popularity of research on 3D data. 

To avoid the shortcomings of naive voxelization, one option is to explicitly exploit the sparsity of the voxel grids \cite{wang2015voting, li2016fpnn, engelcke2017vote3deep}. Although the sparse design allows higher grid resolution, the induced complexity and limitations make it difficult to realize large scale or flexible deep networks \cite{Riegler2017OctNet}. Another option is to utilize scalable indexing structures including kd-tree \cite{bentley1975multidimensional}, octree \cite{meagher1980octree}. Deep networks based on these structures have shown encouraging results. Compared to tree based structures, point cloud representation is mathematically more concise and straight-forward because each point is simply represented by a 3-vector. Additionally, point clouds can be easily acquired with popular sensors such as the RGB-D cameras, LiDAR, or conventional cameras with the help of the Structure-from-Motion (SfM) algorithm. Despite the widespread usage and easy acquisition of point clouds, recognition tasks with point clouds still remain challenging. Traditional deep learning methods such as ConvNets are not applicable because point clouds are spatially irregular, and can be permutated arbitrarily. Due to these difficulties, few attempts has been made to apply deep learning techniques directly to point clouds until the very recent PointNet \cite{qi2016pointnet}. 

\begin{figure}[t] \centering
\includegraphics[width=0.49\textwidth]{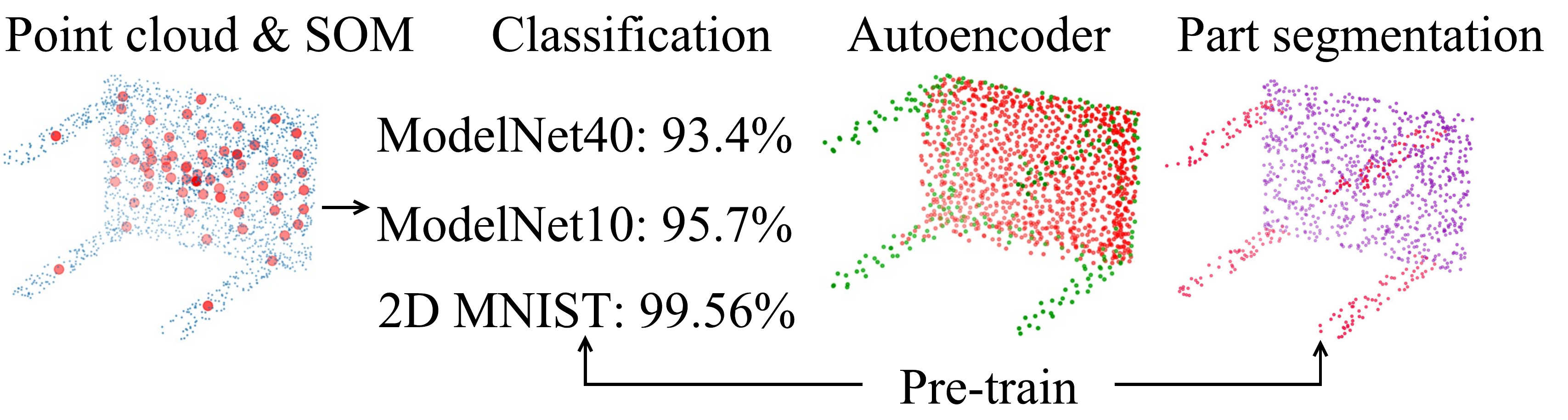}
\caption{Our SO-Net applies hierarchical feature aggregation using SOM. Point clouds are converted into SOM node features and a global feature vector that can be applied to classification, autoencoder reconstruction, part segmentation and shape retrieval etc. 
} \label{fig_intro}
\vspace{-4pt}
\end{figure}
%

Despite being a pioneer in applying deep learning to point clouds, PointNet is unable to handle local feature extraction adequately. PointNet++ \cite{qi2017pointnet++} is later proposed to address this problem by building a pyramid-like feature aggregation scheme, but the point sampling and grouping strategy in \cite{qi2017pointnet++} does not reveal the spatial distribution of the input point cloud. Kd-Net \cite{klokov2017escape} build a kd-tree for the input point cloud, followed by hierarchical feature extractions from the leaves to root. Kd-Net explicitly utilizes the spatial distribution of point clouds, but there are limitations such as the lack of overlapped receptive fields.

In this paper, we propose the SO-Net to address the problems in existing point cloud based networks. Specifically, a SOM \cite{kohonen1998self} is built to model the spatial distribution of the input point cloud, which enables hierarchical feature extraction on both individual points and SOM nodes. Ultimately, the input point cloud can be compressed into a single feature vector. During the feature aggregation process, the receptive field overlap is controlled by performing point-to-node k-nearest neighbor (kNN) search on the SOM. The SO-Net theoretically guarantees invariance to the order of input points, by the network design and our permutation invariant SOM training. Applications of our SO-Net include point cloud based classification, autoencoder reconstruction, part segmentation and shape retrieval, as shown in Fig.~\ref{fig_intro}.
\\\\
\noindent The \textbf{key contributions} of this paper are as follows:
\begin{itemize}
    \item We design a permutation invariant network - the SO-Net that explicitly utilizes the spatial distribution of point clouds. 
    \item With point-to-node $k$NN search on SOM, hierarchical feature extraction is performed with systematically adjustable receptive field overlap.
    \item We propose a point cloud autoencoder as pre-training to improve network performance in various tasks.
    \item Compared with state-of-the-art approaches, similar or better performance is achieved in various applications with significantly faster training speed.
\end{itemize}

\section{Related Work}  \label{sec_related_work}
It is intuitive to represent 3D shapes with voxel grids because they are compatible with 3D ConvNets. \cite{maturana2015voxnet,wu20153d} use binary variable to indicate whether a voxel is occupied or free. Several enhancements are proposed in \cite{qi2016volumetric} - overfitting is mitigated by predicting labels from partial subvolumes, orientation pooling layer is designed to fuse shapes with various orientations, and anisotropic probing kernels are used to project 3D shapes into 2D features. Brock \etal \cite{brock2016generative} propose to combine voxel based variational autoencoders with object recognition networks. Despite its simplicity, voxelization is able to achieve state-of-the-art performance. Unfortunately, it suffers from loss of resolution and the exponentially growing computational cost. Sparse methods \cite{wang2015voting, li2016fpnn, engelcke2017vote3deep} are proposed to improve the efficiency. However, these methods still rely on uniform voxel grids and experience various limitations such as the lack of parallelization capacity \cite{li2016fpnn}. Spectral ConvNets \cite{masci2015geodesic,boscaini2015learning,bruna2013spectral} are explored to work on non-Euclidean geometries, but they are mostly limited to manifold meshes.

Rendering 3D data into multi-view 2D images turns the 3D problem into a 2D problem that can be solved using standard 2D ConvNets.
View-pooling layer \cite{su2015multi} is designed to aggregate features from multiple rendered images. Qi \etal \cite{qi2016volumetric} substitute traditional 3D to 2D rendering with multi-resolution sphere rendering. Wang \etal \cite{wang2017dominant} further propose the dominant set pooling and utilize features like color and surface normal. Despite the improved efficiency compared to 3D ConvNets, multi-view strategy still suffers from information loss \cite{klokov2017escape} and it cannot be easily extended to tasks like per-point labeling.

Indexing techniques such as kd-tree and octree are scalable compared to uniform grids, and their regular structures are suitable for deep learning techniques. To enable convolution and pooling operations over octree, Riegler \etal \cite{Riegler2017OctNet} build a hybrid grid-octree structure by placing several small octrees into a regular grid. With bit string representation, a single voxel in the hybrid structure is fully determined by its bit index. As a result, simple arithmetic can be used to visit the parent or child nodes. Similarly, Wang \etal \cite{wang2017cnn} introduce a label buffer to find correspondence of octants at various depths. Klokov \etal propose the Kd-Net \cite{klokov2017escape} that computes vectorial representations for each node of the pre-built balanced kd-tree. A parent feature vector is computed by applying non-linearity and affine transformation on its two child feature vectors, following the bottom-up fashion.

PointNet \cite{qi2016pointnet} is the pioneer in the direct use of point clouds. It uses the channel-wise max pooling to aggregate per-point features into a global descriptor vector. PointNet is invariant to order permutation of input points because the per-point feature extraction is identical for every point and max pooling operation is permutation invariant. A similar permutation equivariant layer \cite{ravanbakhsh2016deep} is also proposed at almost the same time as \cite{qi2016pointnet}, with the major difference that the permutation equivariant layer is max-normalized. Although the max-pooling idea is proven to be effective, it suffers from the lack of ConvNet-like hierarchical feature aggregation. PointNet++ \cite{qi2017pointnet++} is later designed to group points into several groups in different levels, so that features from multiple scales could be extracted hierarchically.

Unlike networks based on octree or kd-tree, the spatial distribution of points is not explicitly modeled in PointNet++. Instead, heuristic grouping and sampling schemes, \eg multi-scale and multi-resolution grouping, are designed to combine features from multiple scales. In this paper, we propose our SO-Net that explicitly models the spatial distribution of input point cloud during hierarchical feature extraction. In addition, adjustable receptive field overlap leads to more effective local feature aggregation.

\section{Self-Organizing Network}
\begin{figure}[t!]
        \centering
        \subfigure[]{\label{fig_potential_field}\includegraphics[width=0.23\textwidth]{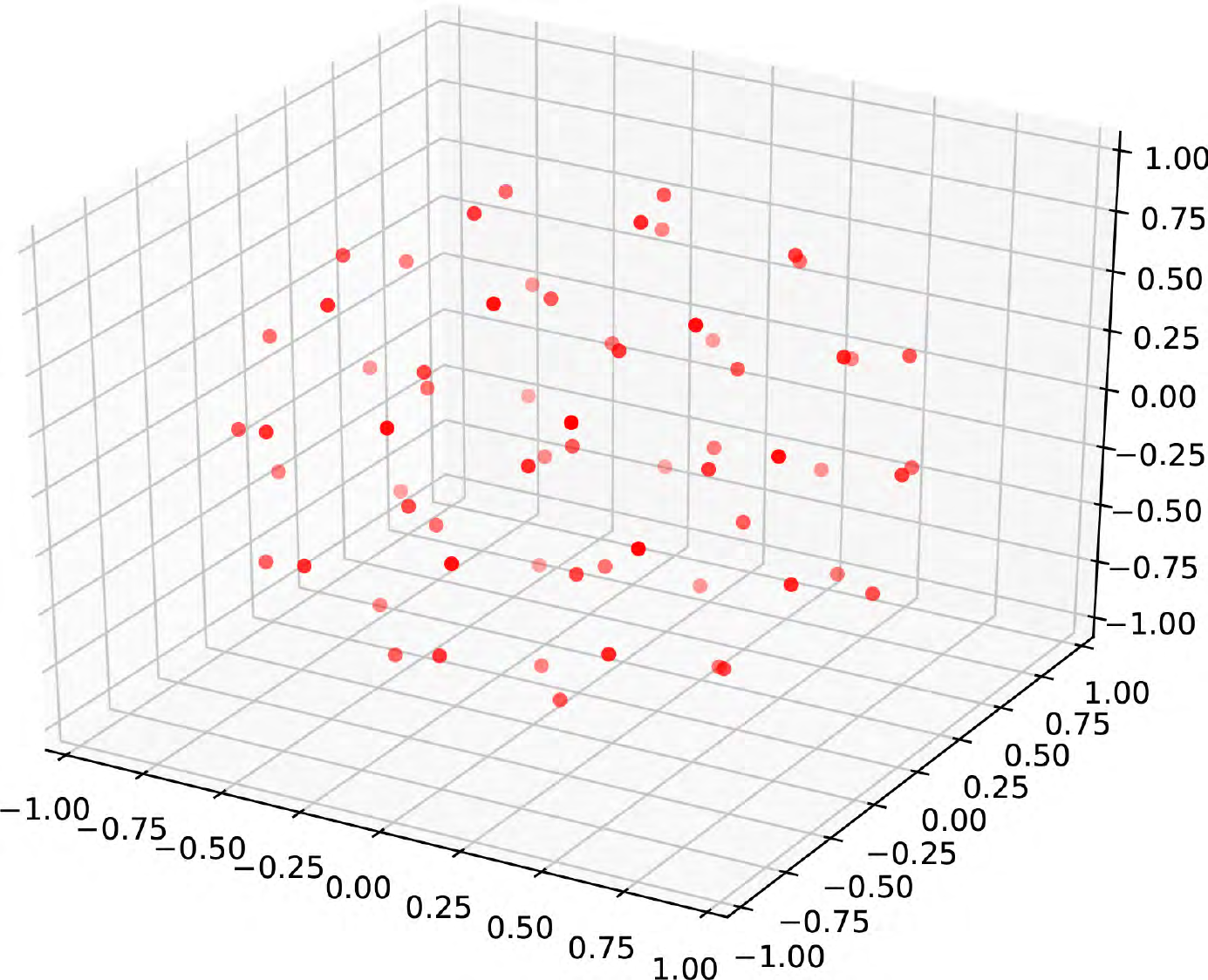}}
        \subfigure[]{\label{fig_som_result}\includegraphics[width=0.23\textwidth]{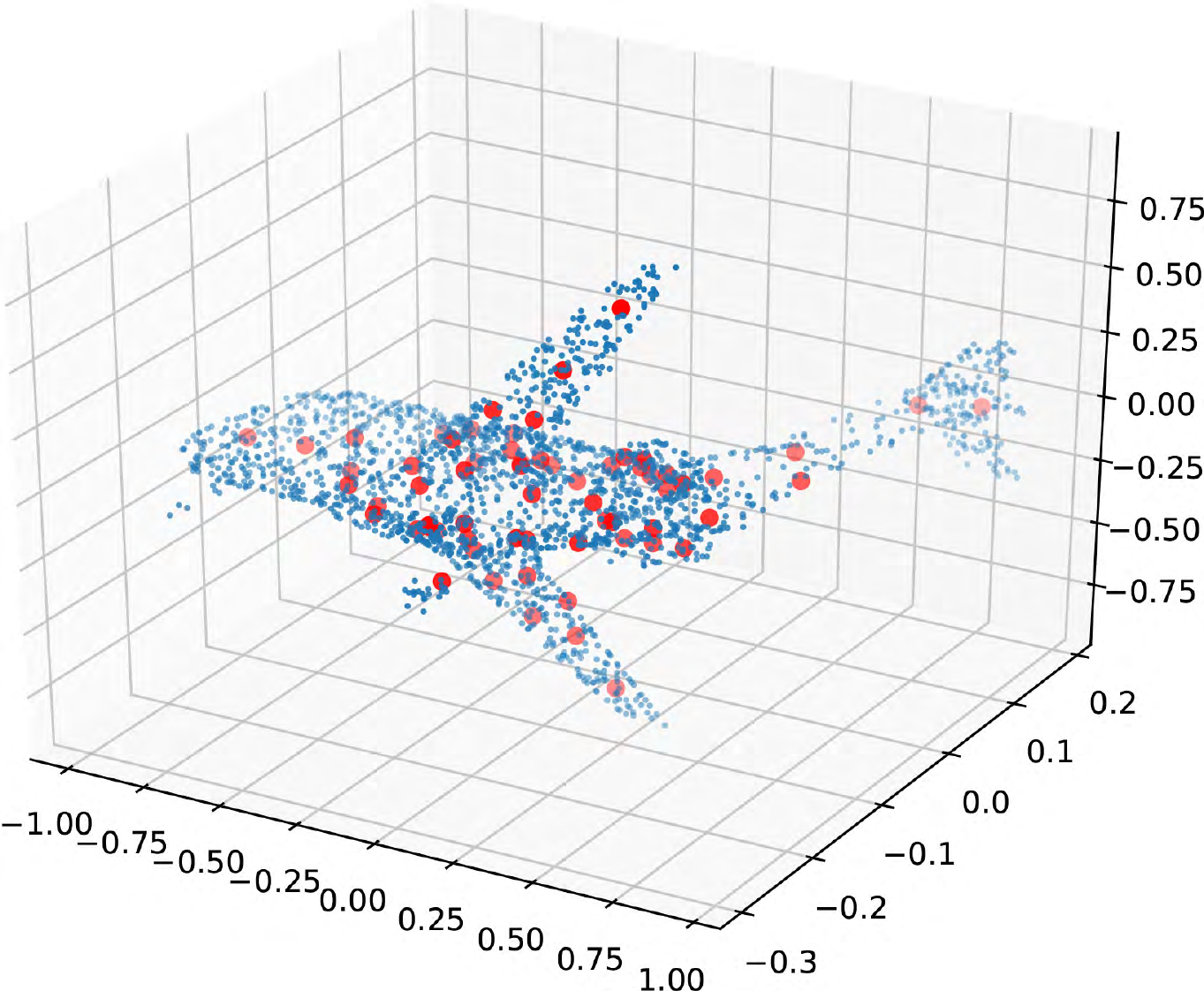}}
        \caption{(a) The initial nodes of an $8\times 8$ SOM. For each SOM configuration, the initial nodes are fixed for every point cloud. (b) Example of a SOM training result.}
        \vspace{-4pt}
\end{figure}
\begin{figure*}[t!] \centering
\includegraphics[width=0.98\textwidth]{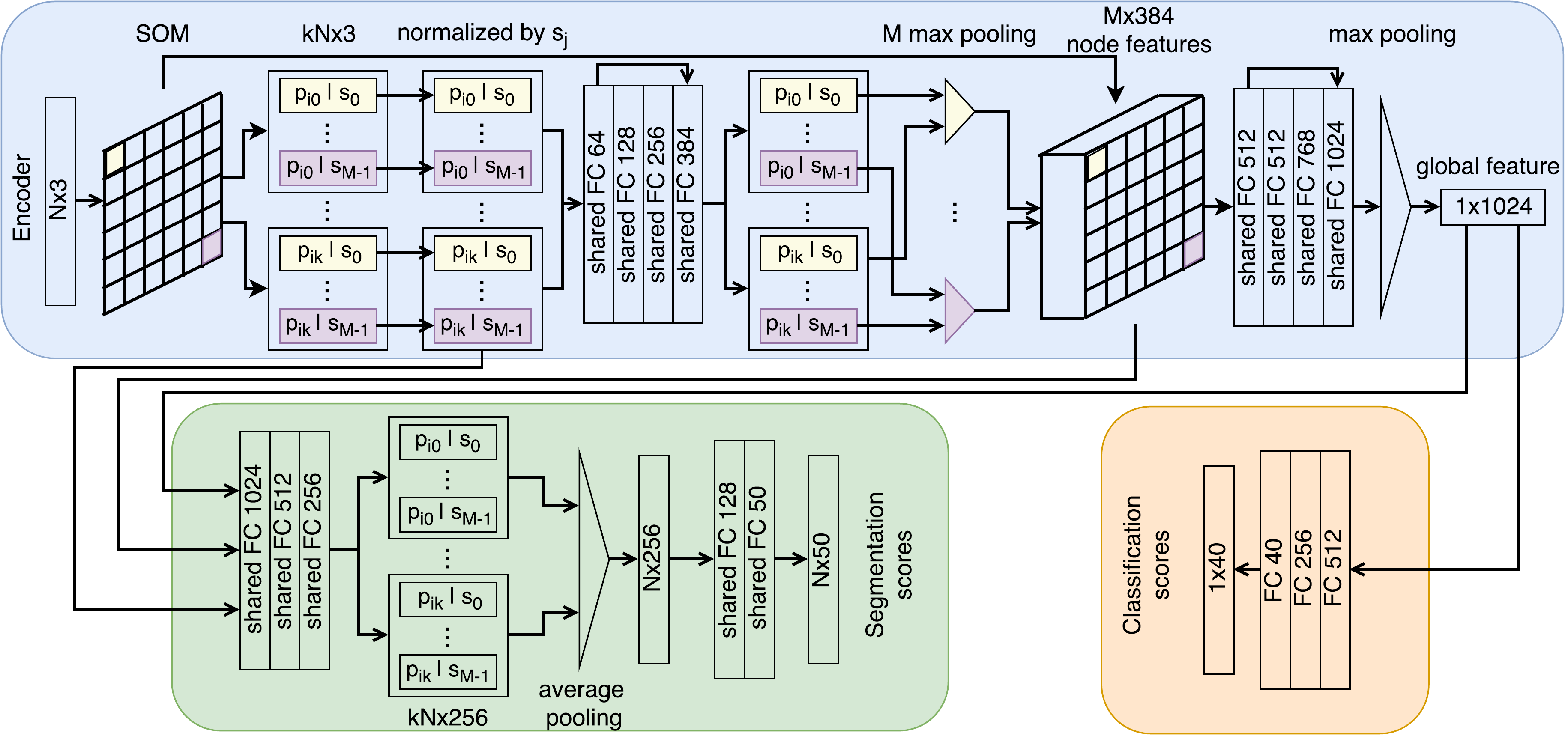}
\caption{The architecture of the SO-Net and its application to classification and segmentation. In the encoder, input points are normalized with the $k$-nearest SOM nodes. The normalized point features are later max-pooled into node features based on the point-to-node kNN search on SOM. $k$ determines the receptive field overlap. In the segmentation network, $M$ node features are concatenated with the $kN$ normalized points following the same kNN association. Finally $kN$ features are aggregated into $N$ features by average pooling. }\label{fig_architecture}
\vspace{-4pt}
\end{figure*}

The input to the network is a point set $P=\{p_i \in \mathbb{R}^3,~i = 0, \cdots, N-1\}$, which will be processed into $M$ SOM nodes $S=\{s_j \in \mathbb{R}^3,~j = 0, \cdots, M-1\}$ as shown in Sec.~\ref{sec_som}. Similarly, in the encoder described in Sec.~\ref{sec_encoder}, individual point features are max-pooled into $M$ node features, which can be further aggregated into a global feature vector. Our SO-Net can be applied to various computer vision tasks including classification, per-point segmentation (Sec.~\ref{sec_segmentation}), and point cloud reconstruction (Sec.~\ref{sec_autoencoder}).

\subsection{Permutation Invariant SOM} \label{sec_som}

SOM is used to produce low-dimensional, in this case two-dimensional, representation of the input point cloud. We construct a SOM with the size of $m \times m$, where $m \in [5,11]$, i.e. the total number of nodes $M$ ranges from 25 to 121. SOM is trained with unsupervised competitive learning instead of the commonly used backpropagation in deep networks. However, naive SOM training schemes are not permutation invariant for two reasons: the training result is highly related to the initial nodes, and the per-sample update rule depends on the order of the input points. 

The first problem is solved by assigning fixed initial nodes for any given SOM configuration. Because the input point cloud is normalized to be within $[-1,1]$ in all three axes, we generate a proper initial guess by dispersing the nodes uniformly inside a unit ball, as shown in Fig.~\ref{fig_potential_field}. Simple approaches such as the potential field can be used to construct such a uniform initial guess. To solve the second problem, instead of updating nodes once per point, we perform one update after accumulating the effects of all the points. This batch update process is deterministic \cite{kohonen1998self} for a given point cloud, making it permutation invariant. Another advantage of batch update is the fact that it can be implemented as matrix operations, which are highly efficient on GPU. Details of the initialization and batch training algorithms can be found in our supplementary material.

\subsection{Encoder Architecture} \label{sec_encoder}
As shown in Fig.~\ref{fig_architecture}, SOM is a guide for hierarchical feature extraction, and a tool to systematically adjust the receptive field overlap. Given the output of the SOM, we search for the $k$ nearest neighbors ($k$NN) on the SOM nodes $S$ for each point $p_i$, i.e., point-to-node $k$NN search:
\begin{equation}
s_{ik} = \text{kNN}(p_i \mid s_j,~j = 0, \cdots, M-1).
\end{equation}

\noindent Each $p_i$ is then normalized into $k$ points by subtraction with its associated nodes:
\begin{equation} \label{equ_pik}
    p_{ik} = p_i - s_{ik}.
\end{equation}

\noindent The resulting $kN$ normalized points are forwarded into a series of fully connected layers to extract individual point features. There is a shared fully connected layer on each level $l$, where $\phi$ is the non-linear activation function. The output of level $l$ is given by
\begin{equation} \label{equ_fc}
    p_{ik}^{l+1} = \phi(W^l p_{ik}^l + b^l).
\end{equation}

\noindent The input to the first layer $p_{ik}^0$ can simply be the normalized point coordinates $p_{ik}$, or the combination of coordinates and other features like surface normal vectors.

Node feature extraction begins with max-pooling the $kN$ point features into $M$ node features following the above kNN association. We apply a channel-wise max pooling operation to get the node feature $s_j^0$ for those point features associated with the same node $s_j$:
\begin{equation} \label{equ_som_pool}
    s_j^0 = \text{max}(\{p_{ik}^l, \forall s_{ik}=s_j\}).
\end{equation}
Since each point is normalized into $k$ coordinates according to the point-to-node kNN search, it is guaranteed that the receptive fields of the $M$ max pooling operations are overlapped. Specifically, $M$ nodes cover $kN$ normalized points. $k$ is an adjustable parameter to control the overlap. 

Each node feature produced by the above max pooling operation is further concatenated with the associated SOM node. The $M$ augmented node features are forwarded into a series of shared layers, and then aggregated into a feature vector that represents the input point cloud.

\paragraph{Feature aggregation as point cloud separation and assembly}
There is an intuitive reason behind the SOM feature extraction and node concatenation. Since the input points to the first layer are normalized with $M$ SOM nodes, they are actually separated into $M$ mini point clouds as shown in Fig.~\ref{fig_architecture}. Each mini point cloud contains a small number of points in a coordinate whose origin is the associated SOM node. For a point cloud of size 2048, and $M=64$ and $k=3$, a typical mini point cloud may consist of around 90 points inside a small space of $x,y,z\in[-0.3, 0.3]$. The number and coverage of points in a mini point cloud are determined by the SOM training and kNN search, i.e. $M$ and $k$. 

The first batch of fully connected layers can be regarded as a small PointNet that encodes these mini point clouds. The concatenation with SOM nodes plays the role of assembling these mini point clouds back into the original point cloud. Because the SOM explicitly reveals the spatial distribution of the input point cloud, our separate-and-assemble process is more efficient than the grouping strategy used in PointNet++ \cite{qi2017pointnet++}, as shown in Sec.~\ref{sec_experiments}.

\paragraph{Permutation Invariance}
There are two levels of feature aggregation in SO-Net, from point features to node features, and from node features to global feature vector. The first phase applies a shared PointNet to $M$ mini point clouds. The generation of these $M$ mini point clouds is irrelevant to the order of input points, because the SOM training in Sec.~\ref{sec_som} and kNN search in Fig.~\ref{fig_architecture} are deterministic. PointNet \cite{qi2016pointnet} is permutation invariant as well. Consequently, both the node features and global feature vector are theoretically guaranteed to be permutation invariant.

\paragraph{Effect of suboptimal SOM training}
It is possible that the training of SOM converges into a local minima with isolated nodes outside the coverage of the input point cloud. In some situations no point will be associated with the isolated nodes during the point-to-node $k$NN search, and we set the corresponding node features to zero. This phenomenon is quite common because the initial nodes are dispersed uniformly in a unit ball, while the input point cloud may occupy only a small corner. Despite the existence of suboptimal SOM, the proposed SO-Net still out-performs state-of-the-art approaches in applications like object classification. The effect of invalid node features is further investigated in Sec.~\ref{sec_experiments} by inserting noise into the SOM results.

\paragraph{Exploration with ConvNets}
It is interesting to note that the node feature extraction has generated an image-like feature matrix, which is invariant to the order of input points. It is possible to apply standard ConvNets to further fuse the node features with increasing receptive field. However, the classification accuracy decreased slightly in our experiments, where we replaced the second batch of fully connected layers with 2D convolutions and pooling. It remains as a promising direction to investigate the reason and solution to this phenomenon.

\begin{figure*}[t] \centering
\includegraphics[width=0.9\textwidth]{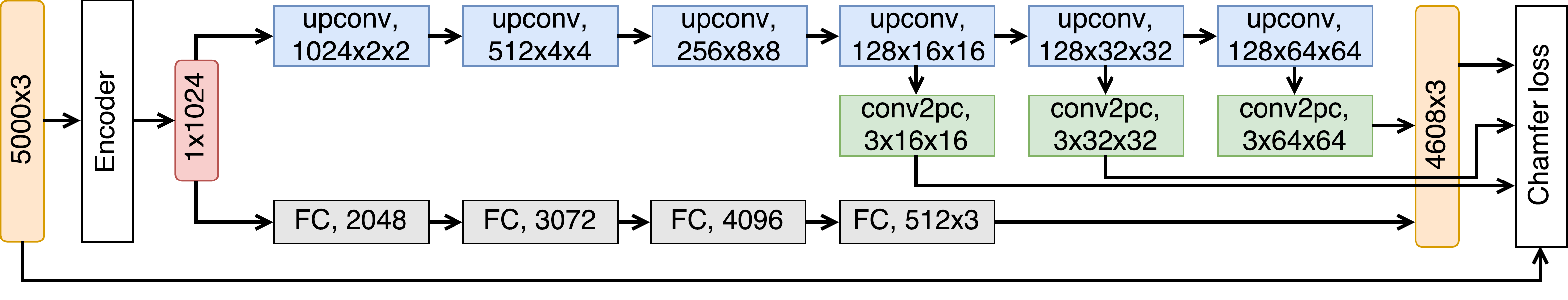}
\caption{The architecture of the decoder that takes 5000 points and reconstructs 4608 points. The up-convolution branch is designed to recover the main body of the input, while the more flexible fully connected branch is to recover the details. The ``upconv'' module consists of a nearest neighbor upsampling layer and a $3\times3$ convolution layer. The ``conv2pc'' module consists of two $1\times1$ convolution layers.}\label{fig_decoder}
\vspace{-4pt}
\end{figure*}

\subsection{Extension to Segmentation} \label{sec_segmentation}
The extension to per-point annotations, \eg segmentation, requires the integration of both local and global features. The integration process is similar to the invert operation of the encoder in Sec.~\ref{sec_encoder}. The global feature vector can be directly expanded and concatenated with the $kN$ normalized points. The $M$ node features are attached to the points that are associated with them during the encoding process. The integration results in $kN$ features that combine point, node and global features, which are then forwarded into a chain of shared fully connected layers.

The $kN$ features are actually redundant to generate $N$ per-point classification scores 
because of the receptive field overlap. Average or max pooling are methods to fuse the redundant information. Additionally, similar to many deep networks, early, middle or late fusion may exhibit different performance \cite{cheng2017locality}. With a series of experiments, we found that middle fusion with average pooling is most effective compared to other fusion methods. 

\subsection{Autoencoder} \label{sec_autoencoder}

In this section, we design a decoder network to recover the input point cloud from the encoded global feature vector. A straightforward design is to stack series of fully connected layers on top of the feature vector, and generate an output vector of length $3\hat{N}$, which can be reshaped into $\hat{N}\times3$. However, the memory and computation footprint will be too heavy if $\hat{N}$ is sufficiently large.

Instead of generating point clouds with fully connected layers \cite{achlioptas2017learning}, we design a network with two parallel branches similar with \cite{fan2016point}, i.e, a fully connected branch and a convolution branch as shown in Fig.~\ref{fig_decoder}. The fully connected branch predicts $\hat{N}_1$ points by reshaping an output of $3\hat{N}_1$ elements. This branch enjoys high flexibility because each coordinate is predicted independently. On the other hand, the convolution branch predicts a feature matrix with the size of $3\times H\times W$, i.e. $\hat{N}_2=H\times W$ points. Due to the spatial continuity of convolution layers, the predicted $\hat{N}_2$ point may exhibit more geometric consistency. Another advantage of the convolution branch is that it requires much less parameters compared to the fully connected branch.

Similar to common practice in many depth estimation networks \cite{garg2016unsupervised,godard2016unsupervised}, the convolution branch is designed as an up-convolution (upconv) chain in a pyramid style. Instead of deconvolution layers, each upconv module consists of a nearest neighbor upsampling layer and a $3\times3$ convolution layer. According to our experiments, this design is much more effective than deconvolution layers in the case of point cloud autoencoder. In addition, intermediate upconv products are converted to coarse reconstructed point clouds and compared with the input. The conversion from upconv products to point clouds is a 2-layer $1\times 1$ convolution stack in order to give more flexibility to each recovered point. The coarse-to-fine strategy produces another boost in the reconstruction performance.

To supervise the reconstruction process, the loss function should be differentiable, ready for parallel computation and robust against outliers \cite{fan2016point}. Here we use the Chamfer loss: 
\begin{equation}\label{equ_chamfer}
\begin{split}
    d(P_s, P_t) = & \frac{1}{|P_s|}\sum_{x\in P_s} \underset{y\in P_t}{\text{min}}\|x-y\|_2 \\
                  & + \frac{1}{|P_t|}\sum_{y\in P_t} \underset{x\in P_s}{\text{min}}\|x-y\|_2.
\end{split}
\end{equation}

\noindent where $P_s$ and $P_t \in \mathbb{R}^3$ represents the input and recovered point cloud respectively. The numbers of points in $P_s$ and $P_t$ are not necessarily the same. Intuitively, for each point in $P_s$, Eq.~(\ref{equ_chamfer}) computes its distance to the nearest neighbor in $P_t$, and vice versa for points in $P_t$. 

\section{Experiments} \label{sec_experiments}

In this section, the performance of our SO-Net is evaluated in three different applications, namely point cloud autoencoder, object classification and object part segmentation. In particular, the encoder trained in the autoencoder can be used as pre-training for the other two tasks. The encoder structure and SOM configuration remain identical among all experiments without delicate finetuning, except for the 2D MNIST classification.

\subsection{Implementation Detail}
Our network is implemented with PyTorch on a NVIDIA GTX1080Ti. We choose a SOM of size $8\times 8$ and $k=3$ in most experiments. We optimize the networks using Adam \cite{kingma2014adam} with an initial learning rate of 0.001 and batch size of 8. For experiments with 5000 or more points as input, the learning rate is decreased by half every 20 epochs, otherwise the learning rate decay is executed every 40 epochs. Generally the networks converge after around 5 times of learning rate decay. Batch-normalization and ReLU activation are applied to every layer.

\subsection{Datasets}
As a 2D toy example, we adopt the MNIST dataset \cite{lecun1998gradient} in Sec.~\ref{sec_exp_cls}. For each digit, 512 two-dimensional points are sampled from the non-zero pixels to serve as our input.

Two variants of the ModelNet \cite{wu20153d}, i.e. ModelNet10 and ModelNet40, are used as the benchmarks for the autoencoder task in Sec.~\ref{sec_exp_ae} and the classification task in Sec.~\ref{sec_exp_cls}. The ModelNet40 contains 13,834 objects from 40 categories, among which 9,843 objects belong to training set and the other 3,991 samples for testing. Similarly, the ModelNet10 is split into 2,468 training samples and 909 testing samples. The original ModelNet provides CAD models represented by vertices and faces. Point clouds are generated by sampling from the models uniformly. For fair comparison, we use the prepared ModelNet10/40 dataset from \cite{qi2017pointnet++}, where each model is represented by 10,000 points. Various sizes of point clouds, \eg, 2,048 or 5,000, can be sampled from the 10k points in different experiments.

Object part segmentation is demonstrated with the ShapeNetPart dataset \cite{yi2016scalable}. It contains 16,881 objects from 16 categories, represented as point clouds. Each object consists of 2 to 6 parts, and in total there are 50 parts in the dataset. We sample fixed size point clouds, \eg 1,024, in our experiments.

\paragraph{Data augmentation}
Input point clouds are normalized to be zero-mean inside a unit cube. The following data augmentations are applied at training phase: (a) Gaussian noise $\mathcal{N}(0, 0.01)$ is added to the point coordinates and surface normal vectors (if applicable). (b) Gaussian noise $\mathcal{N}(0, 0.04)$ is added to the SOM nodes. (c) Point clouds, surface normal vectors (if applicable) and SOM nodes are scaled by a factor sampled from an uniform distribution $\mathcal{U}(0.8, 1.2)$. Further augmentation like random shift or rotation do not improve the results.

\subsection{Point Cloud Autoencoder} \label{sec_exp_ae}
%
\begin{figure}[h]
        \centering
        \includegraphics[width=0.15\textwidth]{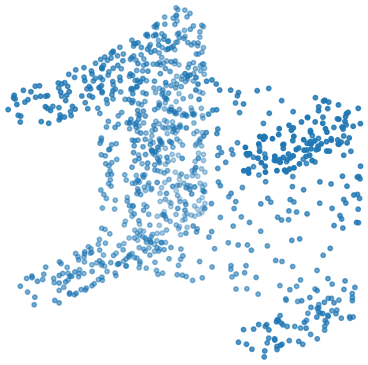}
        \includegraphics[width=0.15\textwidth]{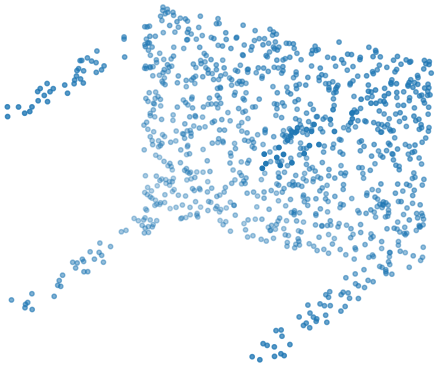}
        \includegraphics[width=0.15\textwidth]{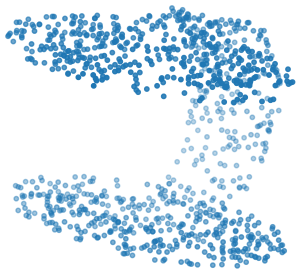}
        \includegraphics[width=0.15\textwidth]{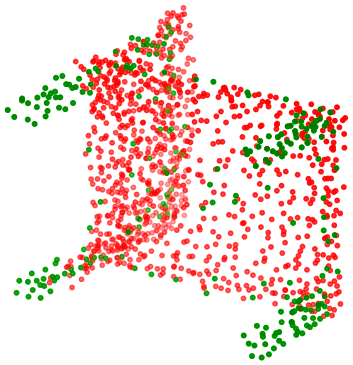}
        \includegraphics[width=0.15\textwidth]{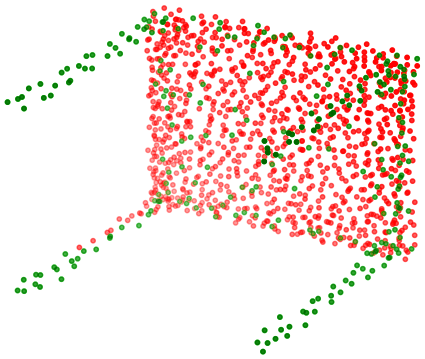}
        \includegraphics[width=0.15\textwidth]{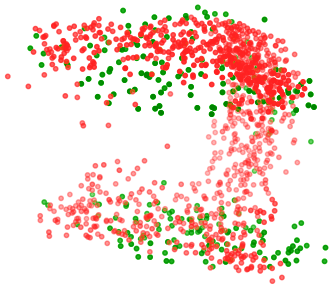}
        \caption{Examples of point cloud autoencoder results. First row: input point clouds of size 1024. Second row: reconstructed point clouds of size 1280. From left to right: chair, table, earphone.}
        \label{fig_ae_selected}
        \vspace{-4pt}
\end{figure}
%

\begin{table*}[t]
\centering
{
\setlength\tabcolsep{4.5pt} 
\begin{tabular}{ll||c|cc|ccc||cc}
\hline
\multirow{2}{*}{Method} & \multirow{2}{*}{Representation} & \multirow{2}{*}{Input} & \multicolumn{2}{c|}{ModelNet10} & \multicolumn{3}{c||}{ModelNet40}  & \multicolumn{2}{c}{MNIST} \\
                        &                                 &                        & Class         & Instance        & Class & Instance & Training & Input      & Error rate     \\ \hline
PointNet, \cite{qi2016pointnet}& points                          & $1024\times 3$                   & -             & -               & 86.2  & 89.2     & 3-6h          & $256\times 2$      & 0.78         \\
PointNet++, \cite{qi2017pointnet++}& points + normal                 & $5000\times 6$                   & -             & -               & -     & 91.9     & 20h           & $512\times 2$      & 0.51         \\
DeepSets, \cite{ravanbakhsh2016deep, zaheer2017deep}& points                          & $5000\times 3$                   & -             & -               & -     & 90.0     & -             &-            & -            \\
Kd-Net, \cite{klokov2017escape}& points                          & $2^{15}\times 3$   & 93.5          & 94.0            & 88.5  & 91.8     & 120h          & $1024\times 2$     & 0.90         \\
ECC, \cite{simonovsky2017dynamic}& points                          & $1000\times3$                   & 90.0          & 90.8            & 83.2  & 87.4     & -             & -          & 0.63         \\
OctNet, \cite{Riegler2017OctNet}& octree                          & $128^3$  & 90.1          & 90.9            & 83.8  & 86.5     & -             &-            & -            \\
O-CNN, \cite{wang2017cnn}& octree                          & $64^3$   & -             & -               & -     & 90.6     & -             &-            & -            \\ 
\hline
Ours (2-layer)*         & points + normal                 & $5000\times 6$                   & 94.9          & 95.0            & 89.4  & 92.5     & 3h            & -      & -         \\
Ours (2-layer)          & points + normal                 & $5000\times 6$                   & 94.4          & 94.5            & 89.3  & 92.3     & 3h            & -           & -            \\
Ours (2-layer)          & points                          & $2048\times 3$                   & 93.9          & 94.1            & 87.3  & 90.9     & 3h            & $512\times 2$            & \textbf{0.44}            \\ \hline
Ours (3-layer)          & points + normal                 & $5000\times 6$                   & \textbf{95.5} & \textbf{95.7}   & \textbf{90.8} & \textbf{93.4}     & \textbf{3h}            & -      & -         \\
\hline
\end{tabular}
}
\caption{Object classification results for methods using scalable 3D representations like point cloud, kd-tree and octree. Our network produces the best accuracy with significantly faster training speed. * represents pre-training.}
\vspace{-4pt}
\label{tbl_cls}
\end{table*}

\begin{figure*}[h!] 
        \centering
        \subfigure[]{\includegraphics[width=0.24\textwidth]{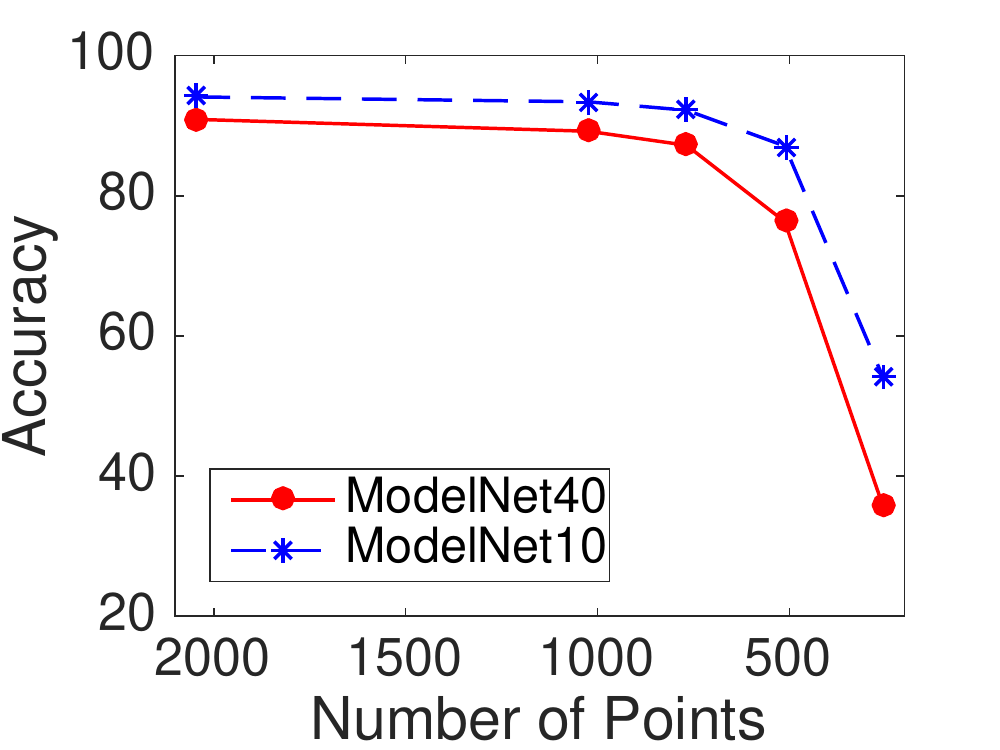} \label{fig_cls_robustness_point}}
        \subfigure[]{\includegraphics[width=0.24\textwidth]{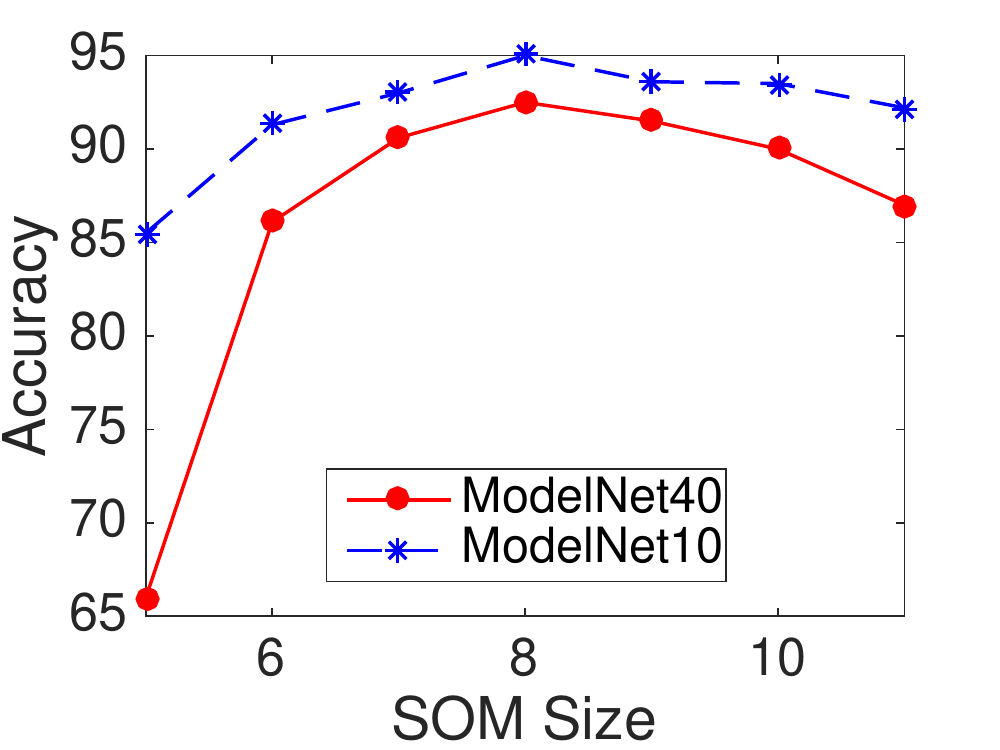} \label{fig_cls_robustness_som_size}}
        \subfigure[]{\includegraphics[width=0.24\textwidth]{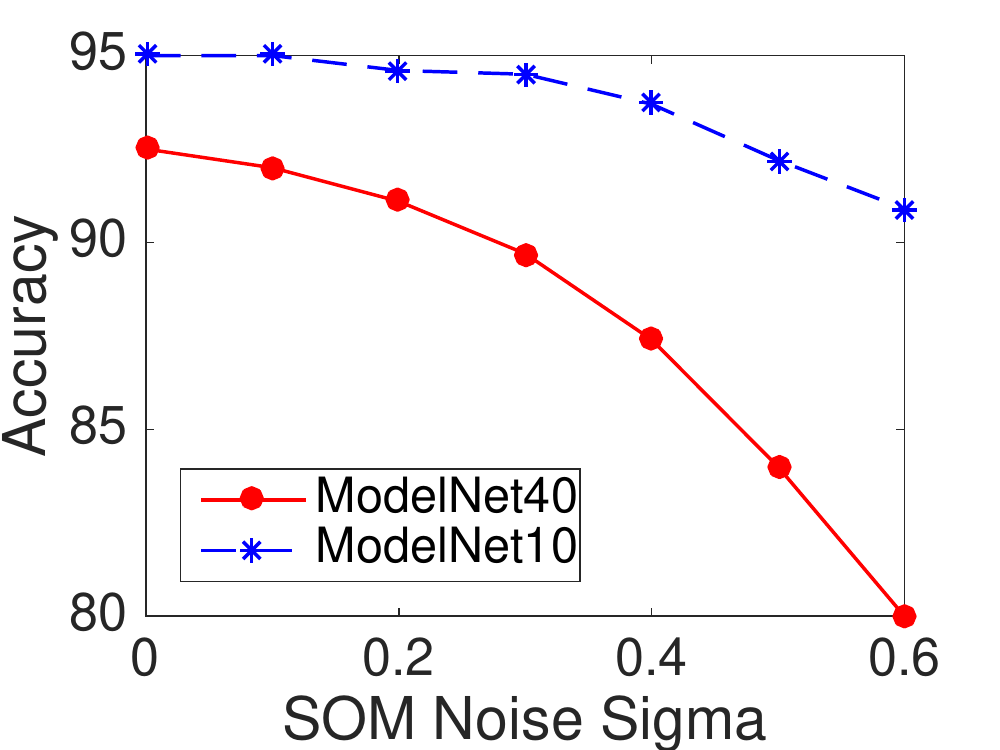} \label{fig_cls_robustness_som_noise}}
        \subfigure[]{\includegraphics[width=0.21\textwidth]{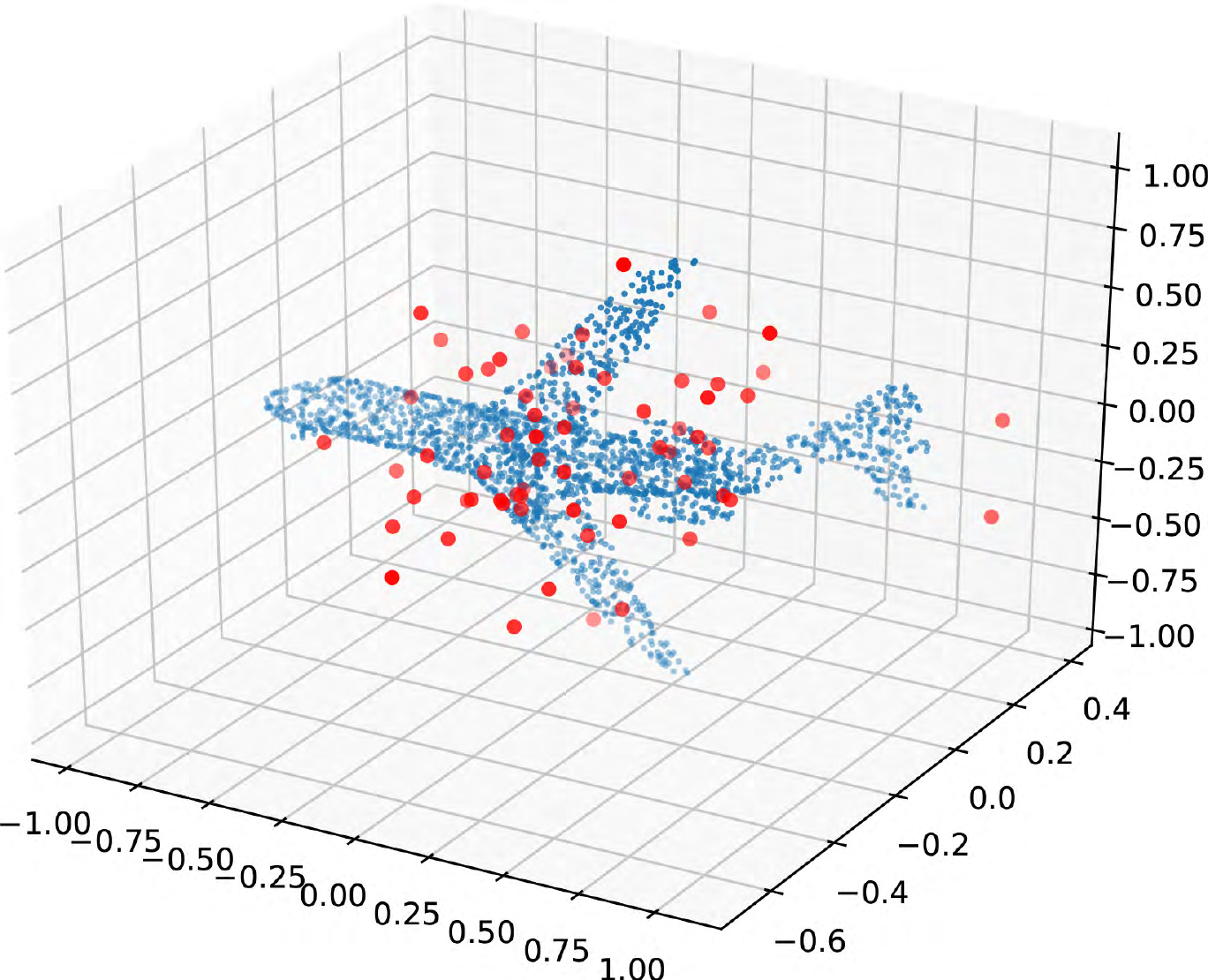} \label{fig_cls_robustness_som_noisy_example}}
        \caption{Robustness test on point or SOM corruption. (a) The network is trained with point clouds of size 2048, while there is random point dropout during testing. (b) The network is trained with SOM of size $8\times 8$, but SOMs of various sizes are used at testing phase. (c) Gaussian noise $\mathcal{N}(0, \sigma)$ is added to the SOM during testing. (d) Example of SOM with Gaussian noise $\mathcal{N}(0, 0.2)$.} \label{fig_cls_robustness}
        \vspace{-4pt}
\end{figure*}
\begin{figure}[t] 
        \centering
        \includegraphics[width=0.23\textwidth]{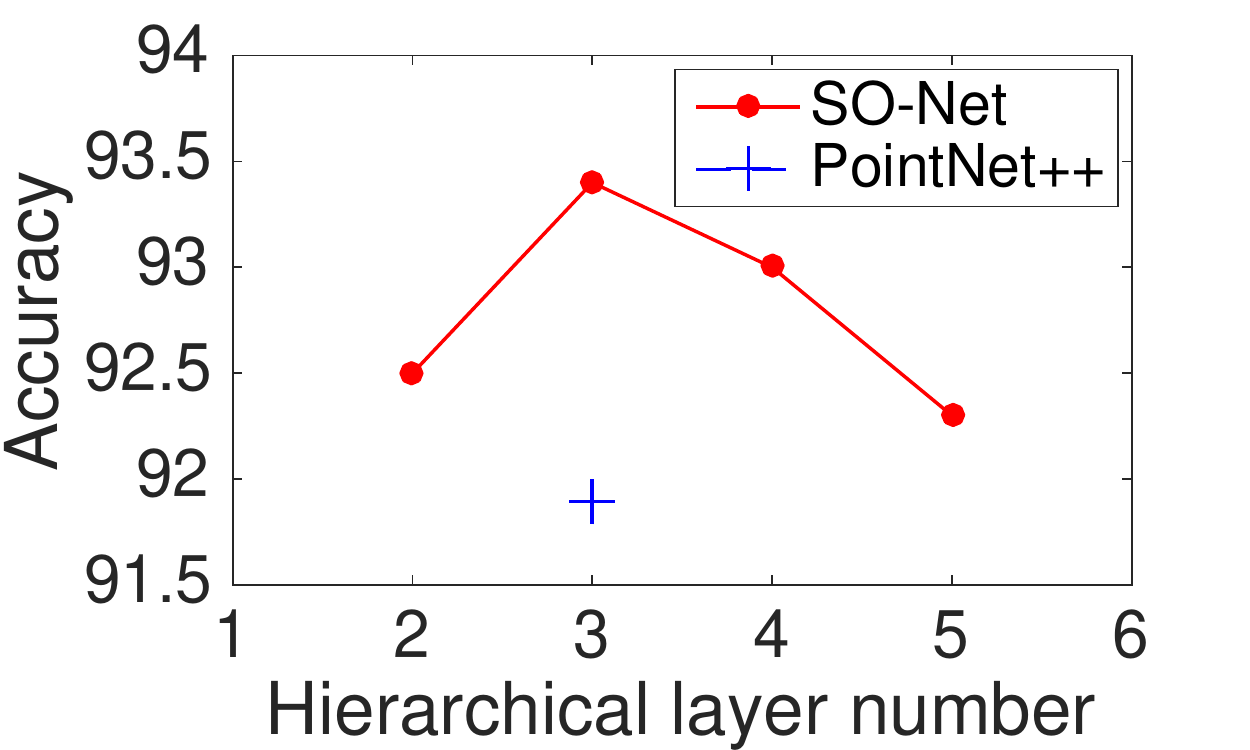}
        \includegraphics[width=0.23\textwidth]{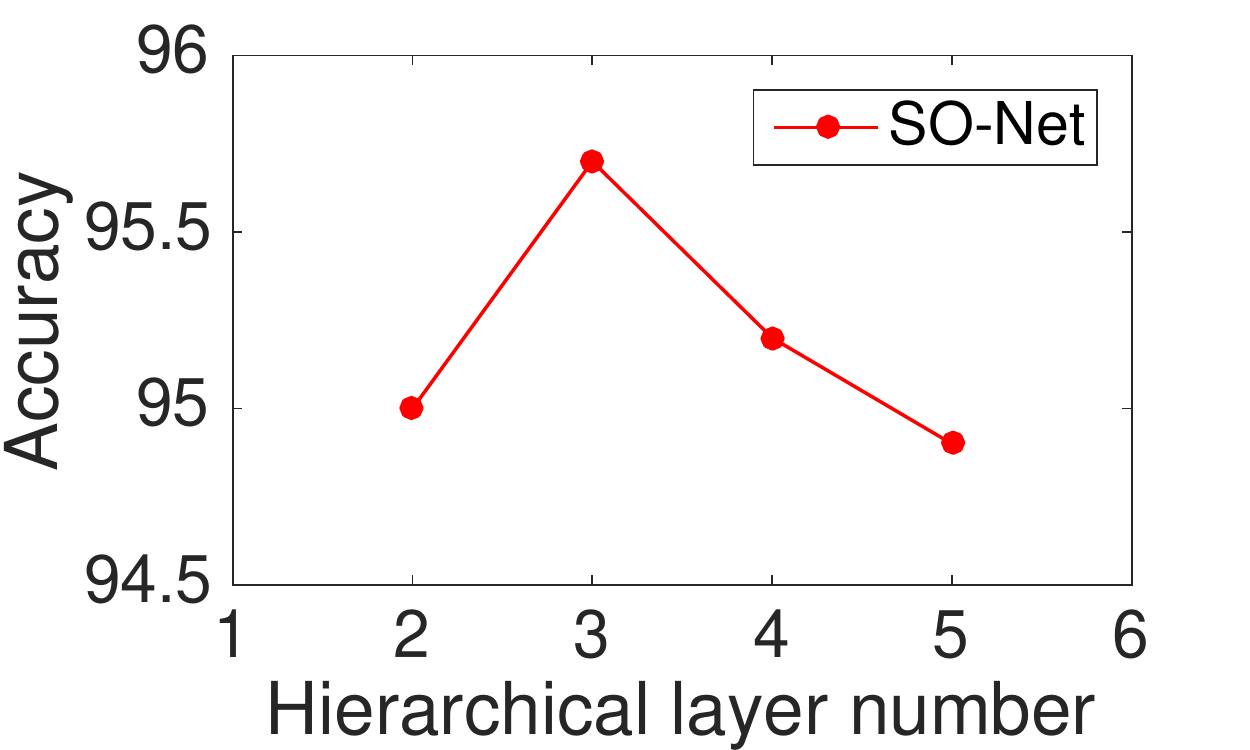}
        \caption{Effect of layer number on classification accuracy with ModelNet40 (left) and ModelNet10 (right).}
        \label{fig_layer_number}
        \vspace{-12pt}
\end{figure}

In this section, we demonstrate that a point cloud can be reconstructed from the SO-Net encoded feature vector, \eg a vector with length of 1024. The nearest neighbor search in Chamfer distance (Eq.~\ref{equ_chamfer}) is conducted with Facebook's faiss \cite{johnson2017billion}. There are two configurations for the decoder to reconstruct different sizes of point clouds. The first configuration generates $64\times64$ points from the convolution branch and 512 points from the fully connected branch. The other one produces $32\times32$ and 256 points respectively, by removing the last upconv module of Fig.~\ref{fig_decoder}.

It is difficult to provide quantitative comparison for the point cloud autoencoder task because little research has been done on this topic. The most related work is the point set generation network \cite{fan2016point} and the point cloud generative models \cite{achlioptas2017learning}. Examples of our reconstructed ShapeNetPart point clouds are visualized in Fig.~\ref{fig_ae_selected}, where 1024 points recovered from the convolution branch are denoted in red and the other 256 points in green. The overall testing Chamfer distance (Eq.~\ref{equ_chamfer}) is 0.033. Similar to the results in \cite{fan2016point}, the convolution branch recovers the main body of the object, while the more flexible fully connected branch focuses on details such as the legs of a table. Nevertheless, many finer details are lost. For example, the reconstructed earphone is blurry. This is probably because the encoder is still not powerful enough to capture fine-grained structures.

Despite the imperfect reconstruction, the autoencoder enhances SO-Net's performance in other tasks by providing a pre-trained encoder, illustrated in Sec.~\ref{sec_exp_cls} and \ref{sec_exp_seg}. More results are visualized in the supplementary materials.

\subsection{Classification Tasks} \label{sec_exp_cls}

To classify the point clouds, we attach a 3-layer multi-layer perceptron (MLP) on top of the encoded global feature vector. Random dropout is applied to the last two layers with keep-ratio of 0.4. Table \ref{tbl_cls} illustrates the classification accuracy for state-of-the-art methods using scalable 3D representations, such as point cloud, kd-tree and octree. In MNIST dataset, our network achieves a relative 13.7\% error rate reduction compared with PointNet++. In ModelNet10 and ModelNet40, our approach out-performs state-of-the-art methods by 1.7\% and 1.5\% respectively in terms of instance accuracy. Our SO-Net even out-performs single networks using multi-view images or uniform voxel grids as input, like qi-MVCNN \cite{qi2016volumetric} (ModelNet40 at 92.0\%) and VRN \cite{brock2016generative} (ModelNet40 at 91.3\%). Methods that integrate multiple networks, i.e., qi-MVCNN-MultiRes \cite{qi2016volumetric} and VRN Ensemble \cite{brock2016generative}, are still better than SO-Net in ModelNet classification, but their multi-view / voxel grid representations are far less scalable and flexible than our point cloud representation, as illustrated in Sec.~\ref{sec_intro} and \ref{sec_related_work}. 


%
\begin{table*}[t]
\centering
{
\setlength\tabcolsep{2.1pt} 
\begin{tabular}{l|lllllllllllllllll}
\hline
\multirow{2}{*}{} & \multicolumn{17}{c}{Intersection over Union (IoU)}                                                                                                 \\ \cline{2-18} 
                  & \multicolumn{1}{l|}{mean} & air  & bag  & cap  & car  & chair & ear. & gui. & knife & lamp & lap. & motor & mug  & pistol & rocket & skate & table \\ \hline
PointNet \cite{qi2016pointnet}          & \multicolumn{1}{l|}{83.7} & 83.4 & 78.7 & 82.5 & 74.9 & 89.6  & 73.0 & 91.5 & 85.9  & 80.8 & 95.3 & 65.2  & 93.0 & 81.2   & 57.9   & 72.8  & 80.6  \\
PointNet++ \cite{qi2017pointnet++}        & \multicolumn{1}{l|}{85.1} & 82.4 & 79.0 & 87.7 & 77.3 & 90.8  & 71.8 & 91.0 & 85.9  & 83.7 & 95.3 & 71.6  & 94.1 & 81.3   & 58.7   & 76.4  & 82.6  \\
Kd-Net \cite{klokov2017escape}            & \multicolumn{1}{l|}{82.3} & 80.1 & 74.6 & 74.3 & 70.3 & 88.6  & 73.5 & 90.2 & 87.2  & 81.0 & 94.9 & 57.4  & 86.7 & 78.1   & 51.8   & 69.9  & 80.3  \\
O-CNN + CRF \cite{wang2017cnn}            & \multicolumn{1}{l|}{\textbf{85.9}} & 85.5 & 87.1 & 84.7 & 77.0 & 91.1  & 85.1 & 91.9 & 87.4  & 83.3 & 95.4 & 56.9  & 96.2 & 81.6   & 53.5   & 74.1  & 84.4  \\ \hline
Ours (pre-trained)             & \multicolumn{1}{l|}{84.9} & 82.8 & 77.8 & 88.0 & 77.3 & 90.6  & 73.5 & 90.7 & 83.9  & 82.8 & 94.8 & 69.1  & 94.2 & 80.9   & 53.1   & 72.9  & 83.0  \\
Ours                   & \multicolumn{1}{l|}{84.6} & 81.9 & 83.5 & 84.8 & 78.1 & 90.8  & 72.2 & 90.1 & 83.6  & 82.3 & 95.2 & 69.3  & 94.2 & 80.0   & 51.6   & 72.1  & 82.6  \\ \hline
\end{tabular}
}
\caption{Object part segmentation results on ShapeNetPart dataset.}
\vspace{-4pt}
\label{tbl_seg}
\end{table*}

\paragraph{Effect of pre-training}
The performance of the network can be improved with pre-training using the autoencoder in Sec.~\ref{sec_autoencoder}. The autoencoder is trained with ModelNet40, using 5000 points and surface normal vectors as input. The autoencoder brings a boost of 0.5\% in ModelNet10 classification, but only 0.2\% in ModelNet40 classification. This is not surprising because pre-training with a much larger dataset may lead to convergence basins \cite{erhan2010does} that are more resistant to over-fitting.

\paragraph{Robustness to point corruption}
We train our network with point clouds of size 2048 but test it with point dropout. As shown in Fig.~\ref{fig_cls_robustness_point}, our accuracy drops by 1.7\% with 50\% points missing (2048 to 1024), and 14.2\% with 75\% points missing (2048 to 512). As a comparison, the accuracy of PN drops by 3.8\% with 50\% points (1024 to 512).

\paragraph{Robustness to SOM corruption}
One of our major concern when designing the SO-Net is whether the SO-Net relies too much on the SOM. With results shown in Fig.~\ref{fig_cls_robustness}, we demonstrate that our SO-Net is quite robust to the noise or corruption of the SOM results. In Fig.~\ref{fig_cls_robustness_som_size}, we train a network with SOM of size $8\times8$ as the noise-free version, but test the network with SOM sizes varying from $5\times5$ to $11\times11$. It is interesting that the performance decay is much slower if the SOM size is larger than training configuration, which is consistent with the theory in Sec.~\ref{sec_encoder}. The SO-Net separates the input point cloud into $M$ mini point clouds, encodes them into $M$ node features with a mini PointNet, and assembles them during the global feature extraction. In the case that the SOM becomes smaller during testing, the mini point clouds are too large for the mini PointNet to encode. Therefore the network performs worse when the testing SOM is smaller than expected.

In Fig.~\ref{fig_cls_robustness_som_noise}, we add Gaussian noise $\mathcal{N}(0, \sigma)$ onto the SOM during testing. Given the fact that input points have been normalized into a unit cube, a Gaussian noise with $\sigma=0.2$ is rather considerable, as shown in Fig.~\ref{fig_cls_robustness_som_noisy_example}. Even in that difficult case, our network achieves the accuracy of 91.1\% in ModelNet40 and 94.6\% in ModelNet10. 

\paragraph{Effect of hierarchical layer number}
Our framework shown in Fig.~\ref{fig_architecture} can be made to further out-perform state-of-the-art methods by simply adding more layers. The vanilla SO-Net is a 2-layer structure ``grouping\&PN(PointNet) - PN'', where the grouping is based on SOM and point-to-node $k$NN. We make it a 3-layer structure by simply repeating the SOM/$k$NN based ``grouping\&PN'' with this protocol: for each SOM node, find $k'=9$ nearest nodes and process the $k'$ node features with a PointNet. The output is a new SOM feature map of the same size but larger receptive field. Shown in Table~\ref{tbl_cls}, our 3-layer SO-Net increases the accuracy to 1.5\% higher (relatively 19\% lower error rate) than PN++ on ModelNet40, and 1.7\% higher (relatively 28\% lower error rate) than Kd-Net on ModelNet10. The effect of hierarchical layer number is illustrated in Fig.~\ref{fig_layer_number}, where too many layers may lead to over-fitting.

\paragraph{Training speed}
The batch training of SOM allows parallel implementation on GPU. Moreover, the training of SOM is completely deterministic in our approach, so it can be isolated as data preprocessing before network optimization. Compared to the randomized kd-tree construction in \cite{klokov2017escape}, our deterministic design provides great boosting during training. In addition to the decoupled SOM, the hierarchical feature aggregation based on SOM can be implemented efficiently on GPU. As shown in Table \ref{tbl_cls}, it takes about 3 hours to train our best network on ModelNet40 with a GTX1080Ti, which is significantly faster than state-of-the-art networks that can provide comparable performance.

\subsection{Part Segmentation on ShapeNetPart} \label{sec_exp_seg}
%
%
\begin{figure}[h] 
        \centering
        \includegraphics[width=0.15\textwidth]{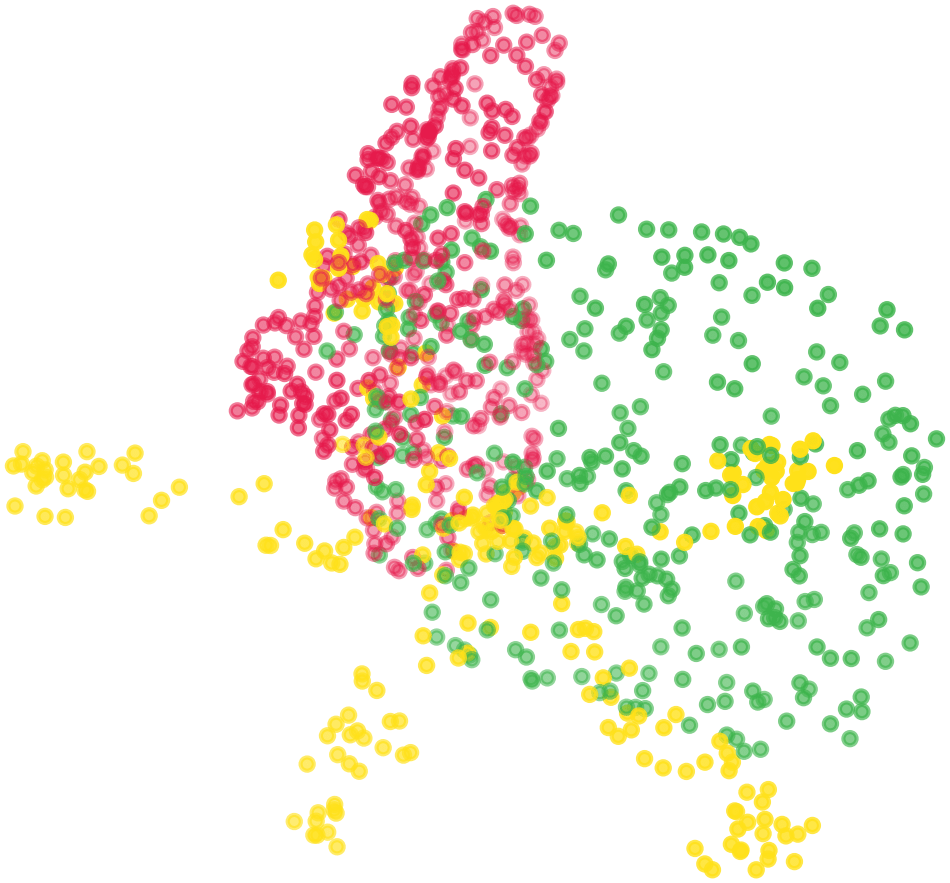}
        \includegraphics[width=0.15\textwidth]{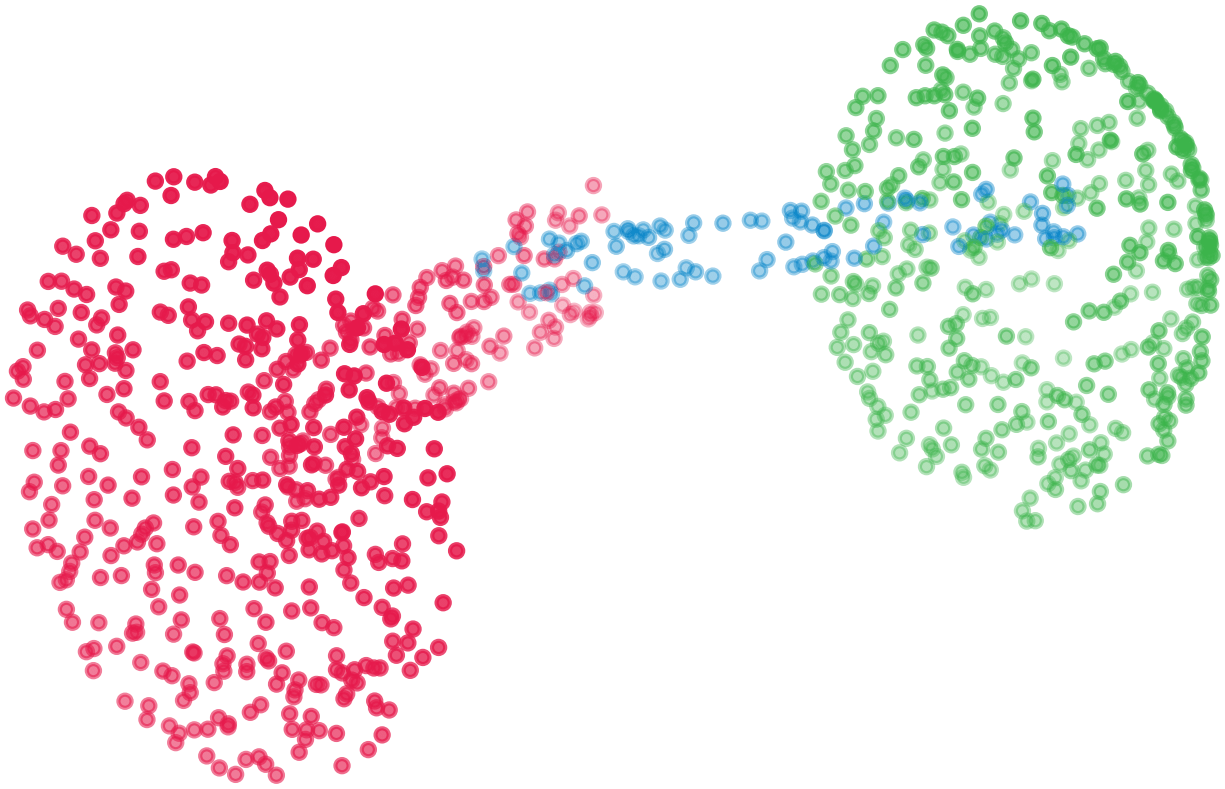}
        \includegraphics[width=0.15\textwidth]{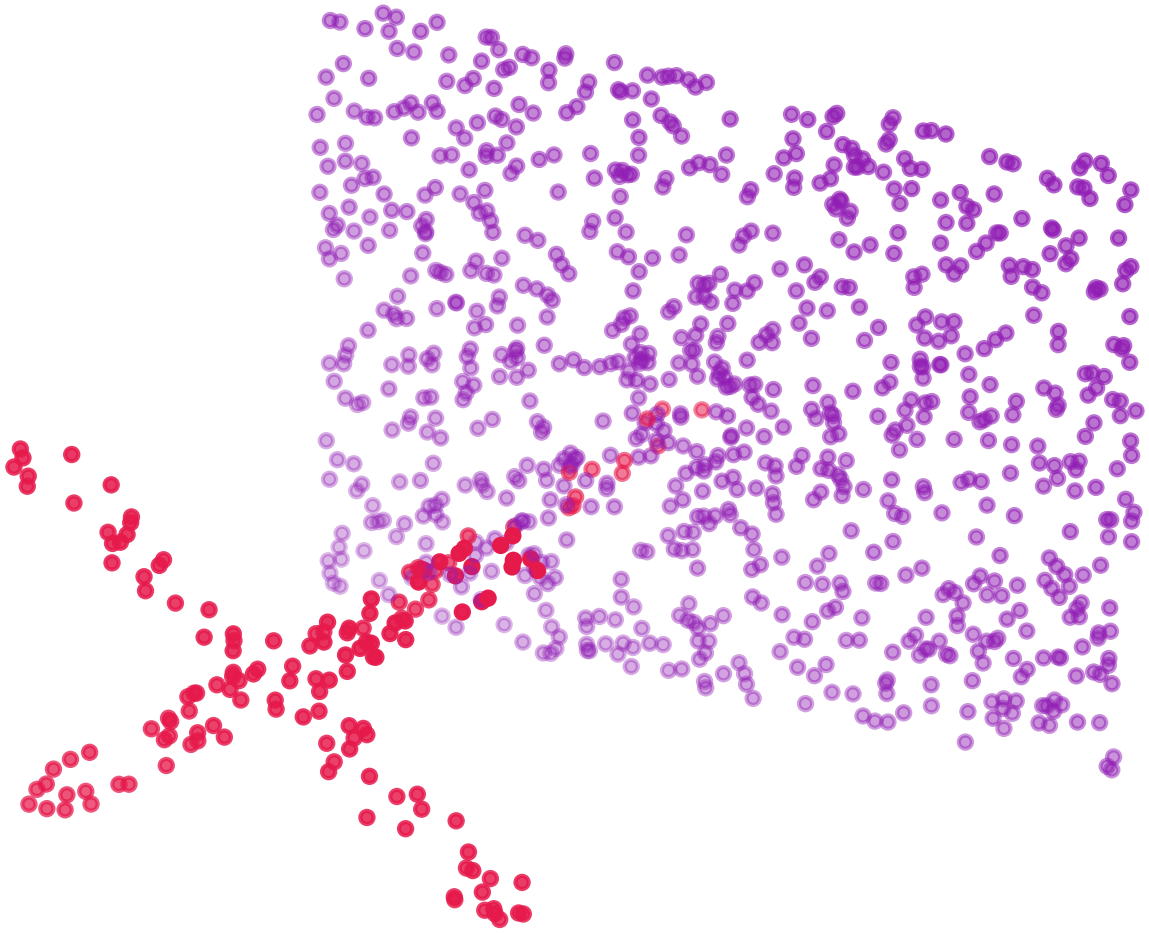}
        \includegraphics[width=0.15\textwidth]{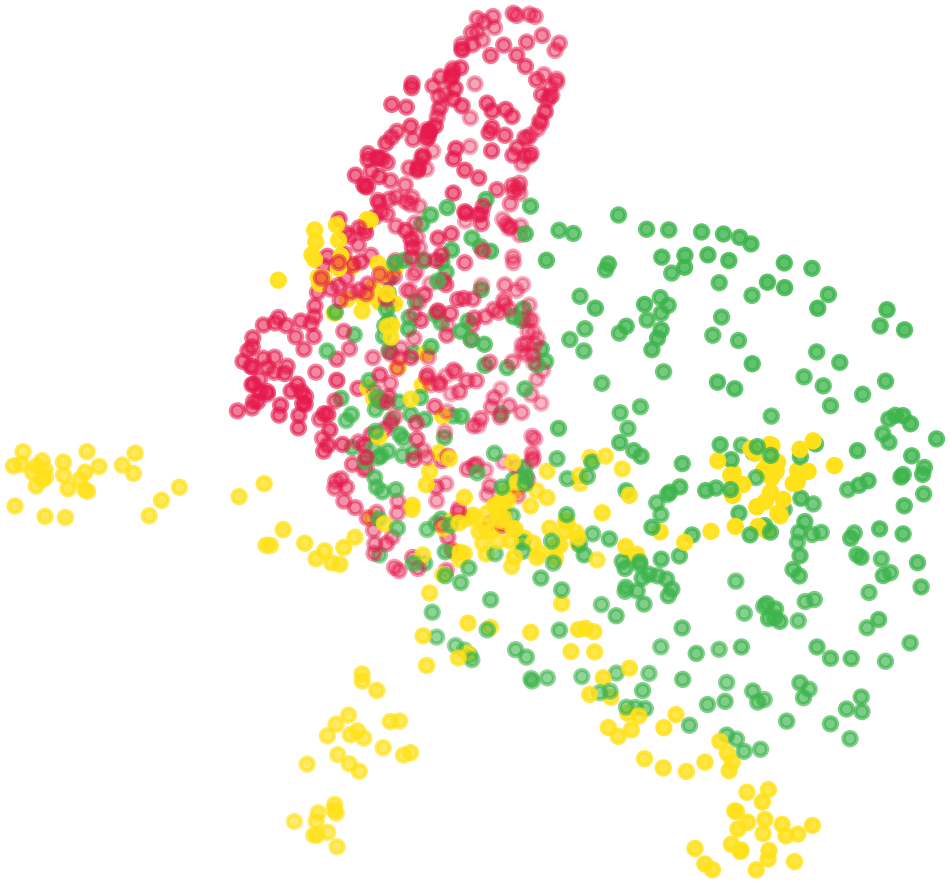}
        \includegraphics[width=0.15\textwidth]{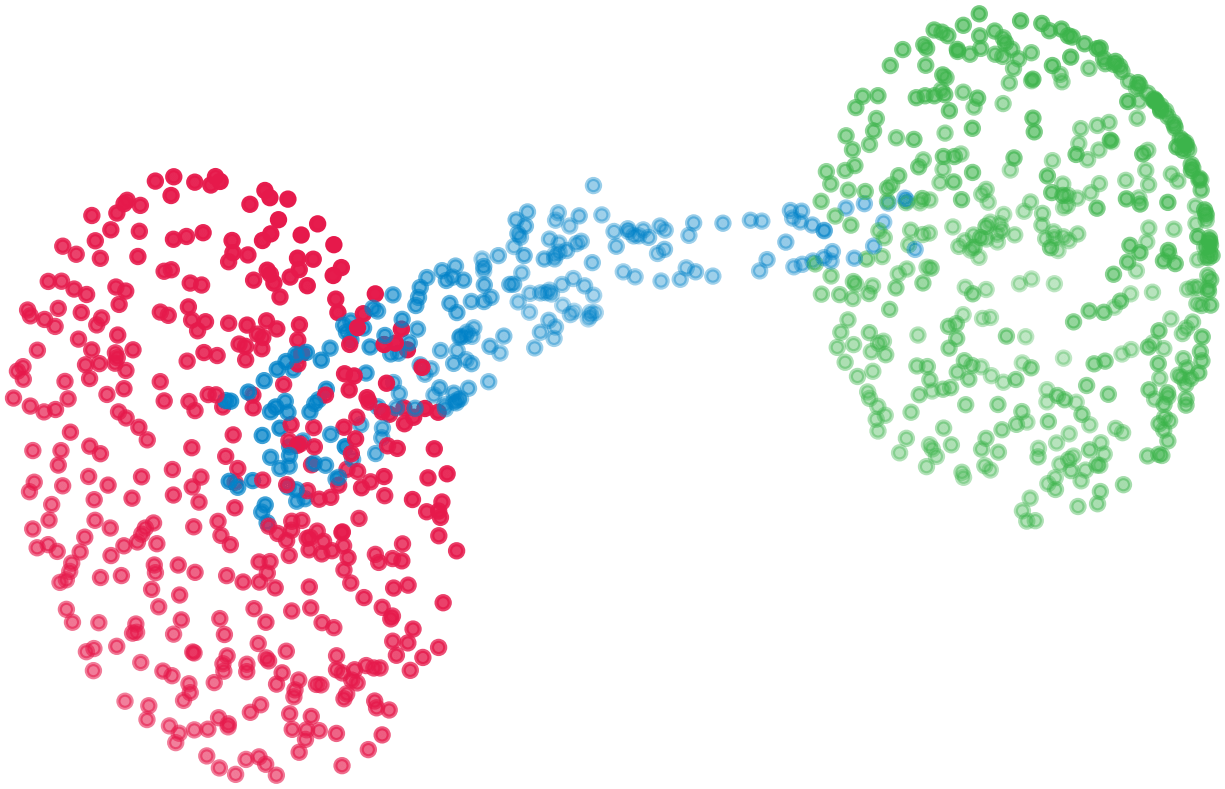}
        \includegraphics[width=0.15\textwidth]{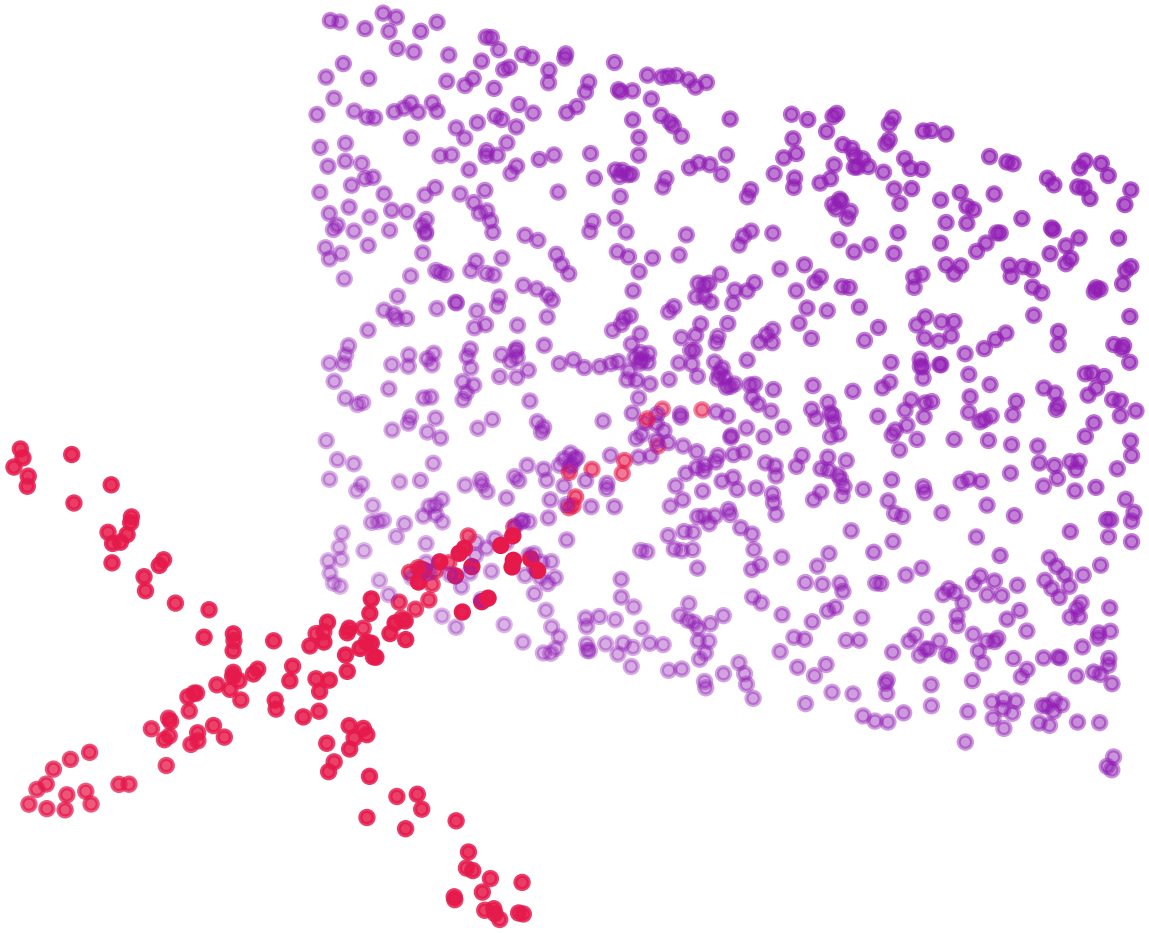}
        \caption{Visualization of object part segmentation results. First row: ground truth. Second row: predicted segmentation. From left to right: chair, lamp, table. }
        \label{fig_seg_selected}
        \vspace{-4pt}
\end{figure}
We formulate the object part segmentation problem as a per-point classification task, as illustrated in Fig.~\ref{fig_architecture}. The networks are evaluated using the mean Intersection over Union (IoU) protocol proposed in \cite{qi2016pointnet}. For each instance, IoU is computed for each part that belongs to that object category. The mean of the part IoUs is regarded as the IoU for that instance. Overall IoU is calculated as the mean of IoUs over all instances, and category-wise IoU is computed as an average over instances under that category. Similar with O-CNN \cite{wang2017cnn} and PointNet++ \cite{qi2017pointnet++}, surface normal vectors are fed into the network together with point coordinates.

By optimizing per-point softmax loss functions, we achieve competitive results as reported in Table \ref{tbl_seg}. Although O-CNN reports the best IoU, it adopts an additional dense conditional random field (CRF) to refine the output of their network while others do not contain this post-processing step. Some segmentation results are visualized in Fig.~\ref{fig_seg_selected} and we further visualize one instance per category in the supplementary material. Although in some hard cases our network may fail to annotate the fine-grained details correctly, generally our segmentation results are visually satisfying. The low computation cost remains as one of our advantages. Additionally, pre-training with our autoencoder produces a performance boost, which is consistent with our classification results.

\section{Conclusion} \label{sec_conclusion}
In this paper, we propose the novel SO-Net that performs hierarchical feature extraction for point clouds by explicitly modeling the spatial distribution of input points and systematically adjusting the receptive field overlap. In a series of experiments including point cloud reconstruction, object classification and object part segmentation, our network achieves competitive performance. In particular, we out-perform state-of-the-art deep learning approaches in point cloud classification and shape retrieval, with significantly faster training speed. As the SOM preserves the topological properties of the input space and our SO-Net converts point clouds into feature matrice accordingly, one promising future direction is to apply classical ConvNets or graph-based ConvNets to realize deeper hierarchical feature aggregation.

\vspace{-8pt}
\paragraph{Acknowledgment.} \label{sec_ack}
This work is supported partially by a ODPRT start-up grant R-252-000-636-133 from the National University of Singapore.

%


{\small
\bibliographystyle{ieee}
\bibliography{CVPR2018_REF}
}




\newpage
\section*{Supplementary}

\appendix

\section{Overview}
This supplementary document provides more technical details and experimental results to the main paper. Shape retrieval experiments are demonstrated with ShapeNet Core55 dataset in Sec.~\ref{sec_shape_retrieval}. The time and space complexity is analyzed in Sec.~\ref{sec_time_space}, followed by detailed illustration of our permutation invariant SOM training algorithms in Sec.~\ref{sec_supplementary_som}. More experiments and results are presented in Sec.~\ref{sec_more_exps}.

\section{Shape Retrieval} \label{sec_shape_retrieval}
Our object classification network can be easily extended to the task of 3D shape retrieval by regarding the classification score as the feature vector. Given a query shape and a shape library, the similarity between the query and the candidates can be computed as their feature vector distances.

\subsection{Dataset}
We perform 3D shape retrieval task using the ShapeNet Core55 dataset, which contains 51,190 shapes from 55 categories and 204 subcategories. Specifically, we adopt the dataset split provided by the 3D Shape Retrieval Contest 2016 (SHREC16), where 70\% of the models are used for training, 10\% for validation and 20\% for testing. Since the 3D shapes are represented by CAD models, i.e., vertices and faces, we sample 5,000 points and surface normal vectors from each CAD model. Data augmentation is identical with the previous classification and segmentation experiments - random jitter and scale.

\subsection{Procedures}
We train a classification network on the ShapeNet Core55 dataset using identical configurations as our ModelNet40 classification experiment, i.e. a SOM of size $8\times 8$ and $k=3$. For simplicity, the softmax loss is minimized with only the category labels (without any subcategory information). The classification score vector of length 55 is used as the feature vector. We calculate the L2 feature distance between 
each shape in the test set 
and all shapes in the same predicted category from the test set (including itself). The corresponding retrieval list is constructed by sorting these shapes according to the feature distances.

\subsection{Performance}
\begin{figure*}[th]
        \centering
        \subfigure[]{\includegraphics[width=0.45\textwidth]{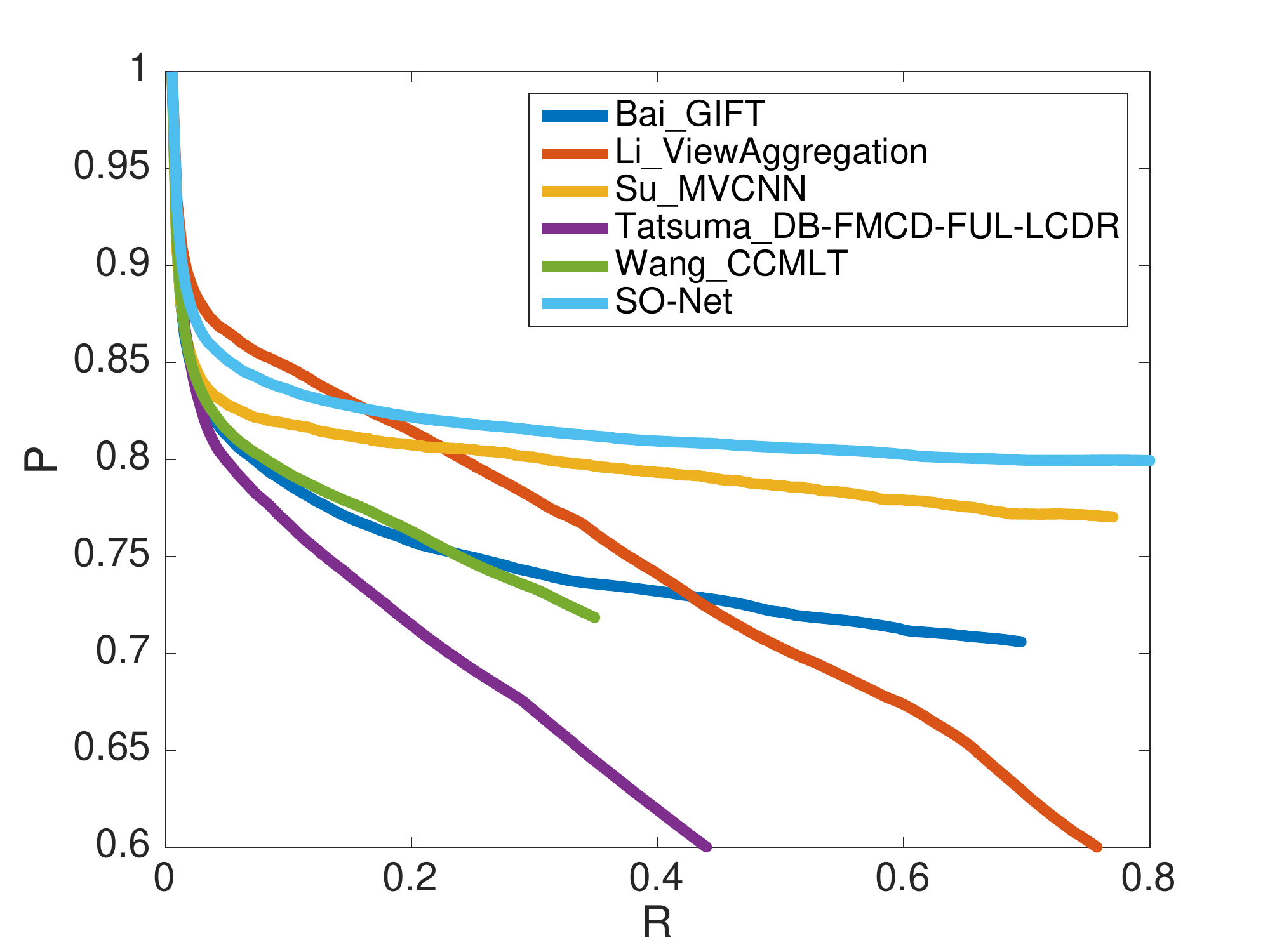}}
        \subfigure[]{\includegraphics[width=0.45\textwidth]{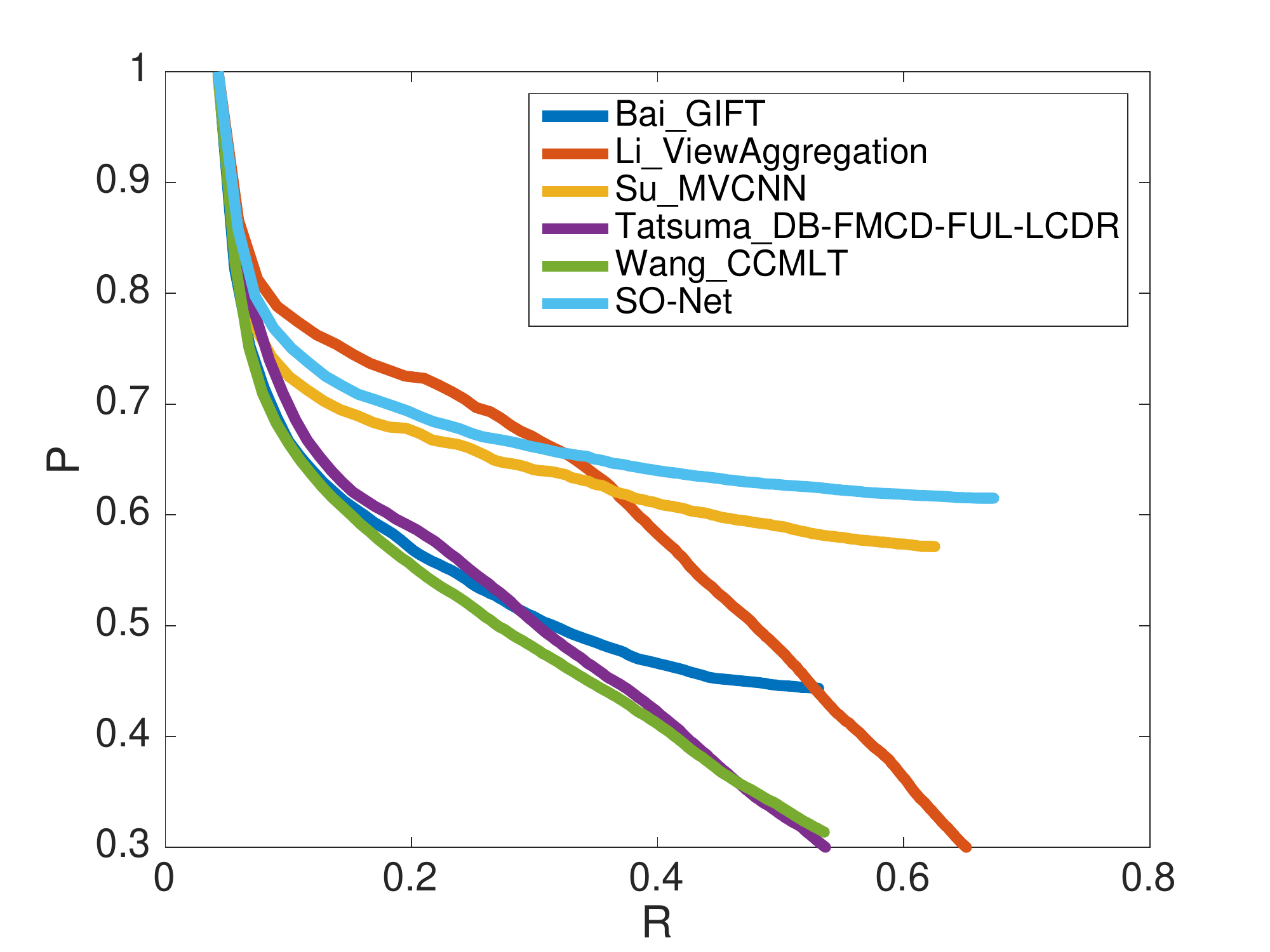}}
        \caption{Precision-recall curve for micro (a) and macro (b) metrics in the 3D shape retrieval task. In both curves, the SO-Net demonstrates the largest AUC.}
        \label{fig_shrec16_pr}
        \vspace{-4pt}
\end{figure*}

SHREC16 provides several evaluation metrics including Precision-Recall curve, F-score, mean average precision (mAP), normalized discounted cumulative gain (NDCG). These metrics are computed under two contexts - macro and micro. Macro metric is a simple average across all categories while micro metric is a weighted average according to the number of shapes in each category. As shown in Table \ref{tbl_shrec16_metric}, our SO-Net out-performs state-of-the-art approaches with most metrics. The precision-recall curves are illustrated in Fig.~\ref{fig_shrec16_pr}, where SO-Net demonstrates the largest area under curve (AUC).
Some shape retrieval results are visualized in Fig.~\ref{fig_shrec16_visualization}.

\begin{table*}[t]
\centering
{
\setlength\tabcolsep{5pt} 
\begin{tabular}{l|ccccc|ccccc}
\hline
\multirow{2}{*}{Method}   & \multicolumn{5}{c|}{Micro}                                        & \multicolumn{5}{c}{Macro}             \\
                          & P@N   & R@N   & F1@N  & mAP   & NDCG@N                            & P@N   & R@N   & F1@N  & mAP   & NDCG@N \\ \hline
Tatsuma                   & 0.427 & 0.689 & 0.472 & 0.728 & 0.875                             & 0.154 & 0.730 & 0.203 & 0.596 & 0.806  \\
Wang\_CCMLT               & 0.718 & 0.350 & 0.391 & 0.823 & 0.886                             & 0.313 & 0.536 & 0.286 & 0.661 & 0.820  \\
Li\_ViewAggregation       & 0.508 & \textbf{0.868} & 0.582 & 0.829 & 0.904                    & 0.147 & \textbf{0.813} & 0.201 & 0.711 & 0.846  \\
Bai\_GIFT \cite{bai2016gift} & 0.706 & 0.695 & 0.689 & 0.825 & 0.896                             & 0.444 & 0.531 & 0.454 & 0.740 & 0.850  \\
Su\_MVCNN \cite{su2015multi} & 0.770 & 0.770 & 0.764 & 0.873 & 0.899                          & 0.571 & 0.625 & 0.575 & \textbf{0.817} & 0.880  \\
Kd-Net \cite{klokov2017escape} & 0.760 & 0.768 & 0.743 & 0.850 & 0.905                        & 0.492 & 0.676 & 0.519 & 0.746 & 0.864  \\
O-CNN \cite{wang2017cnn}     & 0.778 & 0.782 & 0.776 & \textbf{0.875} & 0.905                 & -     & -     & -     & -     & -      \\ \hline
Ours                      & \textbf{0.799} & 0.800 & \textbf{0.795} & 0.869 & \textbf{0.907}  & \textbf{0.615} & 0.673 & \textbf{0.622} & 0.805 & \textbf{0.888}  \\ \hline
\end{tabular}
}
\caption{3D shape retrieval results with SHREC16. Our SO-Net out-performs state-of-the-art deep networks with most metrics.}
\label{tbl_shrec16_metric}
\end{table*}

\section{Time and Space Complexity} \label{sec_time_space}
We evaluate the model size, forward (inference) time and training time of several point cloud based networks in the task of ModelNet40 classification, as shown in Table~\ref{tbl_time_space}. The forward timings are acquired with a batch size of 8 and input point cloud size of 1024. In the comparison, we choose the networks with the best classification accuracy among various configurations of PointNet and PointNet++, i.e., PointNet with transformations and PointNet++ with multi-scale grouping (MSG). Because of the parallelizability and simplicity of our network design, our model size is smaller and the training speed is significantly faster compared to PointNet and its successor PointNet++. Meanwhile, our inference time is around 1/3 of that of PointNet++.

\begin{table}[t!]
\centering
{
\setlength\tabcolsep{6pt} 
\begin{tabular}{l|ccc}
\hline
                                    & Size / MB       & Forward / ms      & Train / h                    \\ \hline
PointNet \cite{qi2016pointnet}      & 40            & \textbf{25.3}   & 3-6                        \\
PointNet++ \cite{qi2017pointnet++}  & 12            & 163.2           & 20                         \\
Kd-Net \cite{klokov2017escape}      & -             & -               & 16                         \\ \hline
Ours                                & \textbf{11.5} & 59.6            & \textbf{1.5}                 \\ \hline
\end{tabular}
}
\caption{Time and space complexity of point cloud based networks in ModelNet40 classification. }
\label{tbl_time_space}
\vspace{-4pt}
\end{table}

\section{Permutation Invariant SOM} \label{sec_supplementary_som}
We apply two methods to ensure that the SOM is invariant to the permutation of the input points - fixed initialization and deterministic training rule.

\subsection{Initialization for SOM Training}
\begin{figure}[t]
        \centering
        \includegraphics[width=0.23\textwidth]{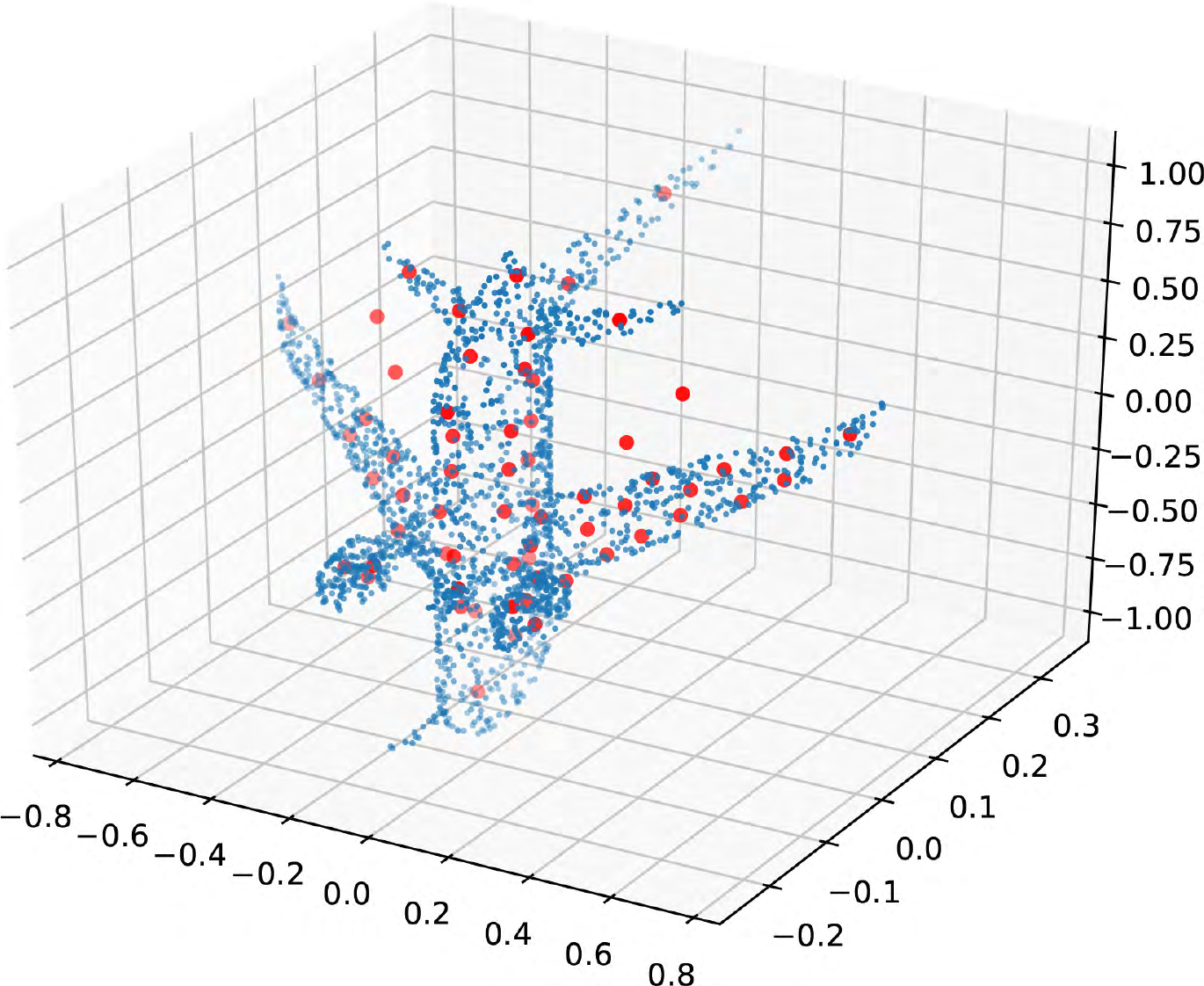}
        \includegraphics[width=0.23\textwidth]{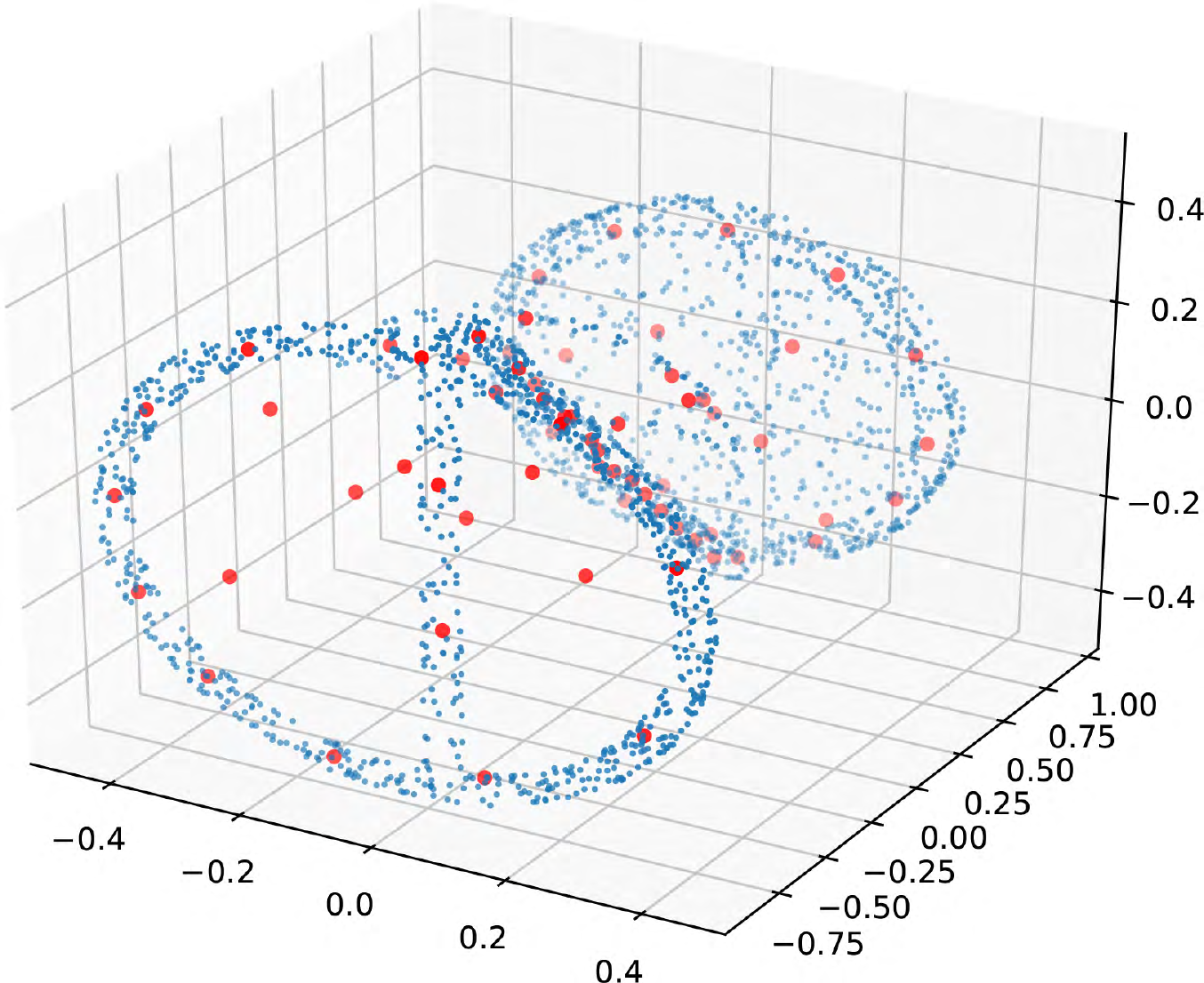}
        \caption{ Results of SOM training with uniform initialization. Isolated nodes are inevitable even with uniform initialization.} \label{fig_som_results}
        \vspace{-4pt}
\end{figure}

In addition to permutation invariance, the initialization should be reasonable so that the SOM training is less prone to local minima. Suboptimal SOM may lead to many isolated nodes outside the coverage of the input points. 
For simplicity, we use fixed initialization for any point cloud input
although there are other initialization approaches that are permutation invariant, \eg, principal component initialization. We generate a set of node coordinates that are uniformly distributed in an unit ball to serve as a reasonable initialization because the input point clouds are in various shapes. Unfortunately,
as shown in Fig.~\ref{fig_som_results},
isolated nodes are inevitable even with uniform initialization. 
Isolated nodes may not be associated during the kNN search, and their corresponding node features will be set to zero, i.e. the node features are invalid. Nevertheless, our SO-Net is robust to small amount of invalid nodes as demonstrated in the experiments.

We propose a simple algorithm based on potential field methods to generate the initialization as shown in Algorithm \ref{alg_potential_field}. $S=\{s_j\in\mathbb{R}^3, j=0, \cdots, M-1\}$ represents the SOM nodes and $\eta$ is the learning rate. The key idea is to apply a repulsion force between any pair of nodes, and external forces to attract nodes toward the origin. The parameter $\lambda$ is used to control the weighting between the repulsion and attraction force, so that the resulting nodes are within the unit ball.
\begin{algorithm}
\caption{Potential field method for SOM initialization} \label{alg_potential_field}
\begin{algorithmic}
\State Set random seed.
\State Random initialization: $S \gets \mathcal{U}(-1,1)$
\Repeat
\ForAll{$s_j \in S $}
    \State $f_j^{wall} \gets -s_j$
    \State $f_j^{node} \gets 0$
    \ForAll{$s_k \in S, k\neq j $}
        \State $f_j^{node} \gets f_j^{node} + \lambda \frac{s_j-s_k}{\|s_j-s_k\|^2_2}$
    \EndFor
\EndFor
\ForAll{$s_j \in S $}
    \State $s_j \gets s_j + \eta (f_j^{wall} + f_j^{node})$
\EndFor
\Until{converge}
\end{algorithmic}
\end{algorithm}

\subsection{Batch Update Training}
Instead of updating the SOM once per point, the batch update rule conducts one update after accumulating the effect of all points in the point cloud. As a result, each SOM update iteration is unrelated to the order of point, i.e., permutation invariant. During SOM training, each training sample affects the winner node and all its neighbors. We define the neighborhood function as a Gaussian distribution as follows:

\begin{equation} \label{equ_joint_gaussian}
\begin{split}
    w_{xy}(x,y | p,q, \sigma_x, \sigma_y) &= \frac{\text{exp}\big(-\frac{1}{2}(x-\mu)^T \Sigma^{-1} (x-\mu)\big)}{\sqrt{(2\pi)^2|\Sigma|}}  \\
    \mu &= \begin{bmatrix}p & q\end{bmatrix}^T \\
    \Sigma &= \begin{bmatrix} \sigma_x^2 & 0 \\ 0 & \sigma_y^2  \end{bmatrix}.
\end{split}
\end{equation}

The pseudo code of the training scheme is shown in Algorithm \ref{alg_som_training}. $P=\{p_i\in\mathbb{R}^3, i=0, \cdots,N-1\}$ and $S=\{s_j\in\mathbb{R}^3, j=0, \cdots, M-1\}$ represent the input points and SOM nodes respectively. The learning rate $\eta_t$ and neighborhood parameter $(\sigma_x, \sigma_y)$ should be decreased slowly during training. In addition, Algorithm \ref{alg_som_training} can be easily implemented as matrix operations which are highly efficient on GPU.

\begin{algorithm}
\caption{SOM batch update rule} \label{alg_som_training}
\begin{algorithmic}
\State Initialize $m\times m $ SOM $S$ with Algorithm \ref{alg_potential_field}
\For{$t < \text{MaxIter}$}
    \State \Comment{Set update vectors to zero}
    \ForAll{ $s_{xy} \in S$ }
        \State $D_{xy} \gets 0$
    \EndFor
    
    \State \Comment{Accumulate the effect of all points}
    \ForAll{ $p_i \in P$ }
        \State Obtain nearest neighbor coordinate $p, q$
        \ForAll{ $s_{xy} \in S$ }
            \State $w_{xy} \gets $ Eq.~(\ref{equ_joint_gaussian})
            \State $D_{xy} \gets D_{xy} + w_{xy}(p_i - s_{xy})$
        \EndFor
    \EndFor
    
    \State \Comment{Conduct one update}
    \ForAll{ $s_{xy} \in S$ }
        \State $s_{xy} \gets s_{xy} + \eta_t D_{xy}$
    \EndFor
    \State $t \gets t+1$
    \State Adjust $\sigma_x$, $\sigma_y$ and $\eta_t$
\EndFor
\end{algorithmic}
\end{algorithm}

\section{More Experiments} \label{sec_more_exps}

\begin{table*}[th]
\centering
\begin{tabular}{ll|ccc}
\hline
Method                                            & Input     & MNIST & ModelNet10 & ModelNet40 \\ \hline
Kd-Net split based MLP \cite{klokov2017escape}    & splits    & 82.40  & 83.4       & 73.2       \\
Kd-Net depth 10 \cite{klokov2017escape}           & point     & 99.10  & 93.3       & 90.6       \\ \hline
Ours - SOM based MLP                              & SOM nodes & 91.37  & 88.9       & 75.7       \\
Ours                                              & point     & 99.56  & 94.5       & 92.3       \\ \hline
\end{tabular}
\caption{Classification results using structure information - SOM nodes and kd-tree split directions. } \label{tbl_som_cls}
\end{table*}

\subsection{MNIST Classification} 
We evaluate our network using the 2D MNIST dataset, which contains 60,000 $28\times 28$ images for training and 10,000 images for testing. 2D coordinates are extracted from the non-zero pixels of the images. In order to upsample these 2D coordinates into point clouds of a certain size, \eg, 512 in our experiment, we augment the original pixel coordinates with Gaussian noise $\mathcal{N}(0, 0.01)$. Other than the acquisition of point clouds, the data augmentation is exactly the same as other experiments using ModelNet or ShapeNetPart. We reduce the SOM size to $4\times 4$ and set $k=4$ because the point clouds are in 2D and the cloud size is relatively small. The neurons in the shared fully connected layers are reduced as well: 2-64-64-128-128 during point feature extraction and (128+2)-256-512-512-1024 during node feature extraction.

Similar to 3D classification tasks, our network out-performs existing point cloud based deep networks although the best performance is still from the well engineered 2D ConvNets as shown in Table \ref{tbl_mnist}. Despite using point cloud representation instead of images, our network demonstrates better results compared with ConvNets such as Network in Network \cite{lin2013network}, LeNet5 \cite{lecun1998gradient}.

\begin{table}[h]
\centering
\begin{tabular}{lc}
\hline
Method                 & Error rate (\%) \\ \hline
Multi-column DNN \cite{ciregan2012multi}       & \textbf{0.23}    \\
Network in Network \cite{lin2013network}     & 0.47    \\
LeNet5 \cite{lecun1998gradient}                 & 0.80    \\
Multi-layer perceptron \cite{simard2003best} & 1.60    \\ 
\hline
PointNet \cite{qi2016pointnet}               & 0.78    \\
PointNet++ \cite{qi2017pointnet++}             & 0.51    \\
Kd-Net \cite{klokov2017escape}                 & 0.90    \\
ECC \cite{simonovsky2017dynamic}                    & 0.63    \\ 
\hline
Ours                   & 0.44    \\ 
\hline
\end{tabular}
\caption{MNIST classification results.}
\label{tbl_mnist}
\end{table}

\subsection{Classification with SOM Only} 
There are two sources of information utilized by the SO-Net - the point cloud and trained SOM. The information from SOM is explicitly used when the nodes are concatenated with the node features at the beginning of node feature extraction. Additionally, the SOM is implicitly utilized because point normalization, kNN search and the max pooling are based on the nodes.
We perform classification using the SOM nodes without the point coordinates of the point cloud to analyze the contribution of the SOM. 
We feed the SOM nodes into a 3-layer MLP with MNIST, ModelNet10 and ModelNet40 dataset. Similarly in the Kd-Net \cite{klokov2017escape}, experiments are conducted using the kd-tree split directions without point information, i.e. feeding directions of the splits into a MLP. The results are shown in Table \ref{tbl_som_cls}.

It is interesting that we can achieve reasonable performance in the classification tasks by
combining SOM and a simple MLP. But there is still a large gap between this variant and the full SO-Net, which suggests that the integration of SOM and point clouds is important. Another intriguing phenomenon is that the SOM based MLP achieves better results than split-based MLP. It suggests that maybe SOM is more expressive than kd-trees in the context of classification.

\subsection{Result Visualization}
\begin{figure*}[h] 
        \centering
        \includegraphics[width=0.15\textwidth]{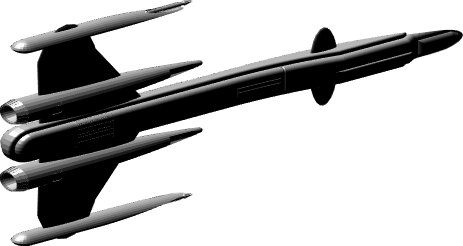}
        \includegraphics[width=0.15\textwidth]{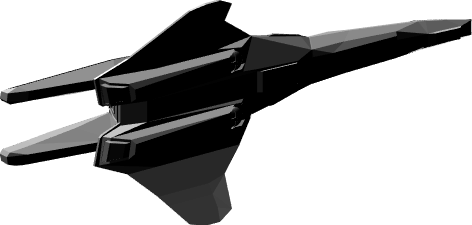}
        \includegraphics[width=0.15\textwidth]{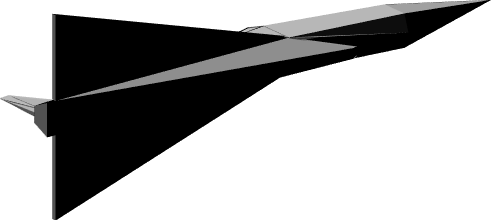}
        \includegraphics[width=0.15\textwidth]{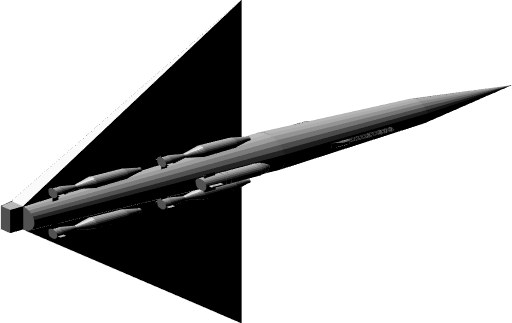}
        \includegraphics[width=0.15\textwidth]{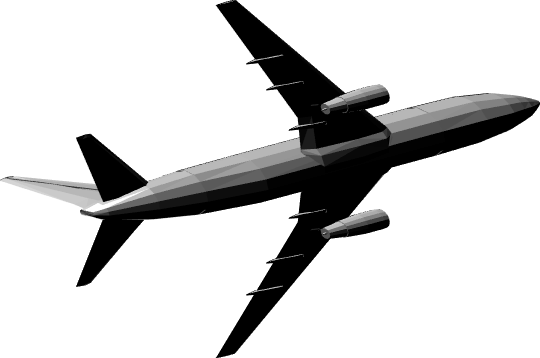}
        \includegraphics[width=0.15\textwidth]{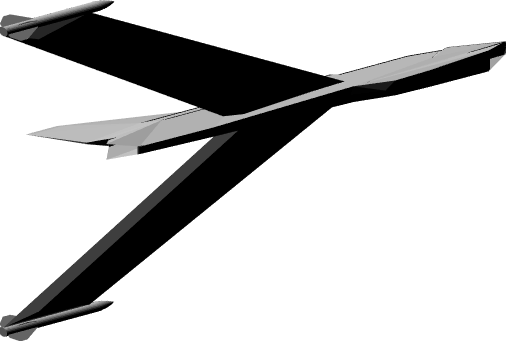}
        
        \includegraphics[width=0.15\textwidth]{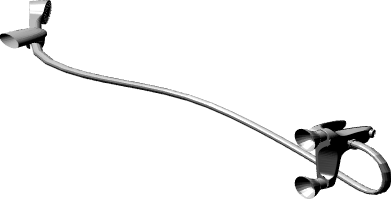}
        \includegraphics[width=0.15\textwidth]{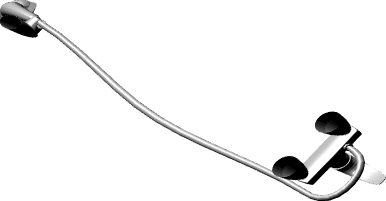}
        \includegraphics[width=0.15\textwidth]{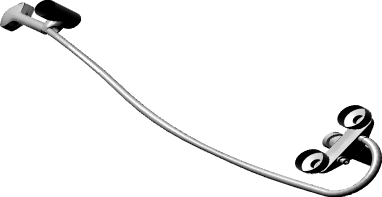}
        \includegraphics[width=0.15\textwidth]{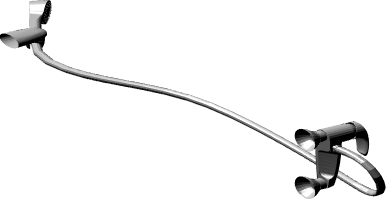}
        \includegraphics[width=0.15\textwidth]{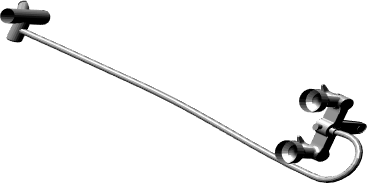}
        \includegraphics[width=0.15\textwidth]{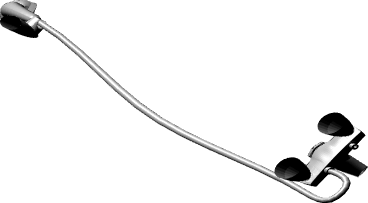}
        
        \includegraphics[width=0.15\textwidth]{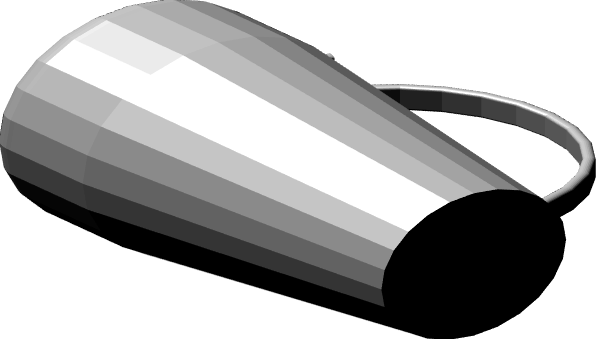}
        \includegraphics[width=0.15\textwidth]{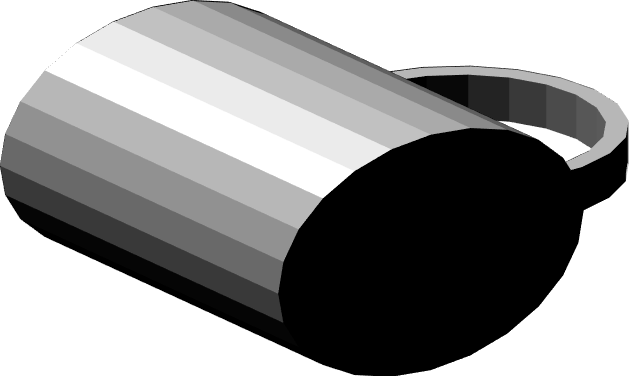}
        \includegraphics[width=0.15\textwidth]{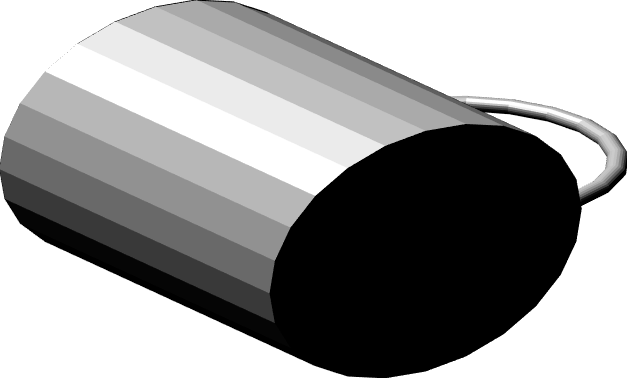}
        \includegraphics[width=0.15\textwidth]{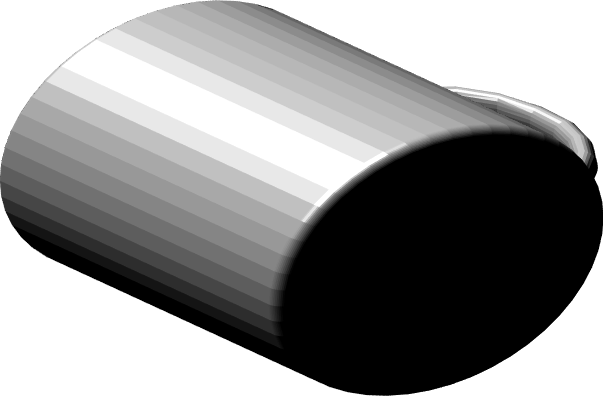}
        \includegraphics[width=0.15\textwidth]{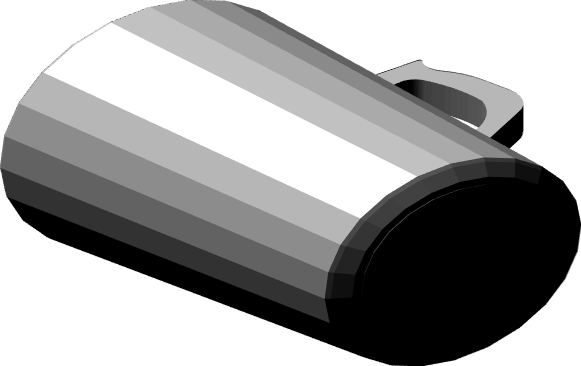}
        \includegraphics[width=0.15\textwidth]{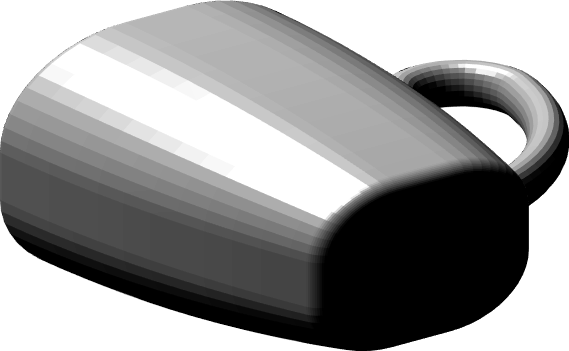}
        
        \includegraphics[width=0.15\textwidth]{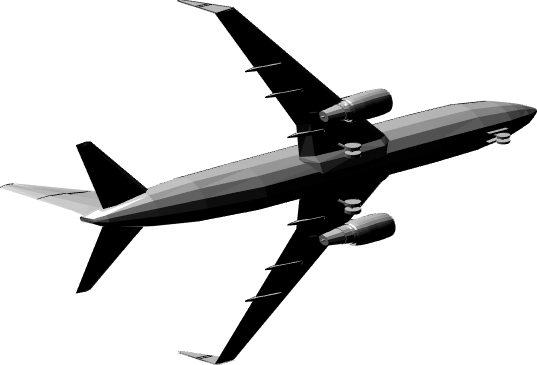}
        \includegraphics[width=0.15\textwidth]{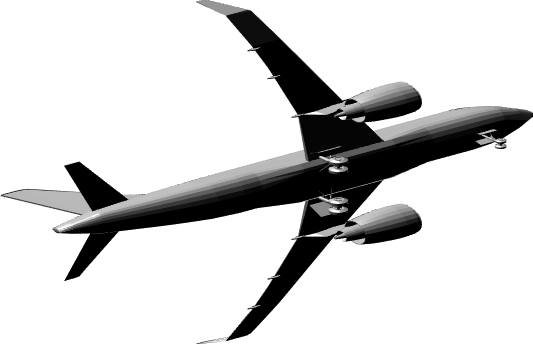}
        \includegraphics[width=0.15\textwidth]{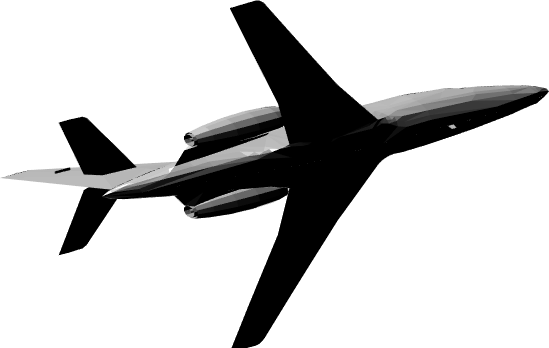}
        \includegraphics[width=0.15\textwidth]{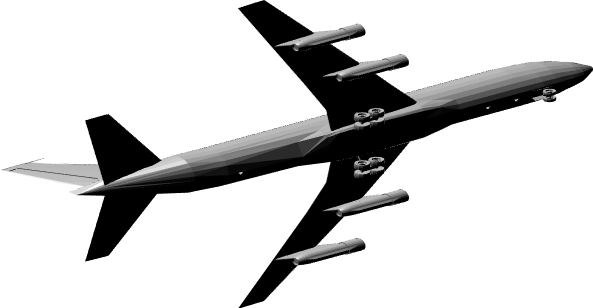}
        \includegraphics[width=0.15\textwidth]{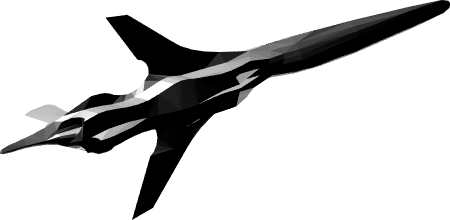}
        \includegraphics[width=0.15\textwidth]{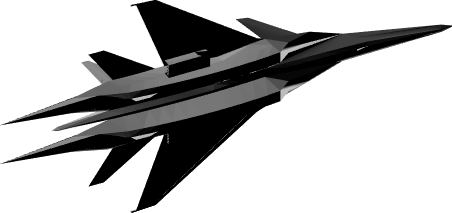}
        
        \includegraphics[width=0.15\textwidth]{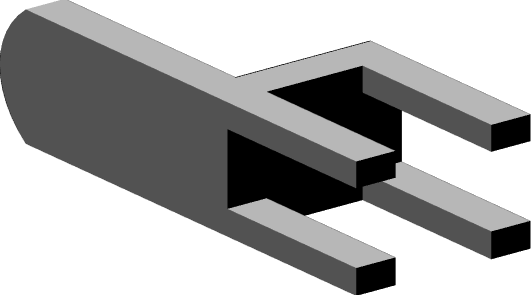}
        \includegraphics[width=0.15\textwidth]{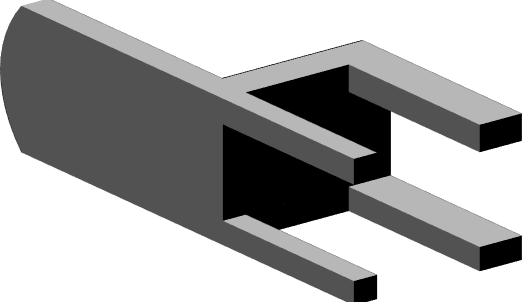}
        \includegraphics[width=0.15\textwidth]{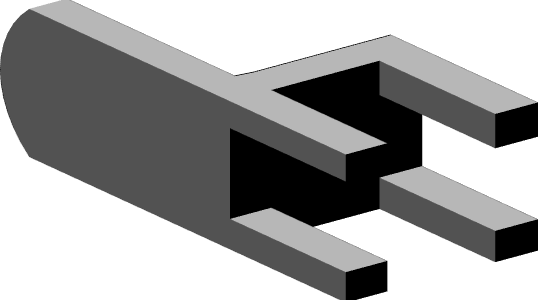}
        \includegraphics[width=0.15\textwidth]{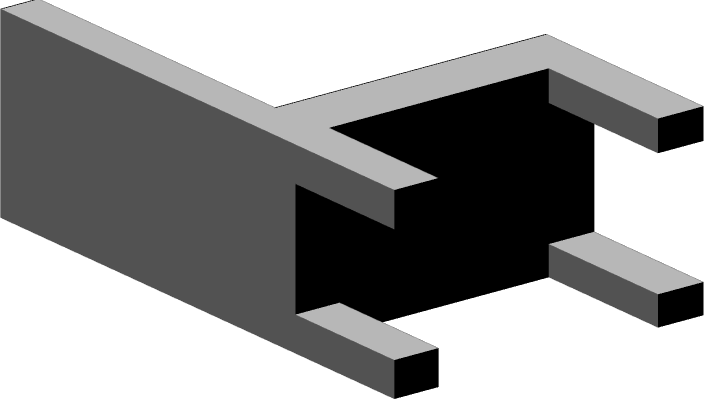}
        \includegraphics[width=0.15\textwidth]{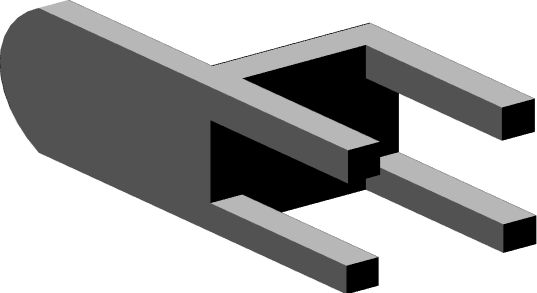}
        \includegraphics[width=0.15\textwidth]{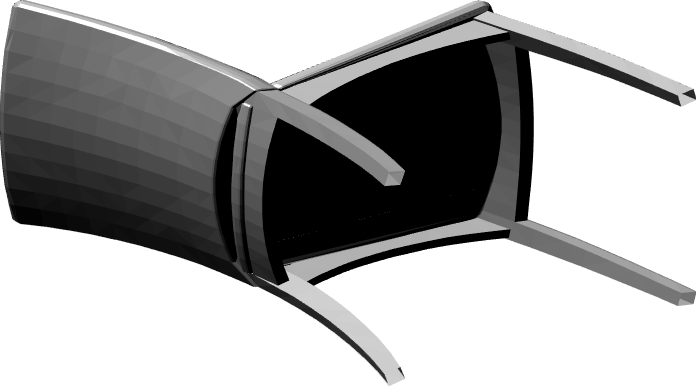}
        
        \includegraphics[width=0.15\textwidth]{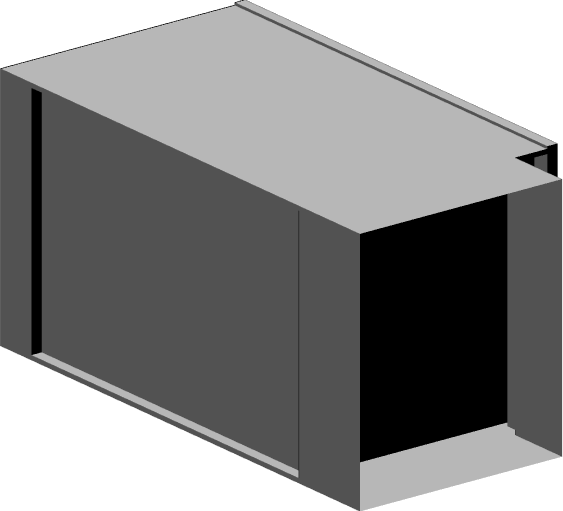}
        \includegraphics[width=0.15\textwidth]{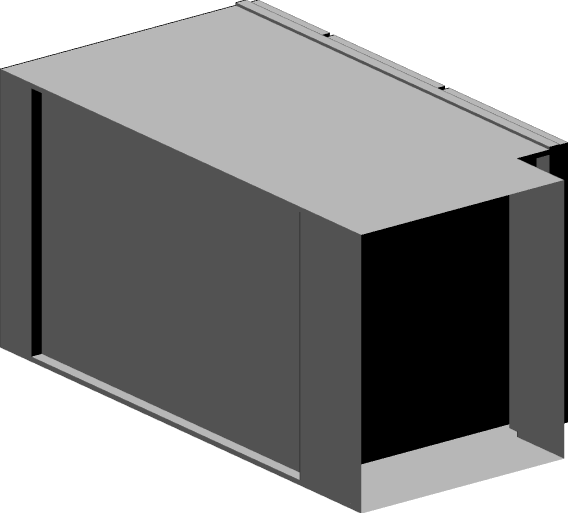}
        \includegraphics[width=0.15\textwidth]{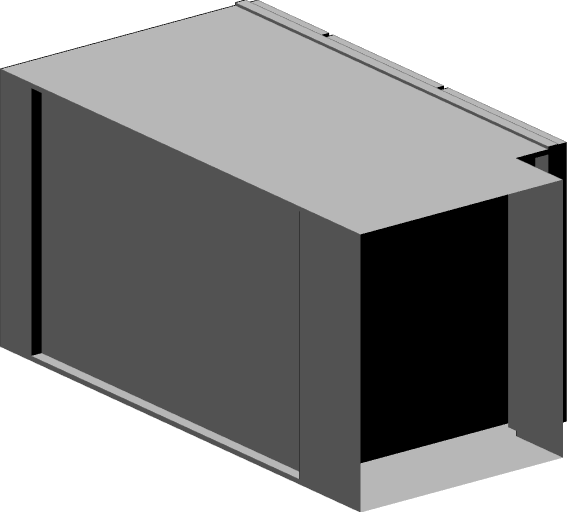}
        \includegraphics[width=0.15\textwidth]{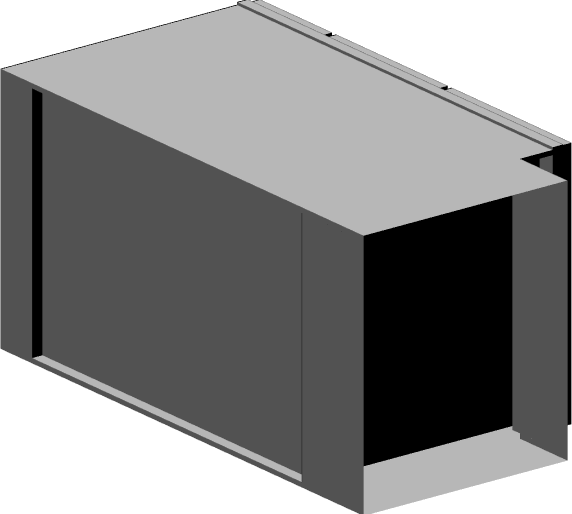}
        \includegraphics[width=0.15\textwidth]{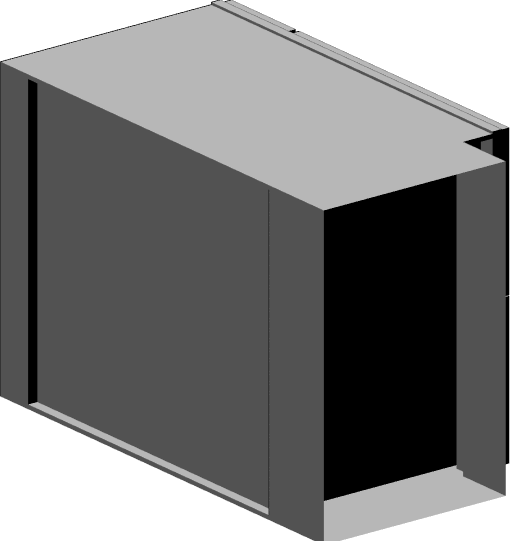}
        \includegraphics[width=0.15\textwidth]{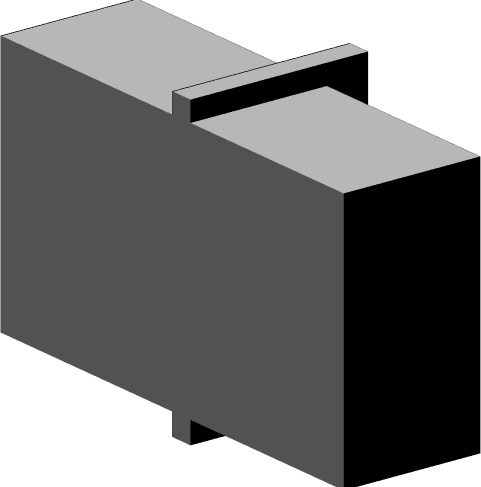}
        
        \includegraphics[width=0.15\textwidth]{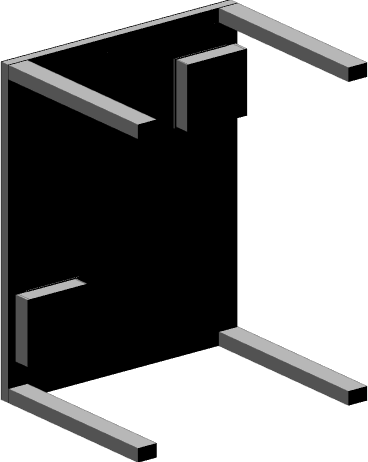}
        \includegraphics[width=0.15\textwidth]{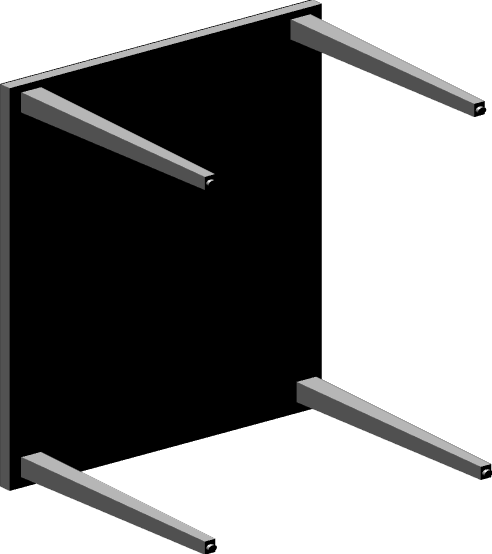}
        \includegraphics[width=0.15\textwidth]{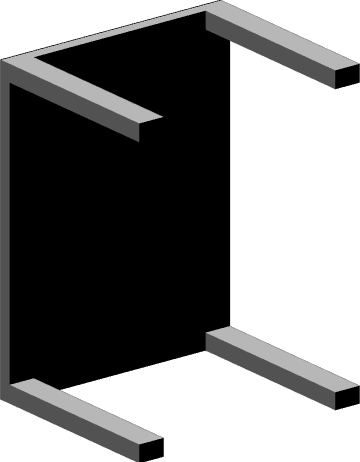}
        \includegraphics[width=0.15\textwidth]{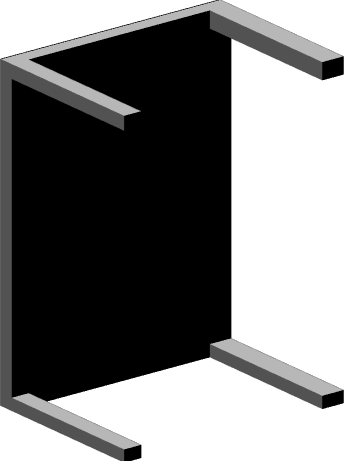}
        \includegraphics[width=0.15\textwidth]{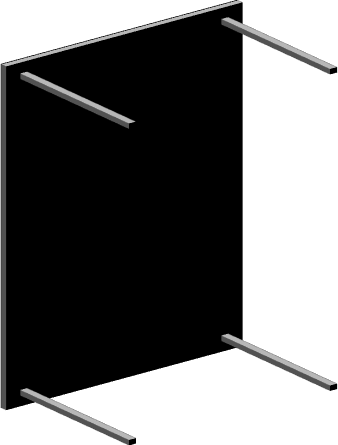}
        \includegraphics[width=0.15\textwidth]{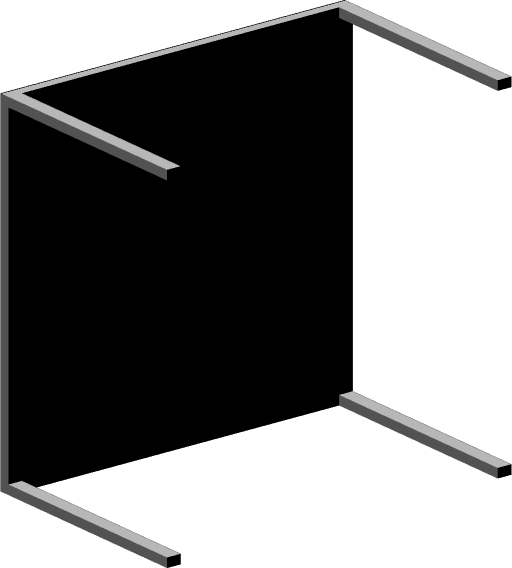}
        
        \includegraphics[width=0.15\textwidth]{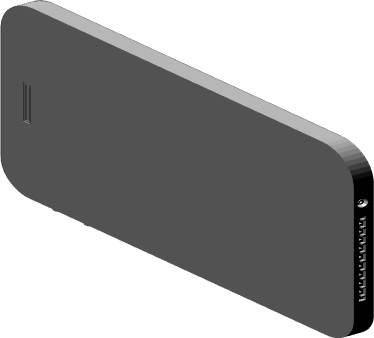}
        \includegraphics[width=0.15\textwidth]{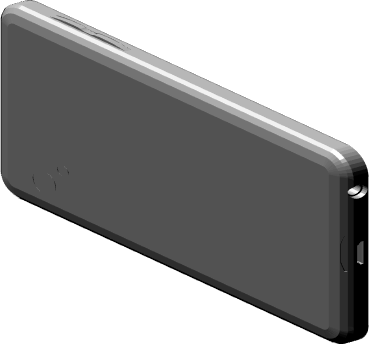}
        \includegraphics[width=0.15\textwidth]{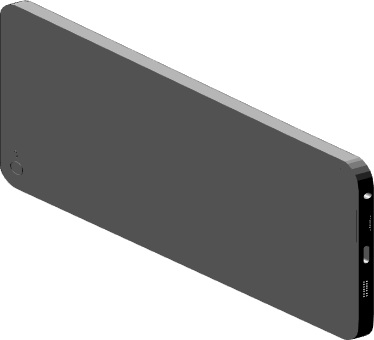}
        \includegraphics[width=0.15\textwidth]{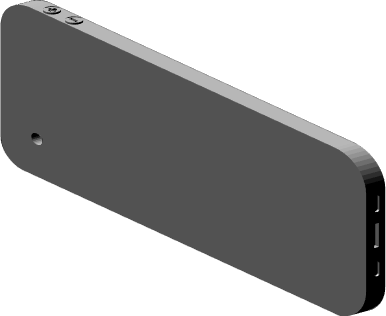}
        \includegraphics[width=0.15\textwidth]{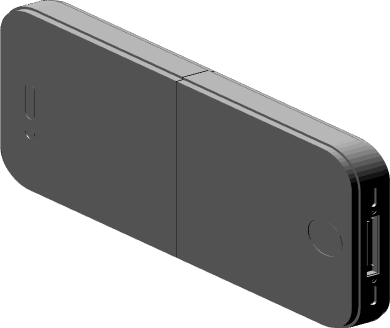}
        \includegraphics[width=0.15\textwidth]{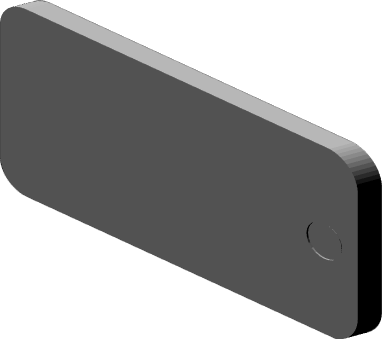}
        
        \includegraphics[width=0.15\textwidth]{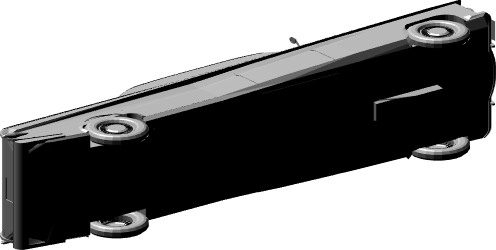}
        \includegraphics[width=0.15\textwidth]{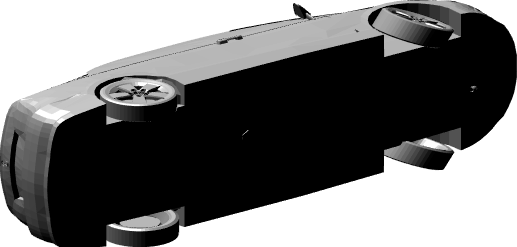}
        \includegraphics[width=0.15\textwidth]{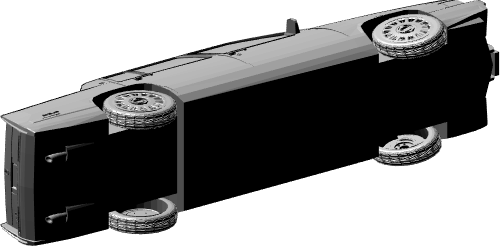}
        \includegraphics[width=0.15\textwidth]{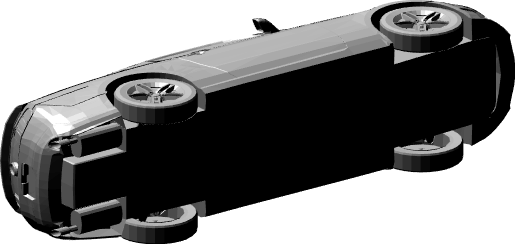}
        \includegraphics[width=0.15\textwidth]{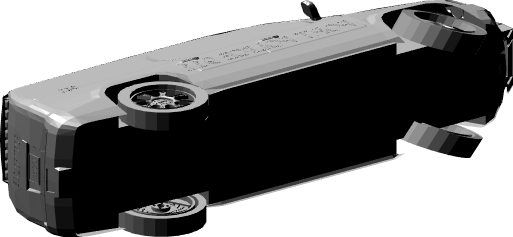}
        \includegraphics[width=0.15\textwidth]{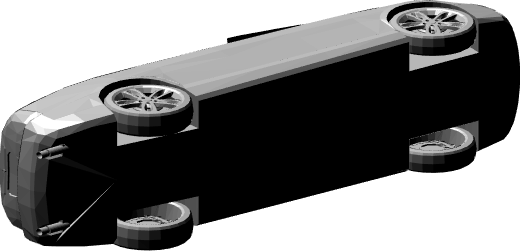}
        
        \includegraphics[width=0.15\textwidth]{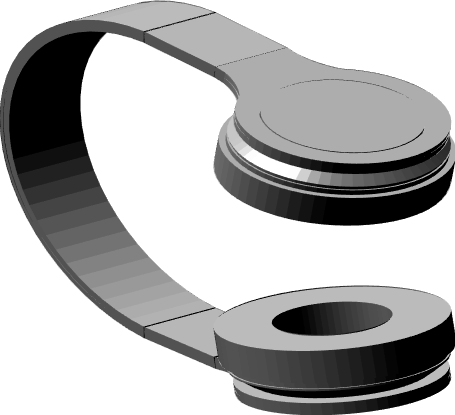}
        \includegraphics[width=0.15\textwidth]{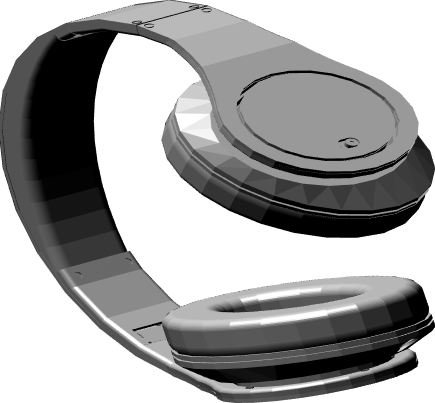}
        \includegraphics[width=0.15\textwidth]{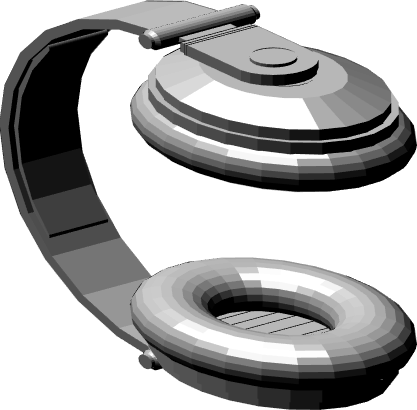}
        \includegraphics[width=0.15\textwidth]{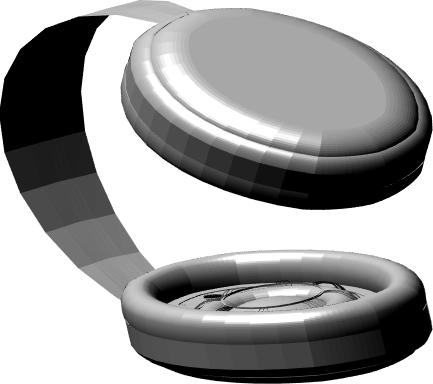}
        \includegraphics[width=0.15\textwidth]{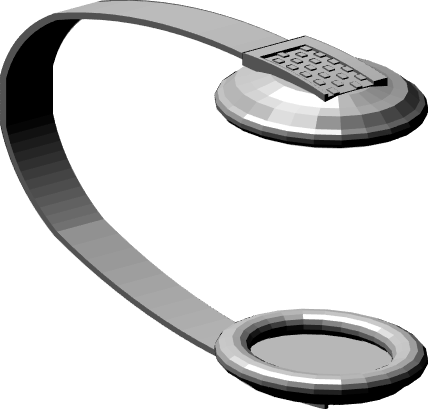}
        \includegraphics[width=0.15\textwidth]{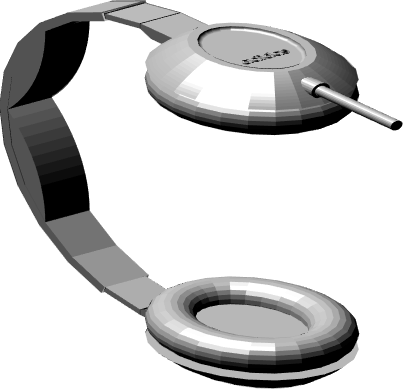}
        
        \caption{Top 5 retrieval results. First column: query shapes. Column 2-6: retrieved shapes ordered by feature similarity.}
        \label{fig_shrec16_visualization}
        \vspace{-4pt}
\end{figure*}

To visualize the shape retrieval results, we present the top 5 retrieval results for a few shapes as shown in Fig.~\ref{fig_shrec16_visualization}

\begin{figure*}[t] 
        \centering
        
        \includegraphics[width=0.12\textwidth]{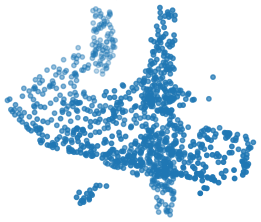}
        \includegraphics[width=0.12\textwidth]{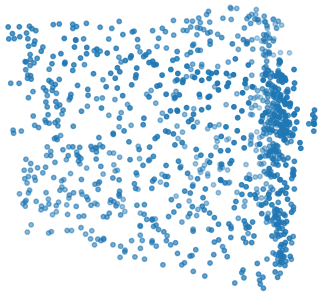}
        \includegraphics[width=0.12\textwidth]{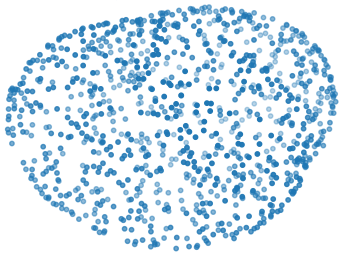}
        \includegraphics[width=0.12\textwidth]{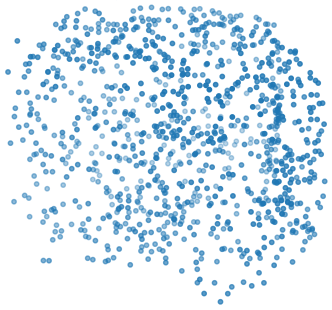}
        \includegraphics[width=0.12\textwidth]{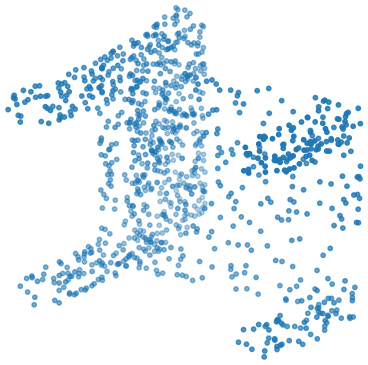}
        \includegraphics[width=0.12\textwidth]{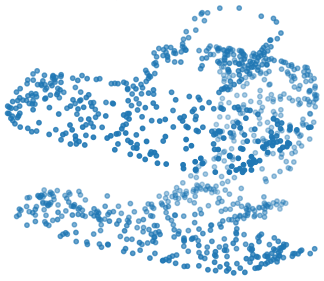}
        \includegraphics[width=0.12\textwidth]{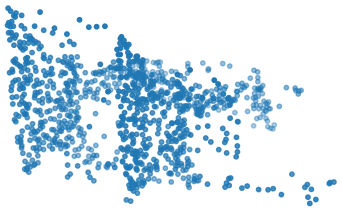}
        \includegraphics[width=0.12\textwidth]{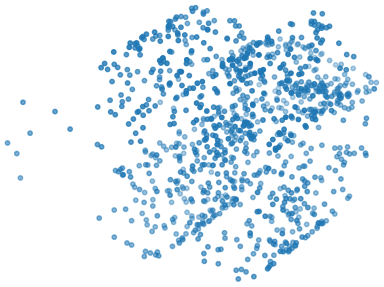}
        
        \includegraphics[width=0.12\textwidth]{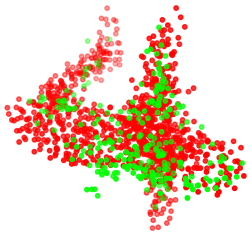}
        \includegraphics[width=0.12\textwidth]{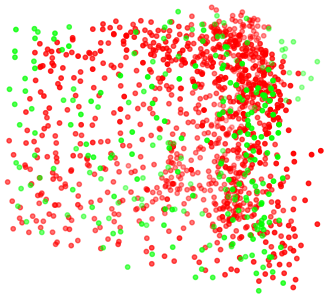}
        \includegraphics[width=0.12\textwidth]{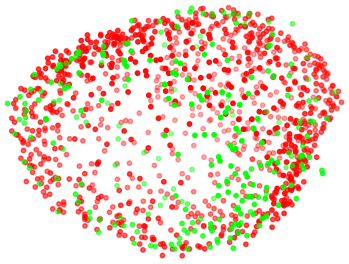}
        \includegraphics[width=0.12\textwidth]{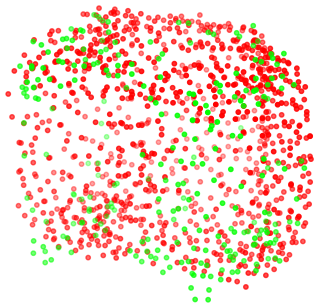}
        \includegraphics[width=0.12\textwidth]{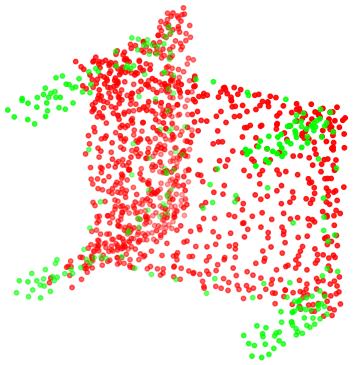}
        \includegraphics[width=0.12\textwidth]{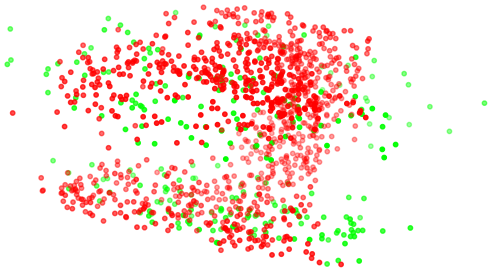}
        \includegraphics[width=0.12\textwidth]{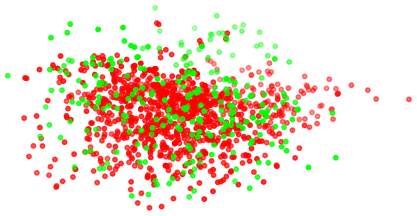}
        \includegraphics[width=0.12\textwidth]{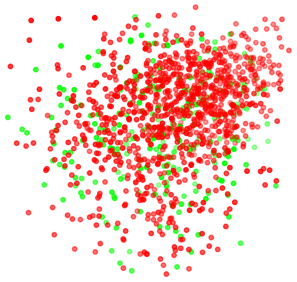}
        
        \includegraphics[width=0.12\textwidth]{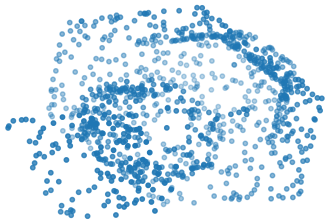}
        \includegraphics[width=0.12\textwidth]{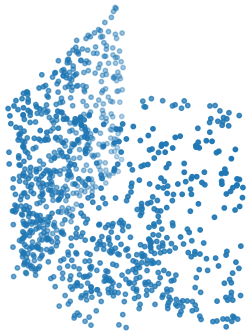}
        \includegraphics[width=0.12\textwidth]{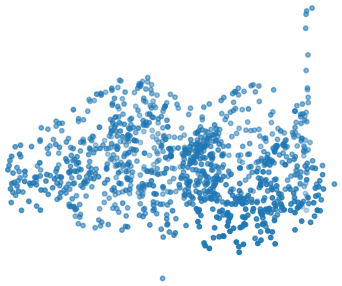}
        \includegraphics[width=0.12\textwidth]{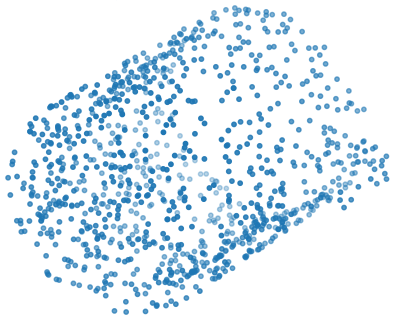}
        \includegraphics[width=0.12\textwidth]{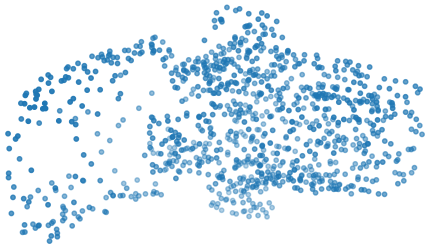}
        \includegraphics[width=0.12\textwidth]{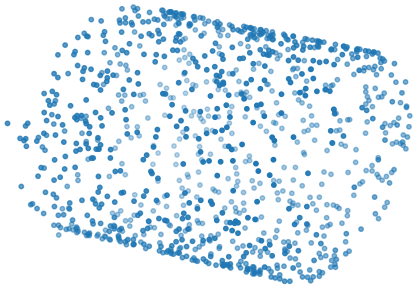}
        \includegraphics[width=0.12\textwidth]{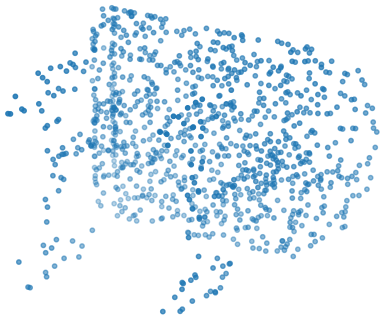}
        \includegraphics[width=0.12\textwidth]{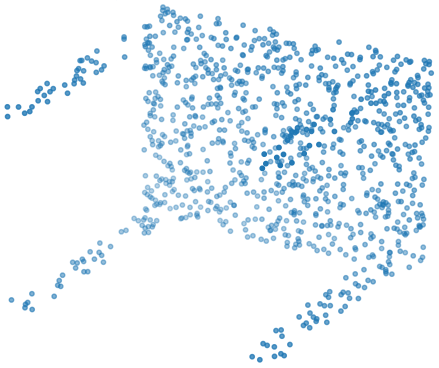}
        
        \includegraphics[width=0.12\textwidth]{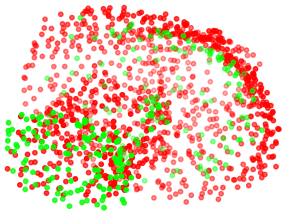}
        \includegraphics[width=0.12\textwidth]{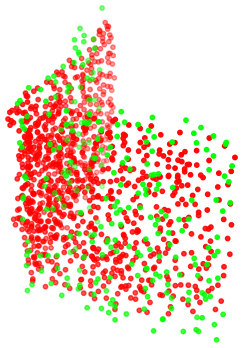}
        \includegraphics[width=0.12\textwidth]{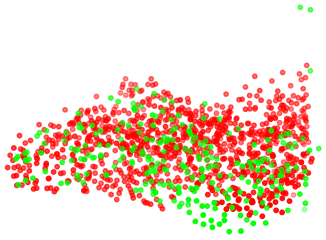}
        \includegraphics[width=0.12\textwidth]{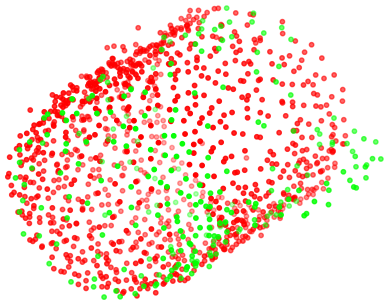}
        \includegraphics[width=0.12\textwidth]{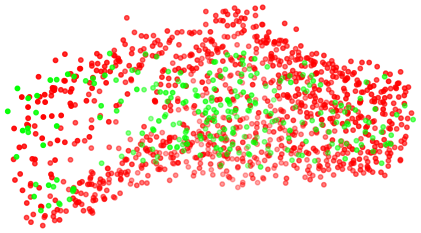}
        \includegraphics[width=0.12\textwidth]{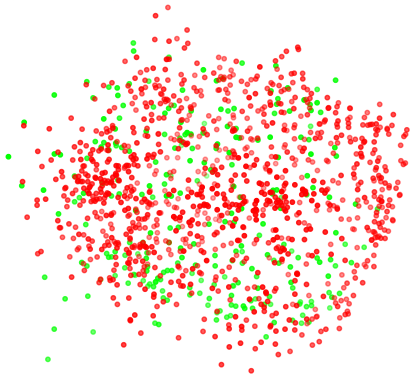}
        \includegraphics[width=0.12\textwidth]{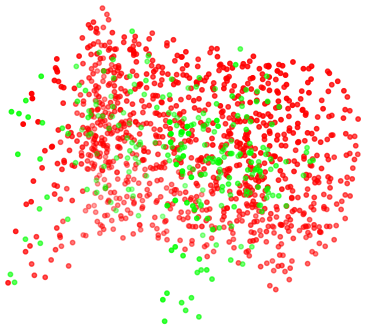}
        \includegraphics[width=0.12\textwidth]{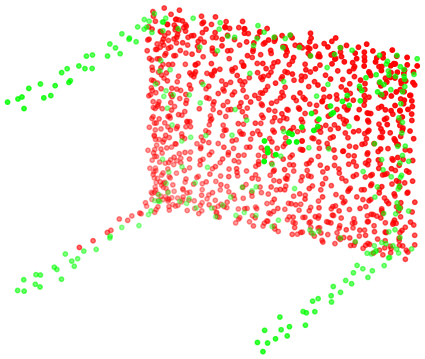}
        
        \caption{Results of our ShapeNetPart autoencoder. Red points are recovered by the convolution branch and green ones are by the fully connected branch. Odd rows: input point clouds. Even rows: reconstructed point clouds.}
        \label{fig_ae_shapenet}
        \vspace{-4pt}
\end{figure*}

\begin{figure*}[t] 
        \centering
        
        \includegraphics[width=0.12\textwidth]{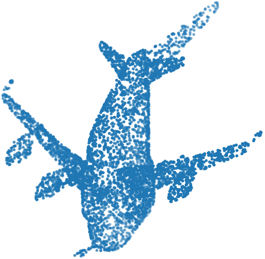}
        \includegraphics[width=0.12\textwidth]{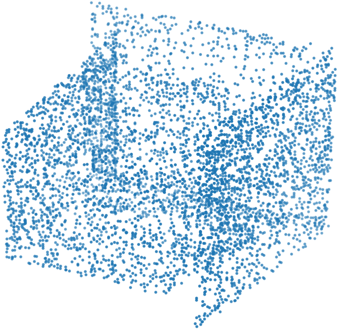}
        \includegraphics[width=0.12\textwidth]{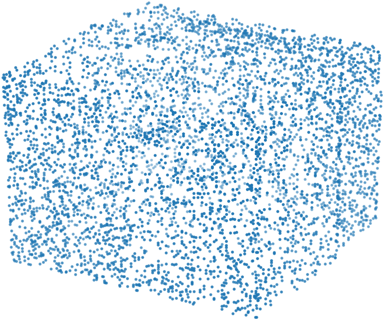}
        \includegraphics[width=0.12\textwidth]{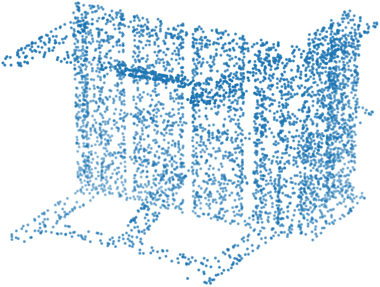}
        \includegraphics[width=0.12\textwidth]{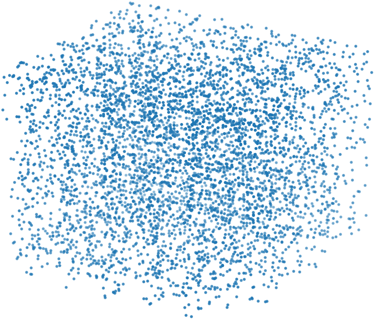}
        \includegraphics[width=0.12\textwidth]{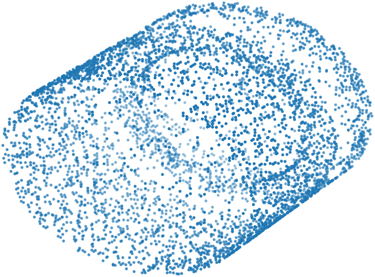}
        \includegraphics[width=0.12\textwidth]{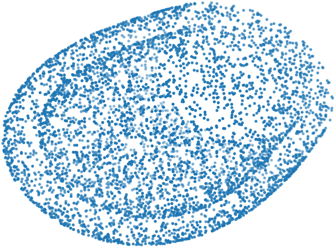}
        \includegraphics[width=0.12\textwidth]{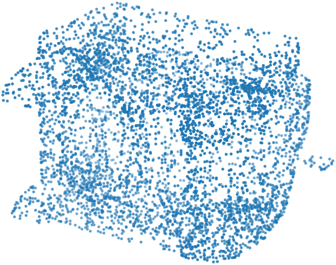}
        
        \includegraphics[width=0.12\textwidth]{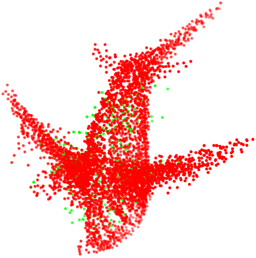}
        \includegraphics[width=0.12\textwidth]{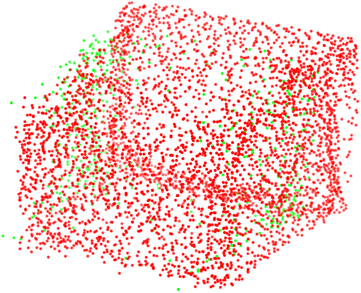}
        \includegraphics[width=0.12\textwidth]{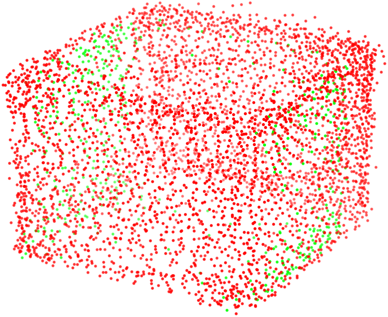}
        \includegraphics[width=0.12\textwidth]{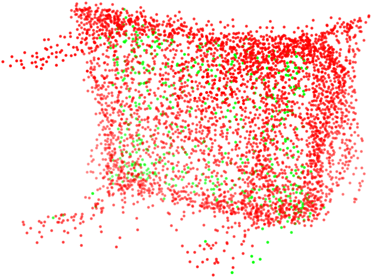}
        \includegraphics[width=0.12\textwidth]{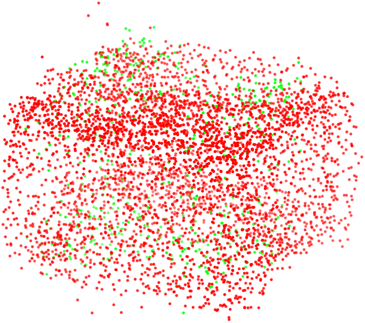}
        \includegraphics[width=0.12\textwidth]{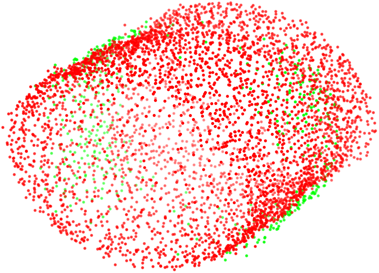}
        \includegraphics[width=0.12\textwidth]{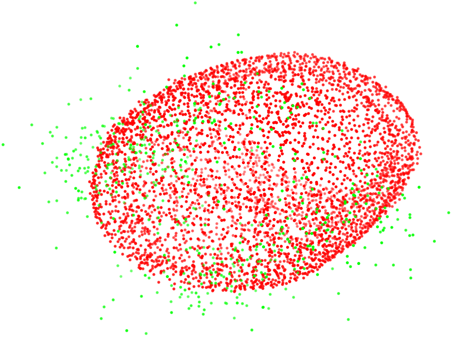}
        \includegraphics[width=0.12\textwidth]{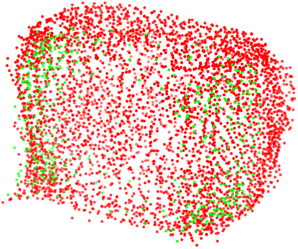}
        
        \includegraphics[width=0.12\textwidth]{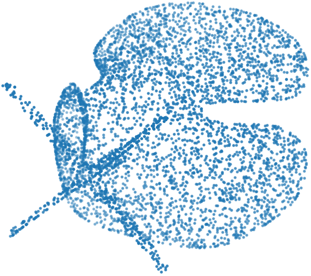}
        \includegraphics[width=0.12\textwidth]{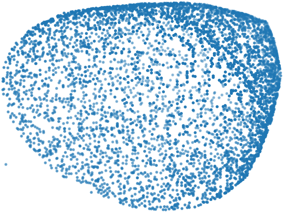}
        \includegraphics[width=0.12\textwidth]{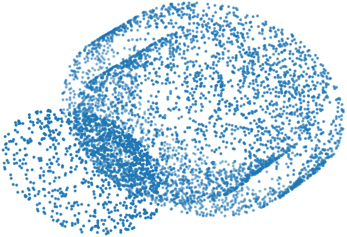}
        \includegraphics[width=0.12\textwidth]{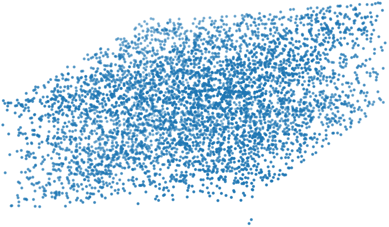}
        \includegraphics[width=0.12\textwidth]{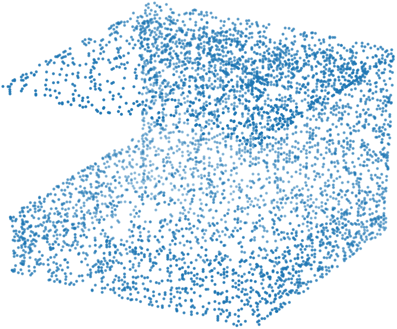}
        \includegraphics[width=0.12\textwidth]{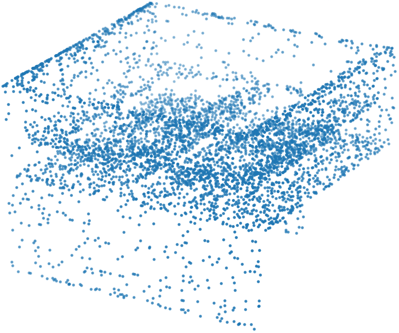}
        \includegraphics[width=0.12\textwidth]{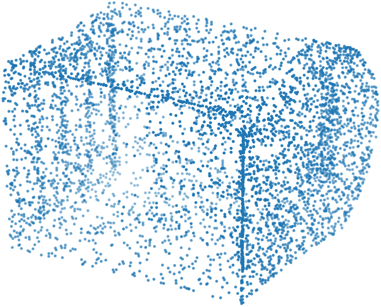}
        \includegraphics[width=0.12\textwidth]{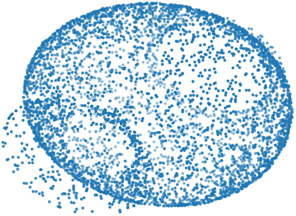}
        
        \includegraphics[width=0.12\textwidth]{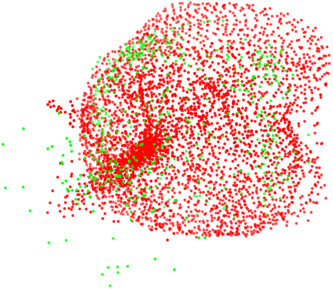}
        \includegraphics[width=0.12\textwidth]{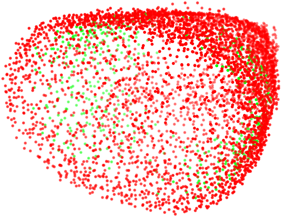}
        \includegraphics[width=0.12\textwidth]{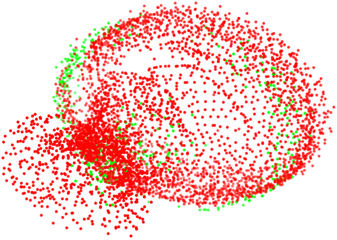}
        \includegraphics[width=0.12\textwidth]{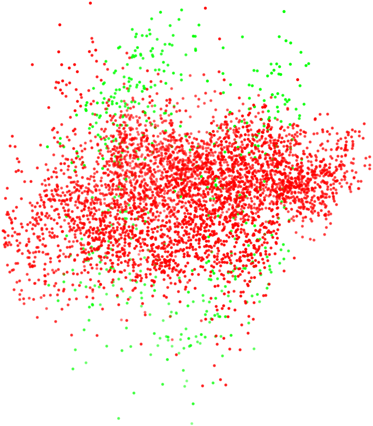}
        \includegraphics[width=0.12\textwidth]{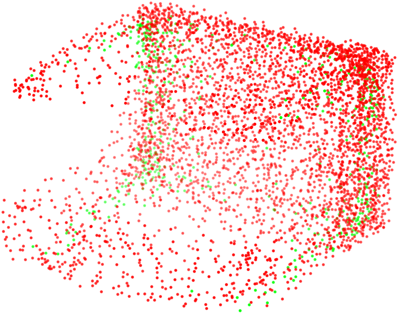}
        \includegraphics[width=0.12\textwidth]{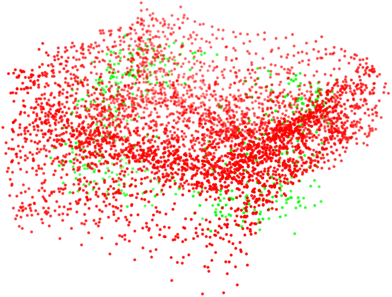}
        \includegraphics[width=0.12\textwidth]{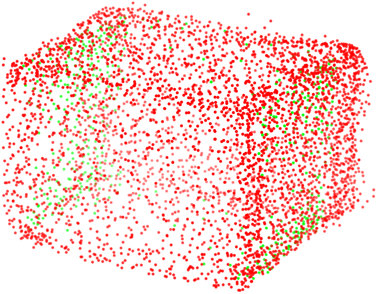}
        \includegraphics[width=0.12\textwidth]{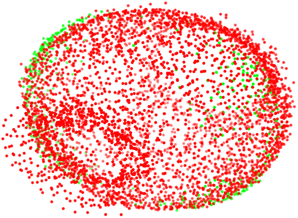}
        
        \includegraphics[width=0.12\textwidth]{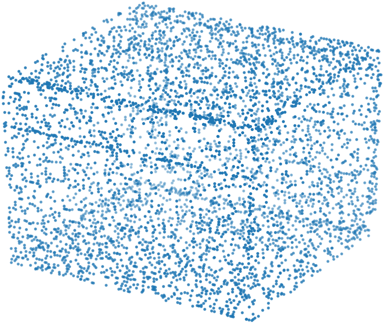}
        \includegraphics[width=0.12\textwidth]{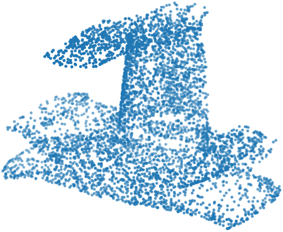}
        \includegraphics[width=0.12\textwidth]{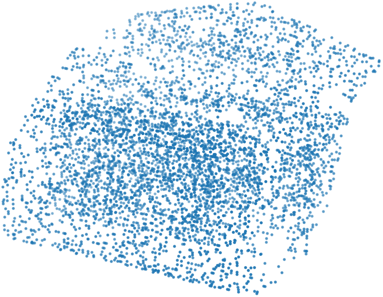}
        \includegraphics[width=0.12\textwidth]{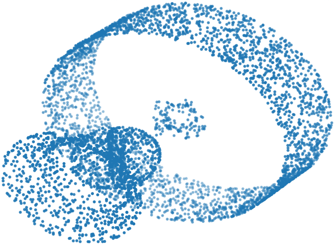}
        \includegraphics[width=0.12\textwidth]{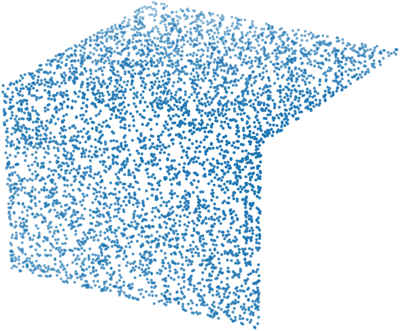}
        \includegraphics[width=0.12\textwidth]{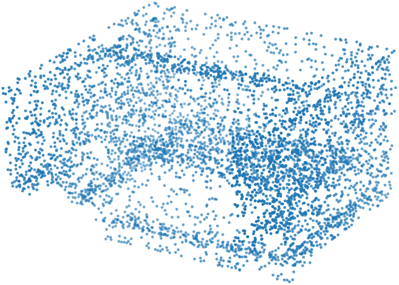}
        \includegraphics[width=0.12\textwidth]{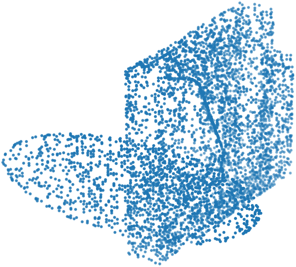}
        \includegraphics[width=0.12\textwidth]{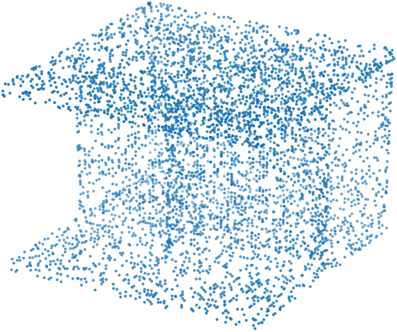}
        
        \includegraphics[width=0.12\textwidth]{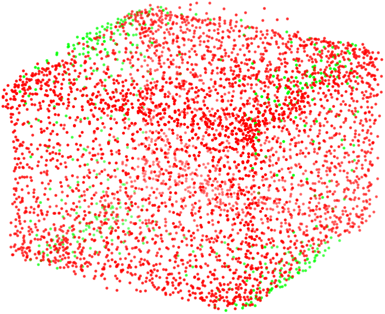}
        \includegraphics[width=0.12\textwidth]{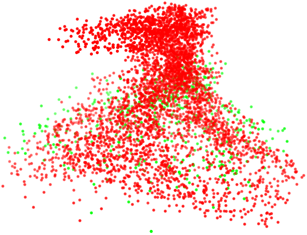}
        \includegraphics[width=0.12\textwidth]{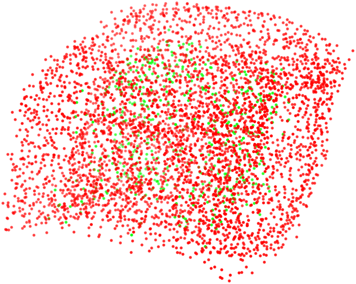}
        \includegraphics[width=0.12\textwidth]{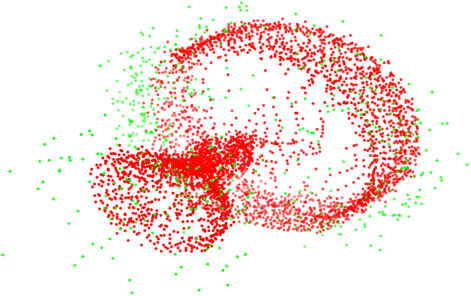}
        \includegraphics[width=0.12\textwidth]{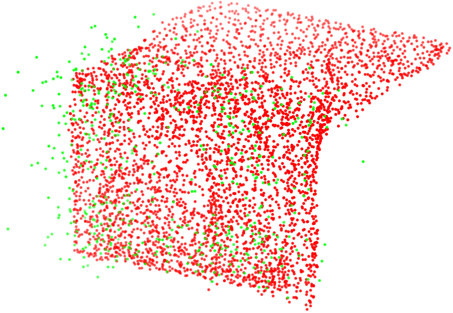}
        \includegraphics[width=0.12\textwidth]{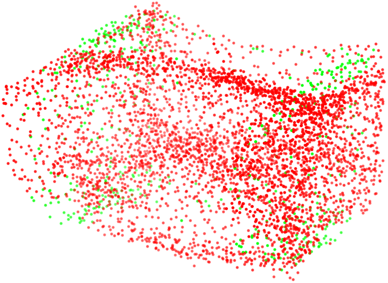}
        \includegraphics[width=0.12\textwidth]{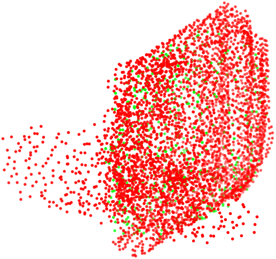}
        \includegraphics[width=0.12\textwidth]{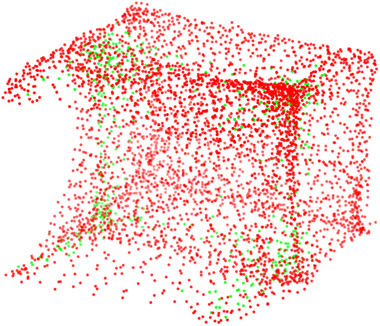}
        
        \includegraphics[width=0.12\textwidth]{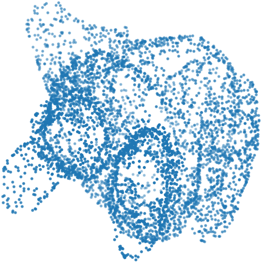}
        \includegraphics[width=0.12\textwidth]{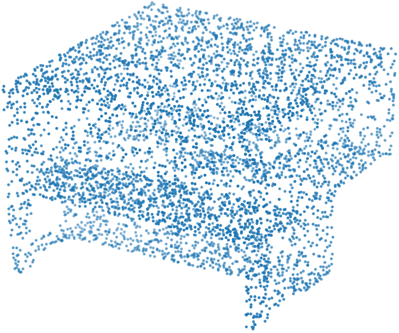}
        \includegraphics[width=0.12\textwidth]{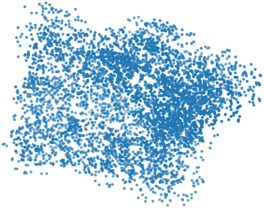}
        \includegraphics[width=0.12\textwidth]{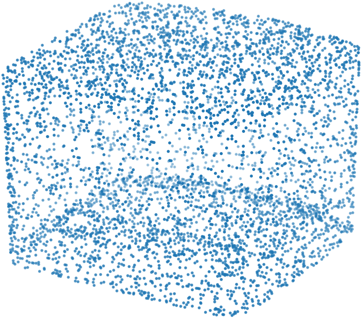}
        \includegraphics[width=0.12\textwidth]{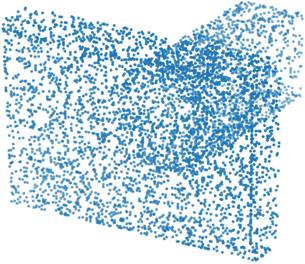}
        \includegraphics[width=0.12\textwidth]{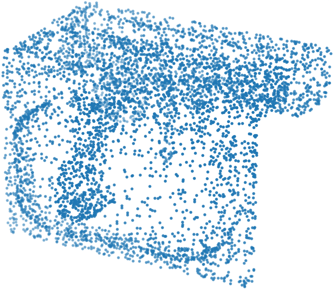}
        \includegraphics[width=0.12\textwidth]{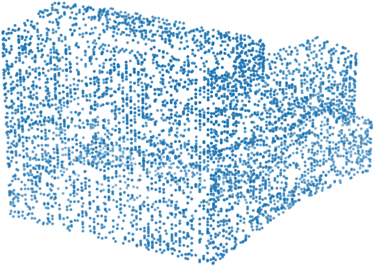}
        \includegraphics[width=0.12\textwidth]{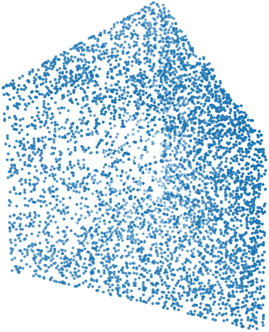}
        
        \includegraphics[width=0.12\textwidth]{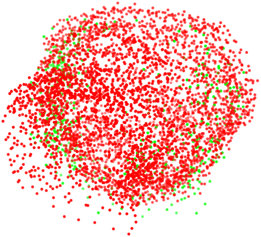}
        \includegraphics[width=0.12\textwidth]{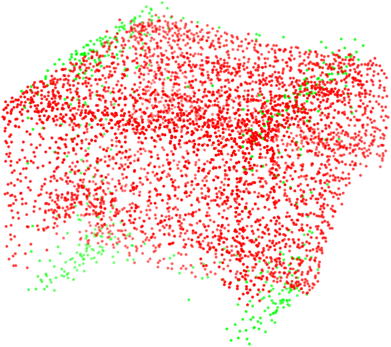}
        \includegraphics[width=0.12\textwidth]{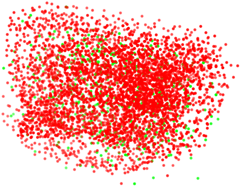}
        \includegraphics[width=0.12\textwidth]{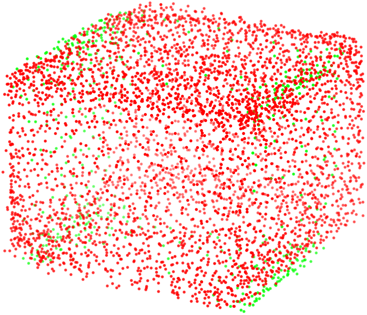}
        \includegraphics[width=0.12\textwidth]{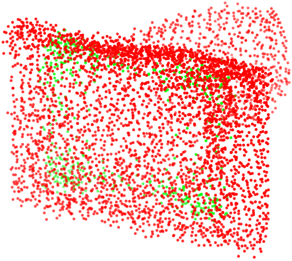}
        \includegraphics[width=0.12\textwidth]{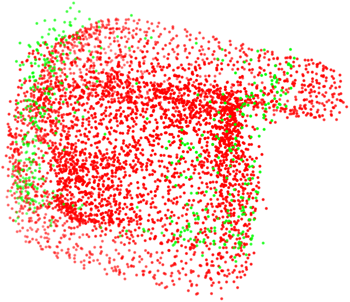}
        \includegraphics[width=0.12\textwidth]{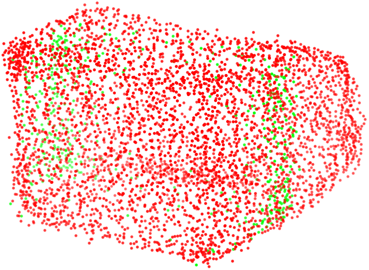}
        \includegraphics[width=0.12\textwidth]{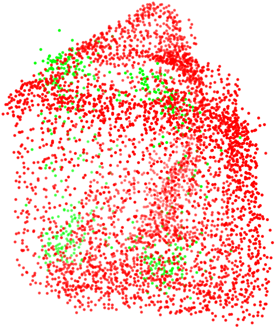}
        
        \includegraphics[width=0.12\textwidth]{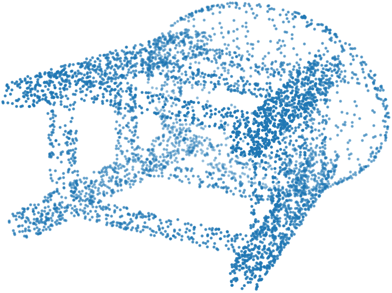}
        \includegraphics[width=0.12\textwidth]{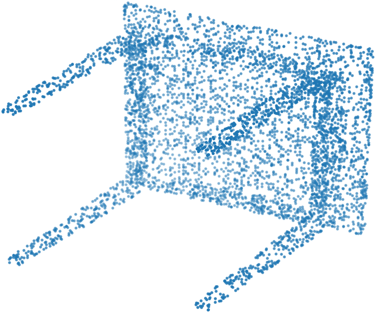}
        \includegraphics[width=0.12\textwidth]{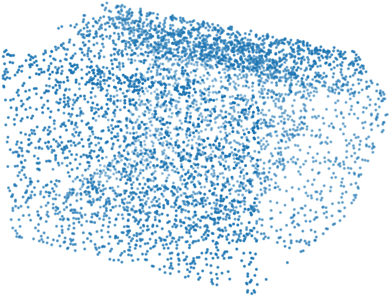}
        \includegraphics[width=0.12\textwidth]{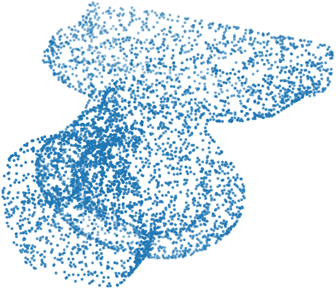}
        \includegraphics[width=0.12\textwidth]{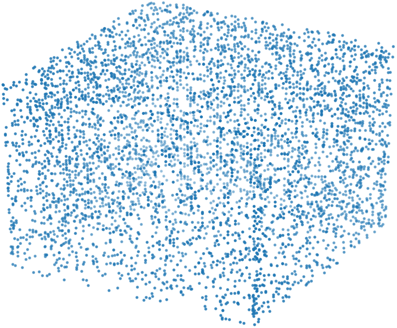}
        \includegraphics[width=0.12\textwidth]{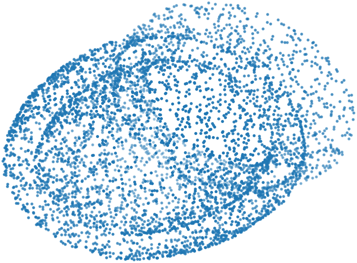}
        \includegraphics[width=0.12\textwidth]{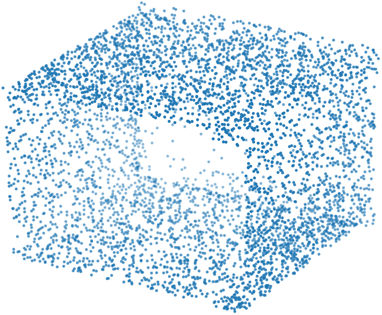}
        \includegraphics[width=0.12\textwidth]{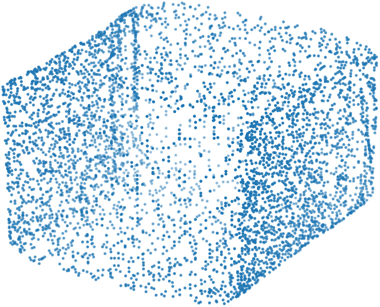}
        
        \includegraphics[width=0.12\textwidth]{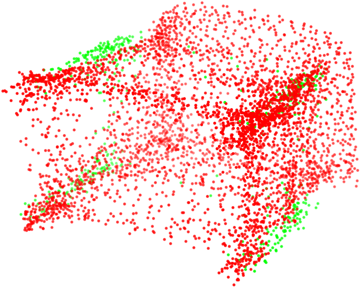}
        \includegraphics[width=0.12\textwidth]{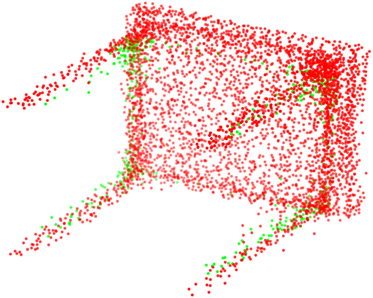}
        \includegraphics[width=0.12\textwidth]{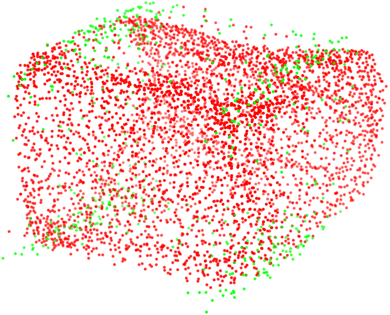}
        \includegraphics[width=0.12\textwidth]{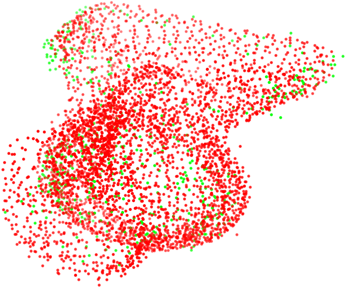}
        \includegraphics[width=0.12\textwidth]{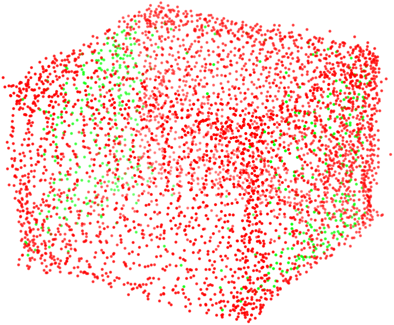}
        \includegraphics[width=0.12\textwidth]{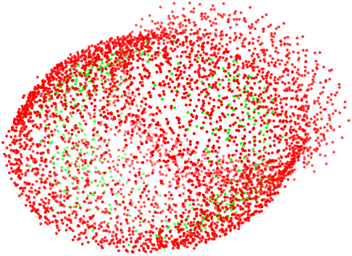}
        \includegraphics[width=0.12\textwidth]{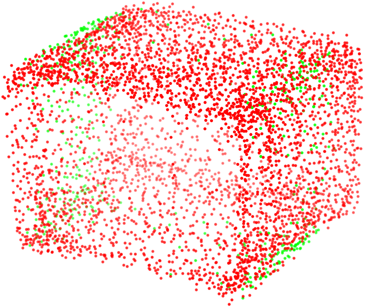}
        \includegraphics[width=0.12\textwidth]{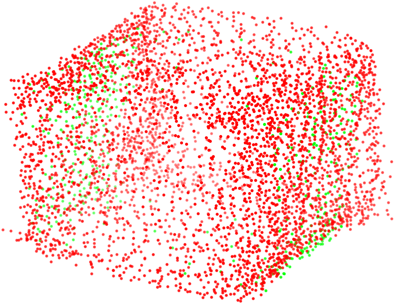}
        \caption{Results of our ModelNet40 autoencoder. Red points are recovered by the convolution branch and green ones are by the fully connected branch. Odd rows: input point clouds. Even rows: reconstructed point clouds.}
        \label{fig_ae_modelnet}
        \vspace{-4pt}
\end{figure*}

For the point cloud autoencoder, we present results from two networks. The first network consumes 1024 points and reconstructs 1280 points with the ShapeNetPart dataset (Fig.~\ref{fig_ae_shapenet}), while the second one consumes 5000 points and reconstructs 4608 points using the ModelNet40 dataset (Fig.~\ref{fig_ae_modelnet}). We present one instance for each category. 

\begin{figure*}[t] 
        \centering
        \includegraphics[width=0.13\textwidth]{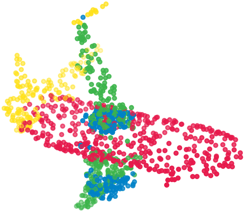} \hspace{10pt}
        \includegraphics[width=0.13\textwidth]{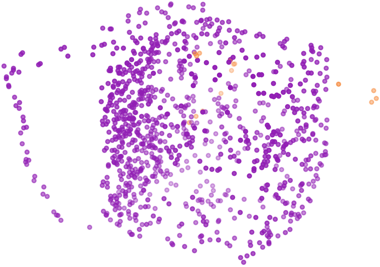} \hspace{10pt}
        \includegraphics[width=0.13\textwidth]{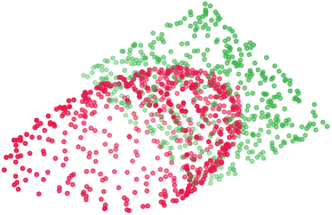} \hspace{10pt}
        \includegraphics[width=0.13\textwidth]{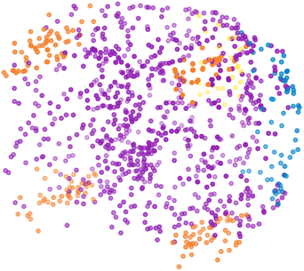} \hspace{10pt}
        \includegraphics[width=0.13\textwidth]{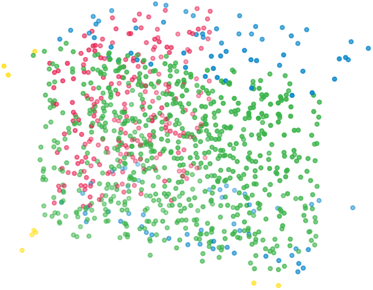} \hspace{10pt}
        \includegraphics[width=0.13\textwidth]{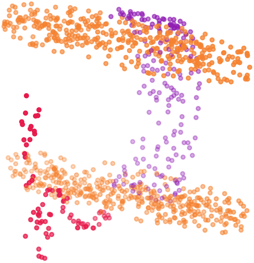}
        
        \includegraphics[width=0.13\textwidth]{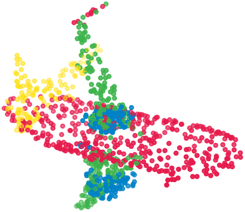} \hspace{10pt}
        \includegraphics[width=0.13\textwidth]{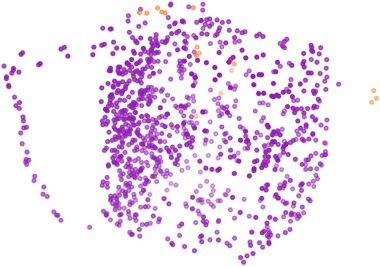} \hspace{10pt}
        \includegraphics[width=0.13\textwidth]{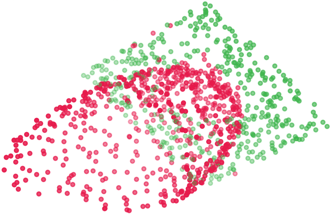} \hspace{10pt}
        \includegraphics[width=0.13\textwidth]{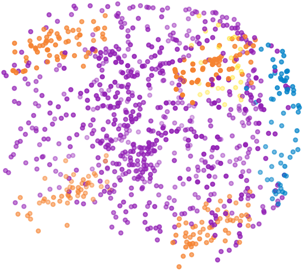} \hspace{10pt}
        \includegraphics[width=0.13\textwidth]{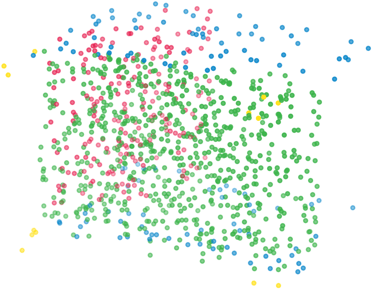} \hspace{10pt}
        \includegraphics[width=0.13\textwidth]{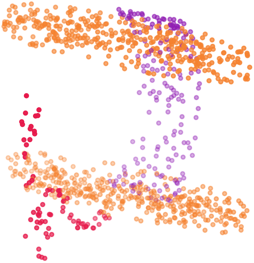}
        
        \vspace{15pt}
        
        \includegraphics[width=0.13\textwidth]{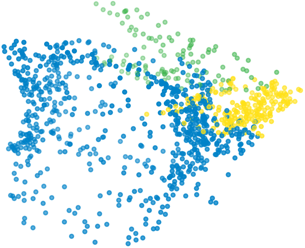} \hspace{10pt}
        \includegraphics[width=0.13\textwidth]{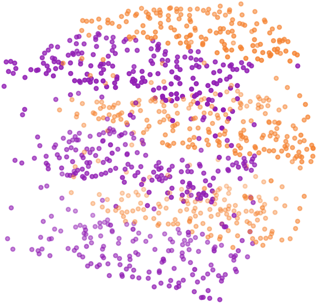} \hspace{10pt}
        \includegraphics[width=0.13\textwidth]{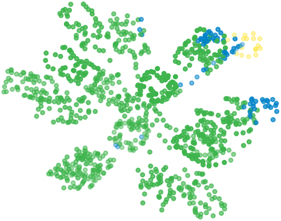} \hspace{10pt}
        \includegraphics[width=0.13\textwidth]{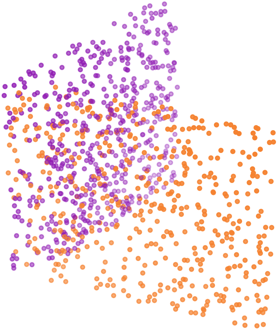} \hspace{10pt}
        \includegraphics[width=0.13\textwidth]{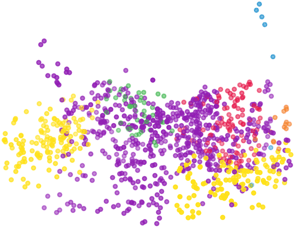} \hspace{10pt}
        \includegraphics[width=0.13\textwidth]{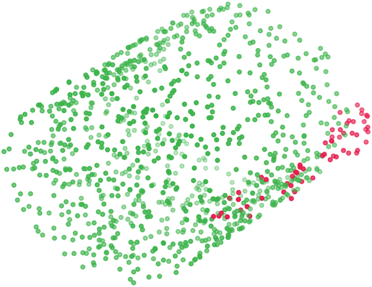}
        
        \includegraphics[width=0.13\textwidth]{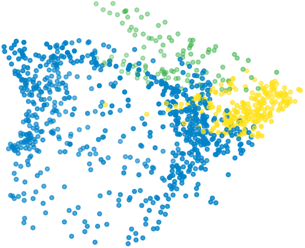} \hspace{10pt}
        \includegraphics[width=0.13\textwidth]{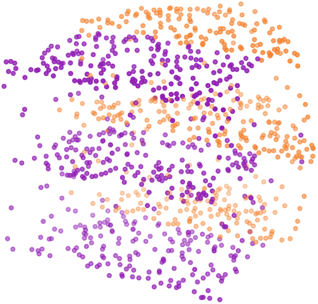} \hspace{10pt}
        \includegraphics[width=0.13\textwidth]{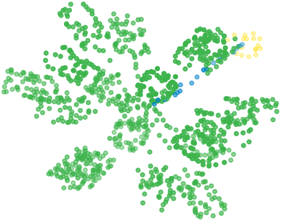} \hspace{10pt}
        \includegraphics[width=0.13\textwidth]{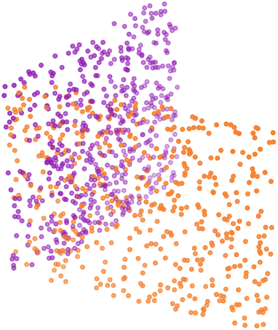} \hspace{10pt}
        \includegraphics[width=0.13\textwidth]{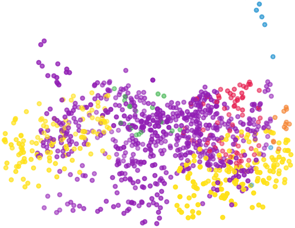} \hspace{10pt}
        \includegraphics[width=0.13\textwidth]{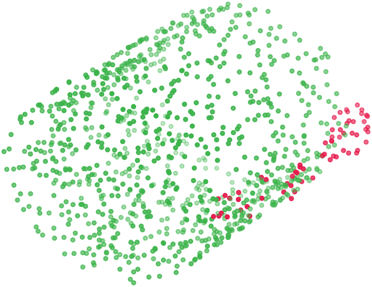}
        
        \vspace{15pt}
        
        \includegraphics[width=0.13\textwidth]{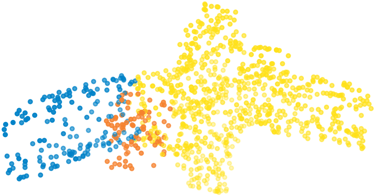} \hspace{10pt}
        \includegraphics[width=0.13\textwidth]{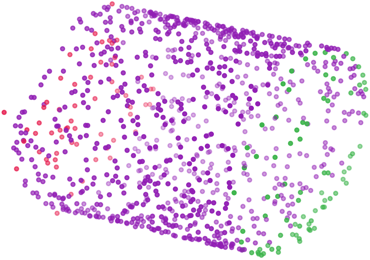} \hspace{10pt}
        \includegraphics[width=0.13\textwidth]{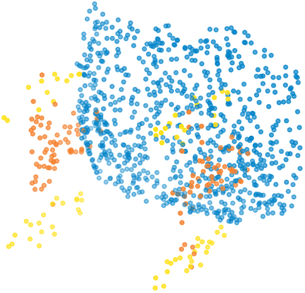} \hspace{10pt}
        \includegraphics[width=0.13\textwidth]{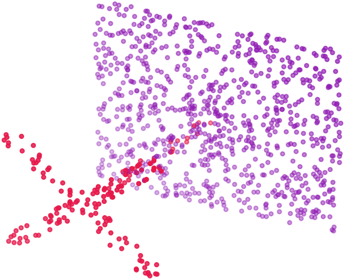}
    
        \includegraphics[width=0.13\textwidth]{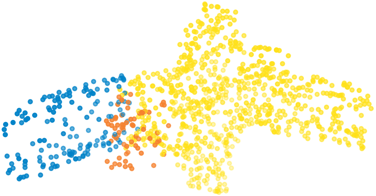} \hspace{10pt}
        \includegraphics[width=0.13\textwidth]{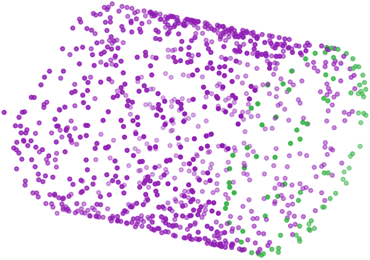} \hspace{10pt}
        \includegraphics[width=0.13\textwidth]{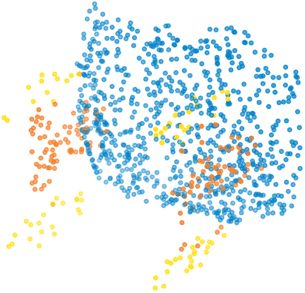} \hspace{10pt}
        \includegraphics[width=0.13\textwidth]{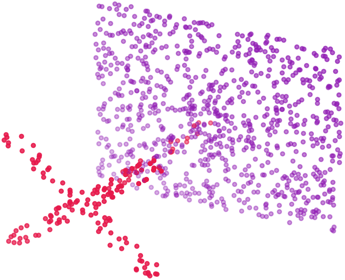}
        
        \caption{Results of object part segmentation. Odd rows: ground truth segmentation. Even rows: predicted segmentation.}
        \label{fig_seg_results}

\end{figure*}

For results of object part segmentation using ShapeNetPart dataset, we visualize one instance per category in Fig.~\ref{fig_seg_results}. The inputs to the network are point clouds of size 1024 and the corresponding surface normal vectors.

\end{document}